\documentclass[preprint,a4paper]{elsarticle}

\usepackage{graphicx}
\usepackage[fleqn]{amsmath}
\usepackage{amsfonts}
\usepackage{amssymb}
\usepackage{float}
\usepackage[dvips]{epsfig}
\usepackage{epstopdf}
\usepackage{rotating}
\usepackage{epigraph}
\usepackage{setspace}

\usepackage{array}
\newcolumntype{P}[1]{>{\centering\arraybackslash}p{#1}}
\newcolumntype{M}[1]{>{\centering\arraybackslash}m{#1}}
\usepackage{hhline}

\usepackage{microtype}
\usepackage{url}
\usepackage{subfiles}
\usepackage{caption}
\usepackage{subcaption}
\usepackage{parskip}
\usepackage[section]{placeins}
\usepackage{xcolor}
\usepackage{framed}
\colorlet{shadecolor}{blue!20}
\usepackage{tabulary}
\usepackage[hidelinks]{hyperref}
\usepackage[capitalize,nameinlink]{cleveref}
\usepackage{flexisym}

\usepackage{amsmath,amsthm,amssymb}

\usepackage{calrsfs}
\DeclareMathAlphabet{\pazocal}{OMS}{zplm}{m}{n}

\usepackage{multirow, bigdelim}

\usepackage{tikz}
\usetikzlibrary{decorations.pathmorphing}
\usetikzlibrary{decorations.markings}
\usetikzlibrary{arrows.meta,bending}
\usetikzlibrary{arrows.meta}
\usepackage{bm}

\usepackage{adjustbox}

\usepackage{makecell}

\journal{Data-centric engineering}

\begin{document}
	
\begin{frontmatter}
		
				\title{On generative models as the basis for digital twins}

    \author[1]{G.\ Tsialiamanis\footnote{Corresponding Author: George Tsialiamanis (g.tsialiamanis@sheffield.ac.uk)}}
    \author[1]{D.J.\ Wagg}
    \author[1]{N.\ Dervilis}
    \author[1]{K.\ Worden}
    \address[1]{Dynamics Research Group, Department of Mechanical Engineering, University of Sheffield \\ Mappin Street, Sheffield S1 3JD}

\begin{abstract}
    
A framework is proposed for generative models as a basis for \textit{digital twins} or \textit{mirrors} of structures. The proposal is based on the premise that deterministic models cannot account for the uncertainty present in most structural modelling applications. Two different types of generative models are considered here. The first is a physics-based model based on the stochastic finite element (SFE) method, which is widely used when modelling structures that have material and loading uncertainties imposed. Such models can be calibrated according to data from the structure and would be expected to outperform any other model if the modelling accurately captures the true underlying physics of the structure. The potential use of SFE models as digital mirrors is illustrated via application to a linear structure with stochastic material properties. For situations where the physical formulation of such models does not suffice, a data-driven framework is proposed, using machine learning and conditional generative adversarial networks (cGANs). The latter algorithm is used to learn the distribution of the quantity of interest in a structure with material nonlinearities and uncertainties. For the examples considered in this work, the data-driven cGANs model outperform the physics-based approach. Finally, an example is shown where the two methods are coupled such that a hybrid model approach is demonstrated.
\end{abstract}
		

\end{frontmatter}
	
	
\section{Introduction}
\label{sec:introduction}
\label{sec:intro}
A recent innovation in the field of system simulation is the creation of \textit{digital twins} for specific systems (called physical twins). For example, attempts have been made to do this in the fields of manufacturing \citep*{Rosen2015, Uhlemann2017}, control systems \& the internet of things \citep*{8289327}, smart cities \citep*{dembski2020urban}, social networks, and management \citep*{macchi2018exploring} --- more detailed literature reviews and descriptions of state-of-the-art research relating to digital twins can be found in the recent review papers \citep*{fuller2020digital,jones2020characterising,wagg2020digital}. Structural dynamics is also a field in which digital twins have been a desired achievement for a number of years --- see \citep*{wagg2020digital} and references therein. One of the motivations for creating a digital twin of a structure (the physical twin) is to enable more accurate prediction of the structure's behaviour under a wider range of different situations. For example, predictions could be used to avoid scenarios under which the structure might be more likely to suffer damage or degradation. Equivalently, in the extreme case, the model might be used to limit the use of the structure in operating conditions where one might be concerned that some form of structural failure might occur (for example using a wind turbine in higher wind speeds than usual). In this context, the overall goal of a digital twin can be viewed as maximising the effective operational life of the structure, and as such, is directly linked to the business objective of minimising cost (or maximising profit) associated with the physical twin.

For complex engineering applications, it is not possible to have complete knowledge of all of the physics of the structure, including all its possible environmental and operational conditions.
Therefore, one of the underlying concepts of a digital twin is that a combination of models is used to capture the overall behaviour of the physical twin.
In particular, a commonly-proposed scenario is that physics-based model(s), such as finite-elements, are combined with data-based techniques, such as machine learning  \citep*{Bishop2, Murphy:2012:MLP:2380985}.
In addition to this, models can be defined for different parts (or sub-structures) of a physical twin and then assembled into a larger model.
This type of assembled modelling approach was discussed in \citep*{worden2020digital}, where the concept of a \textit{digital mirror} was also introduced in order to give a more precise mathematical set of definitions,  and these definition will be used as the framework for the results presented in this paper.

A very important part of building a digital twin is to consider the associated uncertainty of the process. This aspect includes both aleatory uncertainty, which refers to events or quantities that are inherently random and cannot be modelled using deterministic physics-based models (for example measurement uncertainty \citep*{smithUQ}), and epistemic uncertainty which relates to a lack of knowledge about the properties of the physical twin.
A common example of epistemic uncertainty is when the effects of nonlinearity are not captured in a physics-based model, leading to errors between the data acquired from the real structure and the model. Another type of epistemic uncertainty is not knowing all the variables that affect the result of an event. In machine learning these variables are sometimes referred to as ``lurking'' variables \citep*{Bishop2}. Similarly, in probability-based models for uncertainty, such variables are called latent variables \citep*{booyse2020deep}.

Despite the separate definitions of aleatory and epistemic uncertainty, it will typically be a very challenging problem to quantify these separately within a digital twin.
Therefore, the motivation for this work is to use generative models as the basis for a digital twin that can provide estimations of aleatory and epistemic uncertainty.
The probabilistic framework of generative models fits naturally with models for aleatory uncertainty, and epistemic uncertainties can be inferred based on variations between the digital twin outputs and recorded data from the physical twin.
A related approach has been developed in \citep*{booyse2020deep} in order to build a black-box digital twin for a structural health monitoring application.

In the current work, two different types of generative models are studied; the first using the stochastic finite element (SFE) method  \citep*{ghanem2003stochastic, sudret2000stochastic}. SFE models are used to propagate uncertainty from material and loading to quantities of interest via finite element models. They are \textit{white-box} models, directly exploiting knowledge of the physics of the structure.
The second type of generative model considered is the \textit{conditional generative adversarial network} (cGAN) \citep*{mirza2014conditional}. Using this algorithm, one can generate samples of learnt distributions, conditioned on a set of variables. These distributions are of the structural quantities that the model is built to predict. Because cGAN is a machine learning algorithm, it should be able to perform for a wide range of structural or environmental conditions for which there are data; this is in contrast to SFE models, which are able to perform only under pre-defined conditions. However, the machine learning model is limited to a set of system conditions for the data gathered, and cannot extrapolate beyond this, which is in contrast to the SFE model that can be used in a wider predictive role.
A hybrid approach, using both generative models, is an attempt to get the best aspects of both models. Specifically, what is usually expected from hybrid approaches (grey-box models) is: (i) to use the cGAN algorithm to correct the discrepancy of the SFE model in cases where the physical formulation of the finite elements do not suffice, and (ii) to allow the hybrid model to have some predictive capability based on the SFE model away from the operational conditions where data are available; i.e.\ extrapolation capability.

The main thesis of this paper is that one should adopt generative models to properly accommodate uncertainty in potential digital twins. In order to present this idea,
specific modelling technologies are used to illustrate the various shades: stochastic FE for a generative white box and the cGAN for a generative black box; combined
together, these present a fully-generative grey box. The presentation is not intended to suggest that these model types are the {\em only} possibilities; in fact, a
range of paradigms could prove equally powerful. In terms of white generative models, a fairly basic implementation of the stochastic FE method has been presented here,
based on the early polynomial-chaos formulation of \cite{ghanem2003stochastic}; however, more recent variants like the stochastic Galerkin approach
\citep*{Augustin}, have advantages like more general expansion bases for the stochastic space. A very recent methodology {\em StatFEM} \citep*{Girolami1,Girolami2},
provides an elegant Bayesian framework for both building and updating generative FE models. In terms of generative black-box models, GANs are by no means the only option;
in fact one alternative -- the {\em variational auto-encoder} \citep*{Kingma} -- has already proved to be generally useful in engineering problems; particularly in
condition monitoring problems, e.g., \citep*{Mylonas}. Another versatile generative framework is provided by {\em Gaussian processes} (GPs) \citep*{Rasmussen}. Although
GPs are most often used as nonparametric black-box learners, they are also the basis for the StatFEM models mentioned earlier. Furthermore, GPs offer a direct method
of building grey-box models by training them from data, but building {\em a priori} physics into their mean and kernel functions \citep*{Cross4,Cross5,Cross6,Cross7}.

The layout of the paper is as follows. In Section \ref{sec:mirrors}, mirrors are defined based on the work in \citep*{worden2020digital}. In Section \ref{sec:GANs}, details of the GANs and cGANs are given. In Section \ref{sec:SFEM_mirror}, a white-box mirror, based on a SFEM model is presented. In Section \ref{sec:cGAN_DT}, a black-box mirror, based on a cGAN is described. In Section \ref{sec:hybrid}, the combination of the two models into a hybrid mirror is presented and also the extrapolation potential of both the hybrid and the black-box mirrors is studied. Finally, the results are summarised and conclusions are drawn. Further details about the SFE method can be found in the Appendix.

\section{Digital mirrors}
\label{sec:mirrors}

Although the term digital twin has been widely used in many disciplines and in industry sectors, an alternative terminology is used here in order to make the subsequent analysis more precisely defined.
The physical twin (also called the structure and denoted $S$) has $N_{S}$ different states $\underline{s} = \{s_{1}, s_{2}, ...s_{N_{S}}\}$ (in contrast to \citep*{worden2020digital} where each state refers to a time instant $t$, the notation here is simplified for convenience and since the problems to be presented here are not dynamic but static).
Together with the structural states, the environment $E$ of the structure has $N_{E}$ corresponding states, $\underline{e} = \{e_{1}, e_{2}, ...e_{N_{E}}\}$. In general, the set $\underline{s}$ may contain specific displacements (or accelerations when dynamic problems are considered) that are of interest from a structure, or stresses, strains and response spectra that one might monitor in a structural health monitoring (SHM) scheme. The set $\underline{e}$ may include environmental conditions affecting the structure such as temperature, humidity, wind speed etc.

The approach taken in the current work is to build mirrors that predict the behaviour of different parts of the structure $S$, rather than a global model for the whole $S$. This subset of states/quantities that are mirrored define the \textit{context} $C = \{e^{C}_{i} \in E, s^{C}_{j} \in \underline{s};i, j\}$, where $s^{C}_{j}$ is the \textit{response} or \textit{predictive context} and $e^{C}_{i}$ is the environmental context. This formulation is used to define the exact quantities that the model is able to predict and the exact environmental conditions under which the model is able to perform.

Following \citep*{worden2020digital}, definitions are constructed according to the mirror's ability to predict the states $s_{j}^{C}$. This ability is measured using  metrics defined below, and based on this the mirror models can be considered to be either \textit{$\epsilon$-mirrors} or \textit{$\alpha$-mirrors}. A model is considered to be an $\epsilon$-mirror if a distance metric, $d^{C}$, is less than (or equal to) a predefined tolerance, $\epsilon$, such that
\begin{equation}
    \label{eq:epsilon_mirror}
    d^{C}(\underline{p}^{C}, \underline{r}^{C}) \leq \epsilon
\end{equation}
where $\underline{p}^{C}$ is the prediction of the digital mirror within some context $C$ and $\underline{r}^{C}$ is the observation of the response of the structure. The definition simply implies that the response of the physical structure should be within some distance of the prediction of the mirror. If the mirror is based on a deterministic model, then the distance defines some interval or area in the prediction space similar to confidence intervals. Given that the models that are studied here are probabilistic, $d^{C}$ should be some probability distribution distance metric defining the maximum distance between the predicted and real probability distributions of interest.

The second type of mirror is the $\alpha-$mirror. In order for stochastic model to be considered an $\alpha-$mirror, the quantity of interest of the real structure should always be within an interval defined by the output of the model with a given probability $p$, i.e.,
\begin{equation}
    \label{eq:alpha_mirror_prob}
    P(r^{C}_{i} \in [\bar{m}^{C}_{M} - \alpha \sigma^{C}_{M}, \bar{m}^{C}_{M} + \alpha \sigma^{C}_{M}]) = P(\alpha)
\end{equation}
where $r^{C}_{i}$ is an observation, $P$ is the probability that the observation is within the defined interval, $M$ is the model used as a mirror, $m^{C}_{M}$ is the prediction of the model or the mean value of the outputs in the case of a generative model $M$, $\sigma_{M}$ is the corresponding standard deviation, $P(\alpha)$ the probability as a function of the predefined parameter $\alpha$, which controls how wide the interval defined in equation (\ref{eq:alpha_mirror_prob}) is. The probability $P$ is a function of the parameter $\alpha$ and for every mirror, such a function can be defined, according to available data. The curve $P(\alpha)$ can be used to explain the potential of the generative mirror in describing the behaviour of the structure $S$ exploiting only the mean value and the standard deviation of the generated-by-the-model samples. Such an approach provides a way to define an interval, regardless the shape of the distribution, within which all observations $r^{C}_{i}$ would fall into with some probability $P(\alpha)$.

A distinction is introduced within the environmental parameters. The first category is the \textit{controlled variables} ($\underline{e}^{C}_{c}$) which are the variables that are used as deterministic inputs into the mirror $M$. They are quantities that are measured from the environment of the structure and whose effect on the behaviour is modelled. The second category is the \textit{uncontrolled variables} ($\underline{e}^{C}_{u}$), which include parameters that affect the structure but are either unknown or stochastic. A generative model $M^{EC}_u$ that makes the best estimate of $\underline{e}^{C}_{u}$ is needed (in the case of stochastic finite elements, as will be explained later, $M^{EC}_u$ is the stochastic process used for the random quantities of the problem). Given that the model $M$ used as a mirror is a generative model, the model's output under the context $C$, is a probability density function $P$ of the prediction $\underline{p}^{C}$ of the quantities of interest given by,
\begin{equation}
    P_{\underline{p}^{C}} = M(\underline{e}^{C}_{c}, \underline{\hat{e}}^{C}_{u}=M^{EC}_u).
    \label{eq:model_with_known_unkown}
\end{equation}
where $P_{\underline{p}^{C}}$ is the probability density function of the quantity of interest.

Furthermore, following \citep*{worden2020digital} the definition of a virtualisation is provided. Given some context $C$, a virtualisation is defined as the pair,
\begin{equation}
    V^{C} = (M_{\epsilon_1}^{C}, M_{u|\epsilon_2}^{EC})
    \label{eq:virtualisation}
\end{equation}
where $M_{\epsilon 1}^{C}$ is a model calibrated according to data from the physical structure and an established $\epsilon$-mirror for some tolerance $\epsilon_1$ within the context $C$ and $M_{u|\epsilon_2}^{EC}$ is modelling the stochastic uncontrolled variables in the context, $C$, which is also an $\epsilon$-mirror with tolerance $\epsilon_2$, that provides the best estimate for the unknown parameters. A generative model can be considered a virtualisation with clearly separated stochastic and deterministic inputs. The stochastic inputs are modelled by $M_{u|\epsilon_2}^{EC}$ and the model $M_{\epsilon 1}^{C}$ is informed from $M_{u|\epsilon_2}^{EC}$ as well as from deterministic parameters in order to generate probability distributions of the quantities of interest of the outputs; for example, displacements, natural frequencies, accelerations etc.

\begin{figure}[!htbp]
    \centering
    \includegraphics[width=0.58\textwidth]{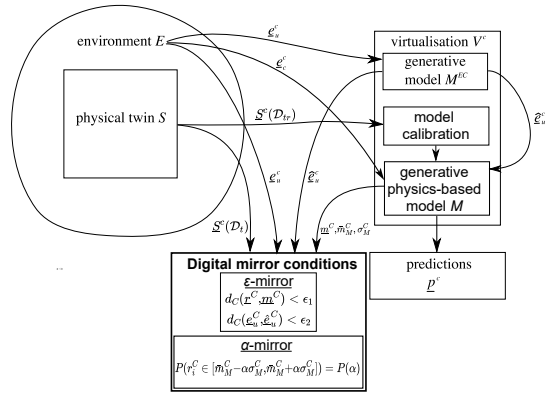}
    \caption{Schematic representation of the proposed framework for a digital mirror.}
    \label{fig:mirror_framework}
\end{figure}

The framework is schematically shown in Figure
\ref{fig:mirror_framework}. As mentioned, the uncontrolled variables $\underline{e}^{C}_{u}$ from the environment are modelled by a generative model $M^{EC}_u$. Of all the data acquired from the physical twin ($\underline{S}^{C}$), a subset is considered to be the training data, $\mathcal{D}_{tr}$, used to calibrate the digital mirror model, i.e.\ $\underline{S}^{C}(\mathcal{D}_{tr})$. The calibrated model is used to yield predictions. Since it is a generative model, some stochastic input is used, which in this case is the best estimate of the uncontrolled environment variables $\underline{\hat{e}}^{C}_{u}$. Some controlled variables of the environment  are also used as inputs $\underline{e}^{C}_{c}$. As far as the evaluation of the model $M$ as a digital mirror is concerned, using some testing data instances $\underline{r}_{C} \in \underline{S}^{c}(\mathcal{D}_{t})$ (where $\underline{S}^{c}(\mathcal{D}_{t})$ are acquired data from the physical twin and considered the testing data) and equations (\ref{eq:epsilon_mirror}) and (\ref{eq:alpha_mirror_prob}), the parameter $\epsilon_{1}$ and the curve $\alpha \to P(\alpha)$  are defined; the latter two describe the ability of the model to perform as a digital mirror. It is worth noting that for a generative model, the distance $d^{C}$ is computed between the probability density function of the predictions $P_{\underline{p}^{C}}$ and the probability density function of the recorded data of the quantity of interest $P_{\underline{r}^{C}}$. Finally, the model is used to get a probability density function of predictions $P_{\underline{p}^{C}}$ corresponding to new values of the controlled variables.

\section{Generative adversarial networks}
\label{sec:GANs}

The stochastic finite element (SFE) method is a white-box physics-based generative modelling method that can be used as a mirror of a structure. The method's performance is largely based on the knowledge one has about the physics of the problem and the finite element formulation. As an alternative, and trying to avoid unnecessary epistemic uncertainty problems, a machine learning black-box solution to the problem is proposed here.

For the purposes of using generative models as mirrors of structures, a very recently-developed neural network architecture is used here. The core algorithm is the \textit{generative adversarial network} (GAN) \citep*{goodfellow2014generative} and a variation of it, the \textit{conditional generative adversarial network} (cGAN) \citep*{mirza2014conditional}. The latter algorithm is used exactly in the same way as an SFE model is used. A deterministic input to the model is defined and the model generates distributions (or samples) of the output quantities. In the current section, the two algorithms are presented and their functionality is explained.

\subsection{Vanilla GANs}
\label{sec:vanilla_gans}

\begin{figure}[!htpb]
    \centering
    \begin{tikzpicture}[scale=0.65, every node/.style={transform shape}]
        \definecolor{blue1}{RGB}{0, 128, 255}
        \node (1) at (0.0, 0.0) [draw, line width=0.5mm, fill=orange] {Noise, $\bm{z}$};

        \node (2) at (3.4, 0.0) [draw, line width=0.5mm, fill=blue1, minimum height=1.5cm, minimum width=2cm] {Generator};
        \draw[-{Latex[width=3mm, length=4mm]}, line width=0.5mm] (1) to (2);

        \node (3) at (6.8, 1.0) [draw, line width=0.5mm, fill=orange, minimum width=3.0cm] {\shortstack{Generated\\samples $G(\bm{z})$}};
        \node (4) at (6.8, -1.0) [draw, line width=0.5mm, fill=orange, minimum width=3.0cm] {\shortstack{Real\\ samples $\bm{x}$}};
        \draw[-{Latex[width=3mm, length=4mm]}, line width=0.5mm] (4.4, 0.0) to (5.3, 1.0);

        \node (5) at (11.2, 0.0) [draw, line width=0.5mm, fill=blue1, minimum height=1.5cm, minimum width=3.0cm] {Discriminator};

        \draw[-{Latex[width=3mm, length=4mm]}, line width=0.5mm] (8.3, 1.0) to (9.7, 0.0);
        \draw[-{Latex[width=3mm, length=4mm]}, line width=0.5mm] (8.3, -1.0) to (9.7, 0.0);

        \node (6) at (15.8, 0.0) [draw, line width=0.5mm, fill=orange, minimum height=1.5cm, minimum width=3.0cm] {Probability $D(G(\bm{z}))$};
        \draw[-{Latex[width=3mm, length=4mm]}, line width=0.5mm] (5) to (6);

    \end{tikzpicture}
    \caption{Vanilla GAN layout.}
    \label{fig:gan_layout}
\end{figure}
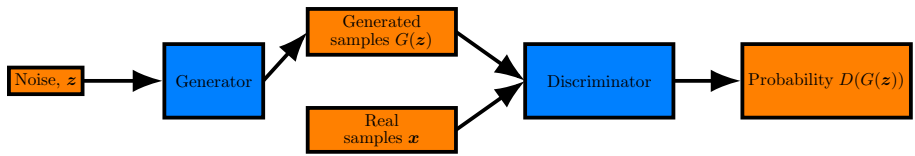

The traditional scheme followed in machine learning is the training of a model to perform classification \citep*{Bishop} or regression \citep*{specht1991general}. To extend this to images, convolutional neural networks were developed \citep*{krizhevsky2012imagenet}, yielding superior performance in the two mentioned tasks. Recently, a new type of neural network has emerged, the \textit{generative adversarial network} (GAN) \citep*{goodfellow2014generative}. The goal of this new scheme was initially to generate images that resemble reality. This task is achieved via the use of \textit{two} neural networks. The first one is termed the \textit{generator} and produces ``fake'' images given a latent noise vector. The second network is the \textit{discriminator}, which tries to identify whether an image, fed to it as an input, is fake (generated by the generator) or real (coming from the available dataset). By training, both of these networks improve towards their objectives and finally, the generator, provided with some latent vector, can generate images that appear to be real. More intuitively, this means that the generator maps a latent vector distribution into a distribution or a manifold of the real data. The layout of the basic (vanilla) GAN can be seen in Figure \ref{fig:gan_layout}.

The generator is commonly a multi-layer perceptron (MLP) \citep*{Bishop}, that takes as input a latent noise vector $\textbf{z}$ coming from a probability distribution $p_z(\textbf{z})$ and maps it into a vector (or an image) $G(\textbf{z})$ of dimension equal to the dimension of the training samples. The discriminator is another MLP that takes as inputs, vectors (or images) $\textbf{x}$, and outputs the probability of the sample being real, $P(\textbf{x}=real) = D(\textbf{x})$. The training of the discriminator is carried out by maximising the probability that it assigns the correct label (``real'' or ``fake'') to the samples. At the same time, the training of the generator, $G$, is accomplished by trying to minimise the probability that the discriminator classifies the generated samples as fake, i.e. minimisation of $\log(1 - D(G(z)))$. Following from \citep*{goodfellow2014generative}, the objective function $\mathcal{L}$ can be interpreted as a two-player game explained by,

\begin{equation} \label{eq:obj_fun}
    \min\limits_{G}\max\limits_{D}\mathcal{L}(D,G)=\mathbb{E}_{\textbf{x} \sim p_{data}(\textbf{x})}[\log D(\textbf{x})] + \mathbb{E}_{\textbf{z} \sim p_{z}(\textbf{z})}[\log 1 - D(G(\textbf{z})))]
\end{equation}

Training of such a network is performed in two steps per epoch. During the first step, random samples are created by the generator and concatenated with a batch of real samples from the dataset. The resulting training batch is used to train the discriminator for one epoch by back-propagating the error of the output. The target label for the real samples is 1 and for the generated ones is 0. The first term of the right-hand side of equation (\ref{eq:obj_fun}) is set in this step as the objective function and its \textbf{maximisation} is attempted. Consequently, the two networks are clipped together as in Figure \ref{fig:gan_layout}, and random samples of the latent vector are generated in order to create random-generated samples. These samples are fed into the whole GAN assembly and the target outputs are labels of 1. The weights of the discriminator's connections are considered as constants during the second training phase and the error is back-propagated in order to train only the generator. This time, the objective function is composed exclusively of the second term of the right-hand side of equation (\ref{eq:obj_fun}) and its \textbf{minimisation} is sought. Following this training scheme, during the first step the discriminator learns to distinguish between real and generated images and the generator to generate images that the discriminator classifies as real and (as shown in \citep*{goodfellow2014generative}) to have probability distribution similar to the real data.

The most straightforward application of GANs is to generate artificial data to augment a dataset. Training neural networks is highly dependent on the size of the available dataset. The rule-of-thumb for training neural networks that generalise well \citep*{tarassenko1998guide}, specifies that for each trainable weight of the neural network, 10 training samples are needed. This statement probably does not stand for GANs, as they are also trained using random noise and generated samples that do not come from the available dataset. Acquiring engineering data is difficult and some times even expensive. Labelled images are hard to obtain and their manual labelling costs both time and money. In cases of image datasets, augmentation can also be achieved by rotation of the pictures or colour change etc. In SHM for example, where acquiring sufficient data is vital in order to efficiently monitor the health state of structures, the securing of data from structures in different damage cases or under different environmental conditions can be very expensive or even impossible; the samples are usually limited and augmentation is not trivial. Especially for deep networks, and even more for deep convolutional neural networks, where the number of trainable parameters is huge, augmentation of available dataset size could yield an efficient way to increase the generalisation performance of models \citep*{frid2018gan}.

\subsection{Conditional generative adversarial networks (cGAN)}
\label{sec:cGAN}
\begin{figure}[!htbp]
    \centering
    \begin{tikzpicture}[scale=0.65, every node/.style={transform shape}]
        \definecolor{blue1}{RGB}{0, 128, 255}
        \node (1) at (0.0, 1.0) [draw, line width=0.5mm, fill=orange] {Noise, $\bm{z}$};

        \node (A) at (0.0, -1.0) [draw, line width=0.5mm, fill=orange] {Code, $\bm{c}$};

        \node (2) at (3.4, 0.0) [draw, line width=0.5mm, fill=blue1, minimum height=1.5cm, minimum width=2cm] {Generator};
        \draw[-{Latex[width=3mm, length=4mm]}, line width=0.5mm] (1) to (2);

        \draw[-{Latex[width=3mm, length=4mm]}, line width=0.5mm] (A) to (2);

        \node (3) at (6.8, 2.0) [draw, line width=0.5mm, fill=orange, minimum width=3.0cm] {\shortstack{Generated\\samples $G(\bm{z, c})$}};
        \node (4) at (6.8, -2.0) [draw, line width=0.5mm, fill=orange, minimum width=3.0cm] {Real samples $\bm{x,c}$};

        \node () at (6.8, 0.0) [draw, line width=0.5mm, fill=orange, minimum width=3.0cm] {Code, $\bm{c}$};;

        \draw[-{Latex[width=3mm, length=4mm]}, line width=0.5mm] (4.4, 0.0) to (5.3, 2.0);

        \node (5) at (11.2, 0.0) [draw, line width=0.5mm, fill=blue1, minimum height=1.5cm, minimum width=3.0cm] {Discriminator};

        \draw[-{Latex[width=3mm, length=4mm]}, line width=0.5mm] (8.3, 2.0) to (9.7, 0.0);
        \draw[-{Latex[width=3mm, length=4mm]}, line width=0.5mm] (8.4, -2.0) to (9.7, 0.0);

        \draw[-{Latex[width=3mm, length=4mm]}, line width=0.5mm] (8.3, 0.0) to (9.7, 0.0);

        \node (6) at (15.8, 0.0) [draw, line width=0.5mm, fill=orange, minimum height=1.5cm, minimum width=3.0cm] {Probability $D(G(\bm{z, c}))$};
        \draw[-{Latex[width=3mm, length=4mm]}, line width=0.5mm] (5) to (6);

    \end{tikzpicture}
    \caption{Layout of a cGAN.}
    \label{fig:cgan_layout}
\end{figure}
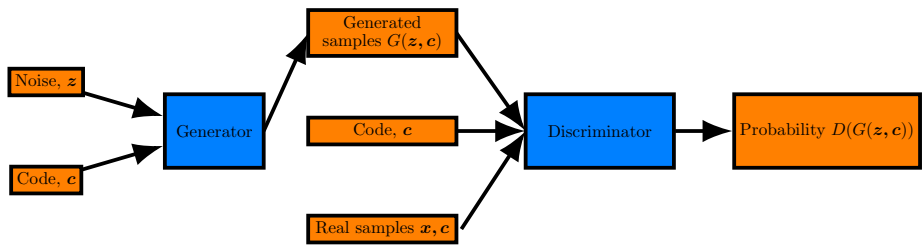

Conditional generative adversarial networks are an attempt to control the output of the generator by conditioning on some variables. In contrast to the traditional GAN layout (Figure \ref{fig:gan_layout}), where the product of the generator is completely controlled by random noise $\textbf{z}$, the output of the generator is here partially controlled by some vector $\textbf{c}$; thus providing a way of learning distribution and manifold transformations parametrised on the code. This code may be a continuous variable or a discrete one. Since the output of the generator depends on the code, the output of the discriminator should also depend on it and the discriminator should also have it as an input. Therefore, the layout of the cGAN is as shown in Figure \ref{fig:cgan_layout}.

Training such an assembly of networks, the discriminator learns that, for each different value of $\textbf{c}$, a different acceptance or rejection boundary is defined in the sample space. Since the decision boundary of the discriminator varies according to the code, the generator also learns to vary its outputs according to $\textbf{c}$ to ``fool'' the discriminator. In cases of discrete or categorical variables, the result is that the generator learns to generate samples belonging to different categories. In \citep*{mirza2014conditional}, an illustration of this result is presented for the MNIST dataset; a collection of hand-drawn digits of numbers. By defining ten binary categorical variables as the code, the generator is able to create sample images in predefined classes, controlled by the code.

The use of continuous variables yields more convenient results for physics modelling. A continuous variable would force the mold (the boundary around the manifold of the data) created by the discriminator, to be gradually transformed as a function of the values of the code. The decision boundaries of the discriminator then force the generator to create samples within the region they define. Consequently the geometry of the generated manifolds is conditioned on the code vector $\textbf{c}$. Using this training scheme, the generator has learnt the transformation of the manifold and the distribution of the generated points, as a function of the code.

The algorithm may be exploited in order to generate artificial data as a function of some code vector but, in the current work, it shall also be exploited in order to learn the transformations of the aforementioned manifolds and distributions, as functions of the code. The output of interest of a generative model is the distribution of some quantity of interest, making the cGAN algorithm a suitable candidate to serve as such a model. The code plays the role of the variable that affects the output distribution and the cGAN is called to learn from the data how the distribution transforms according to the values of the code. The code in a structure modelling context can represent the loading and the environmental conditions of some structure, while the output distributions are the probability distributions of the quantity of interest, i.e. displacement, acceleration, natural frequency etc. A major advantage of using such a model as a digital mirror is that there is no need for modelling the uncontrolled variables from the environment ($\underline{e}^{C}_{u}$), i.e. the generative model $M^{EC}$ step in Figure \ref{fig:mirror_framework} is bypassed. The effect of these variables is taken into consideration via the noise variables of the cGAN model.

\section{Stochastic finite element models as mirrors}
\label{sec:SFEM_mirror}

A stochastic finite element model updated according to data acquired from a structure could be considered as a mirror under the criteria discussed earlier. It could take into account experimental noise and aleatory uncertainty that might exist in a structural problem. The model can also be continuously updated according to newly-acquired data, in order to take into account random events and environmental conditions. In every case, model parameters have to be chosen in order for the model to fit the acquired data.

The parameters of the model that will most probably need calibration, are the parameters describing the stochastic fields of the problem. Some assumptions can be made about the fields; the first might be that the field is stationary and \textit{Gaussian} or \textit{lognormal}. A subsequent assumption might then be about the form of the autocorrelation function. A quite common type of autocorrelation function in SFEM problems is the squared-exponential function, such as $\rho(x, x') = \exp{-(\frac{x - x'}{l})^{2}}$, where $x$ and $x'$ are the points in space and $l$ is a parameter called the \textit{correlation length}.

Having decided on the type of the field and the autocorrelation function, the hyperparameters remaining to be defined are the mean and variance values of the random field (or the mean and variance functions if the field is not stationary). Furthermore, for the aforementioned autocorrelation equation, a third hyperparameter is the correlation length $l$. Fitting can be done in many ways; the most straightforward is an exhaustive search over some set of candidate parameters for the values that yield the best results.

However, a way to evaluate the performance of such generative models is needed. Since it is a generative model and its output is a distribution, a distance metric between the generated and the real distributions should be used as a performance criterion. The \textit{Kullback–Leibler divergence} (KL divergence) \citep*{kullback1997information} is a quantity that measures the ``distance'' between two distributions; it can therefore be used as such a criterion. Regarding mirror terminology, this is the distance $\epsilon$ used to define an $\epsilon$-mirror. The KL divergence between two distributions $P$ and $Q$ is given by,
\begin{equation}
\label{eq:KL_divergence_discr}
    D_{KL}(P||Q) = \sum_{n=1}^{n_{val}} P(x) \log(\frac{P(x)}{Q(x)})
\end{equation}
where $n_{val}$ is the number of available datasets to compute the KL divergence between the predicted and the real distributions (Note that this is the discretised version of the metric.)

Stochastic FEM models take into account uncertainties in the parameters of the structure, but can have a deterministic input. Thus, the output distribution is a function of the input. A model, which might be considered as a mirror of a structure, should be able to perform under different inputs. A simple case to consider is that of a deterministic load input to the model. In this case, the model shall be evaluated for different values of the load and the best one shall be the one with the best average performance amongst all the cases of deterministic loads. Other deterministic inputs might also be the temperature of the environment, seismic accelerations, humidity, etc.

Under the framework of mirrors, the load or any other deterministic input shall be the controlled variable $\underline{e}^{C}_{c}$. Any uncertainties, such as Young's modulus, Poisson ratio etc. and unknown environmental parameters, are included in the uncontrolled variables $\underline{e}^{C}_{u}$. The generative model that estimates the uncontrolled parameters ($M^{EC}_u$) shall be the stochastic process described by the Karhunen-Loeve expansion. The SFE model will be the generative model that will provide the probability density functions of the quantities of interest.

\subsection{Definition of simulation dataset}
\label{sec:GA_application}

In order to test the algorithm, data should be available from some structure of interest. Such data may refer to different deterministic inputs such as load, temperature etc. For every available value of the deterministic input, a set of samples should be available, from which the distribution of the quantity of interest is extracted. The model should perform well in generating distributions close to the ones in the dataset for different values of the inputs. Since SFE models are \textit{white-box} physics-based models, fitting them to data for a set of input values increases the belief that the model will generalise. This assumption is only true if the physical formulation of the model corresponds to the real physical mechanism, i.e. if epistemic uncertainty is not present.

In order to define the required dataset, a simulated structure is considered here in order to generate data. The structure is a simple cantilever with Young's modulus defined as a stochastic field, similar to the one in Figure \ref{fig:cantilever_varying_E} in the Appendix. In real structures, such cases may be observed in a bridge for example, when many heat sources affect the temperature of the structure. This situation would result in fluctuations of the stiffness of the structure within its volume. The final field is of course a stochastic field with some correlation function.

\begin{figure}[!htbp]
    \centering
    \includegraphics[width=0.48\textwidth]{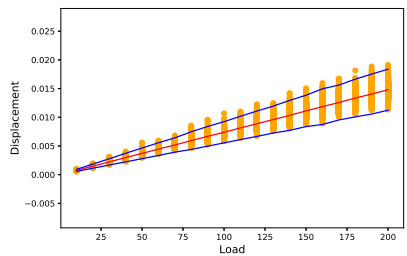}
    \caption{Samples of tip displacements (orange points), their mean values (red line) and $\pm$3 standard deviations (blue line).}
    \label{fig:MC_linear_samples}
\end{figure}

The model cantilever here has length equal to $5$ (length units), a rectangular cross section with height equal to $0.4$ and width equal to $0.1$. The stochastic field was chosen to be a stationary Gaussian stochastic field with mean value equal to $2 \times 10^9$, and standard deviation equal to $0.2 \times 2 \times 10^9$ (pressure units). The correlation length was $3.0$. The procedure described in the Appendix was performed, and equation (\ref{eq:stiffness_realisations_expanded_trunc}) with order of the expansion $m=2$ was used to generate realisations of the stiffness matrix of the structure. As an input, a deterministic distributed load along the cantilever with varying values was considered. The values of the load $f$ were $10, 20, 30..., 200$ force units / length units. The dataset was split into three datasets, one for training, one for validation and one for testing. Samples of the corresponding tip displacements of the cantilever are shown in Figure \ref{fig:MC_linear_samples}. For each load value, $1000$ samples were generated. The results comply with the linearity of the problem, since the mean value (red line) is almost linearly increasing as the load increases.

\subsection{Model calibration (updating) for a simulated structure}
\label{sec:SFEM_fitting_results}

\begin{figure}[!htbp]
    \centering
    \begin{subfigure}[b]{0.49\linewidth}
    \centering
    \includegraphics[width=0.99\linewidth]{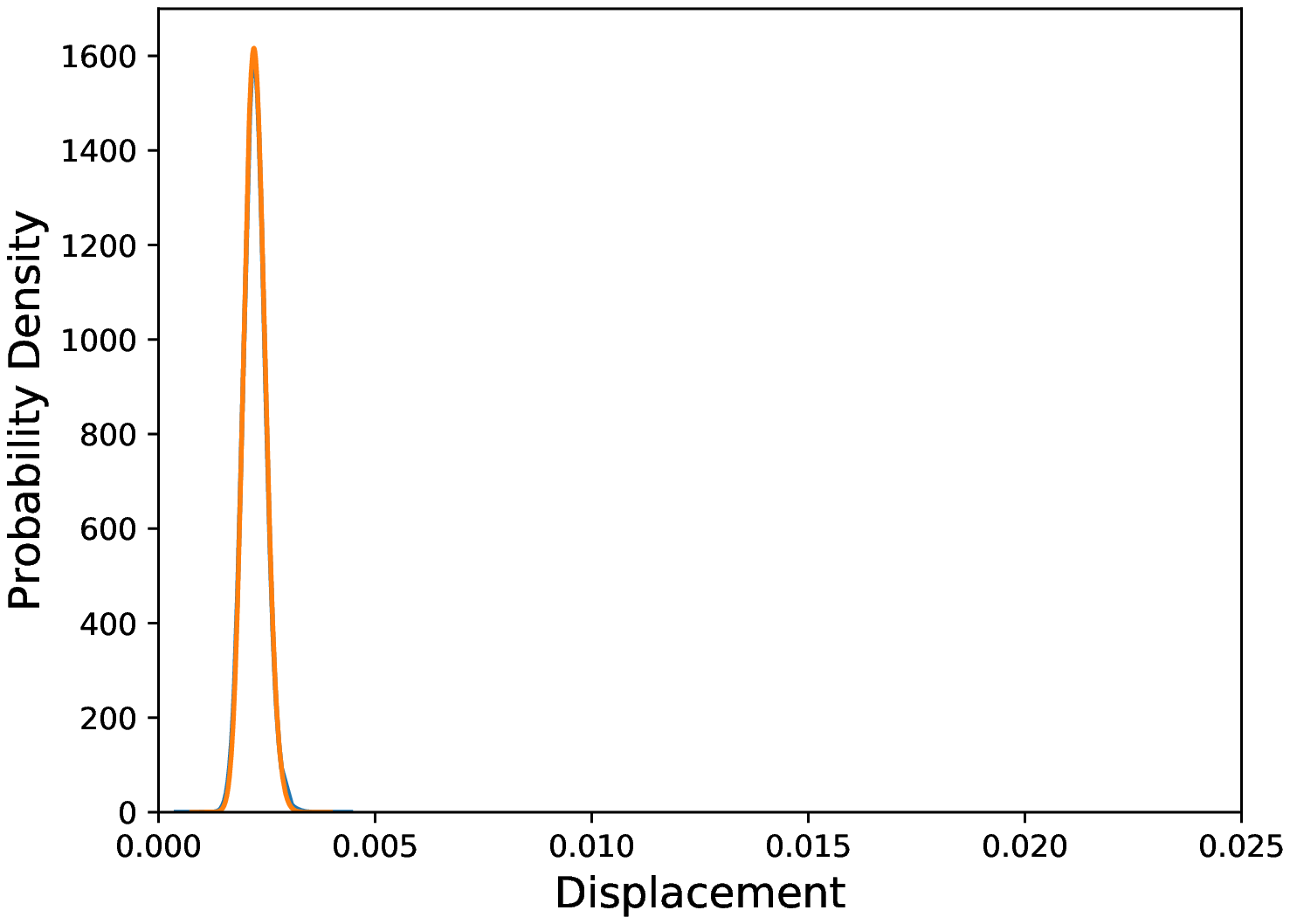}
    \label{fig:sub_a1}
    \caption{}
    \end{subfigure}
    \begin{subfigure}[b]{0.49\linewidth}
    \centering
    \includegraphics[width=0.99\linewidth]{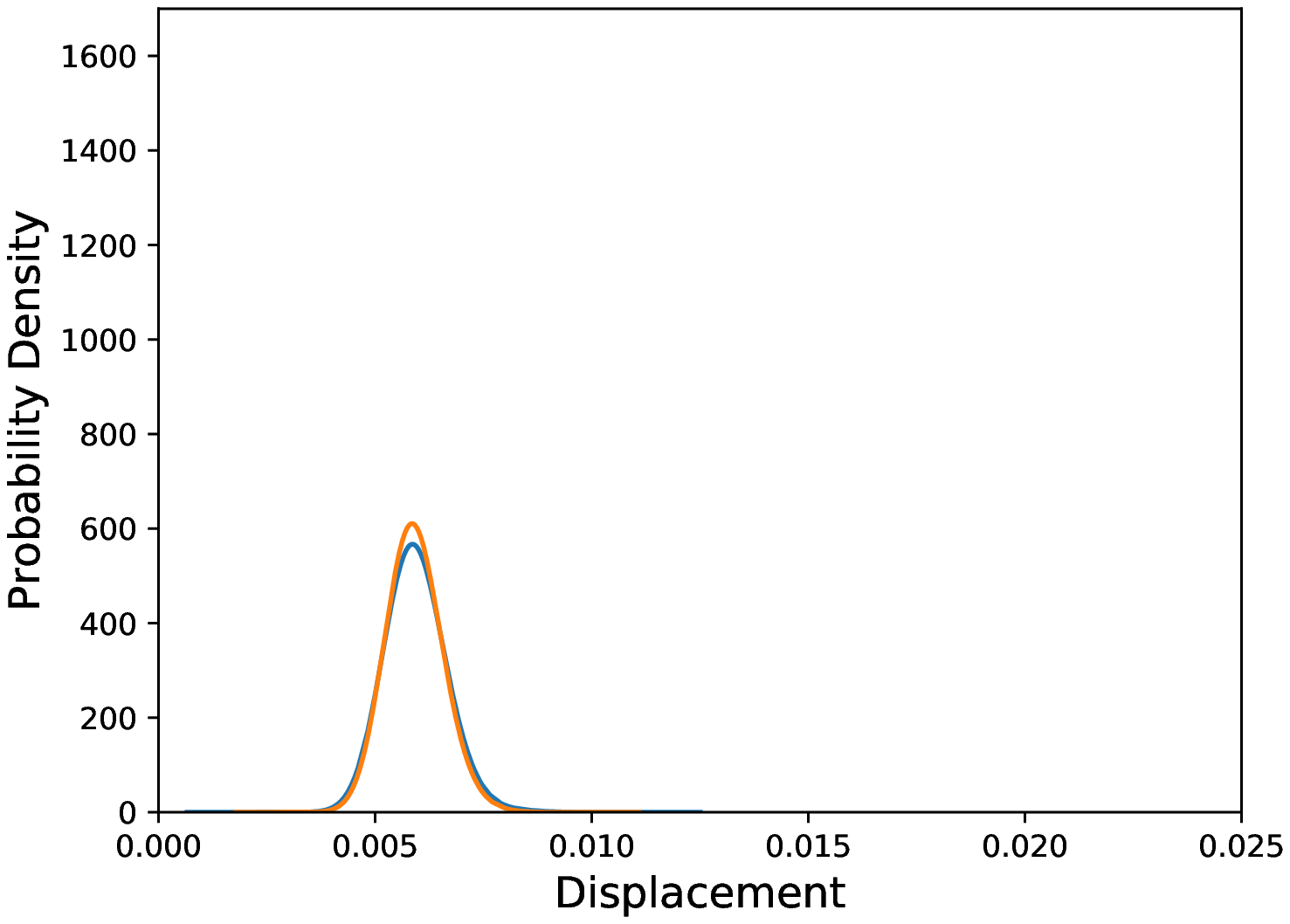}
    \caption{}
    \label{fig:sub_b1}
    \end{subfigure}
    \begin{subfigure}[b]{0.49\linewidth}
    \centering
    \includegraphics[width=0.99\linewidth]{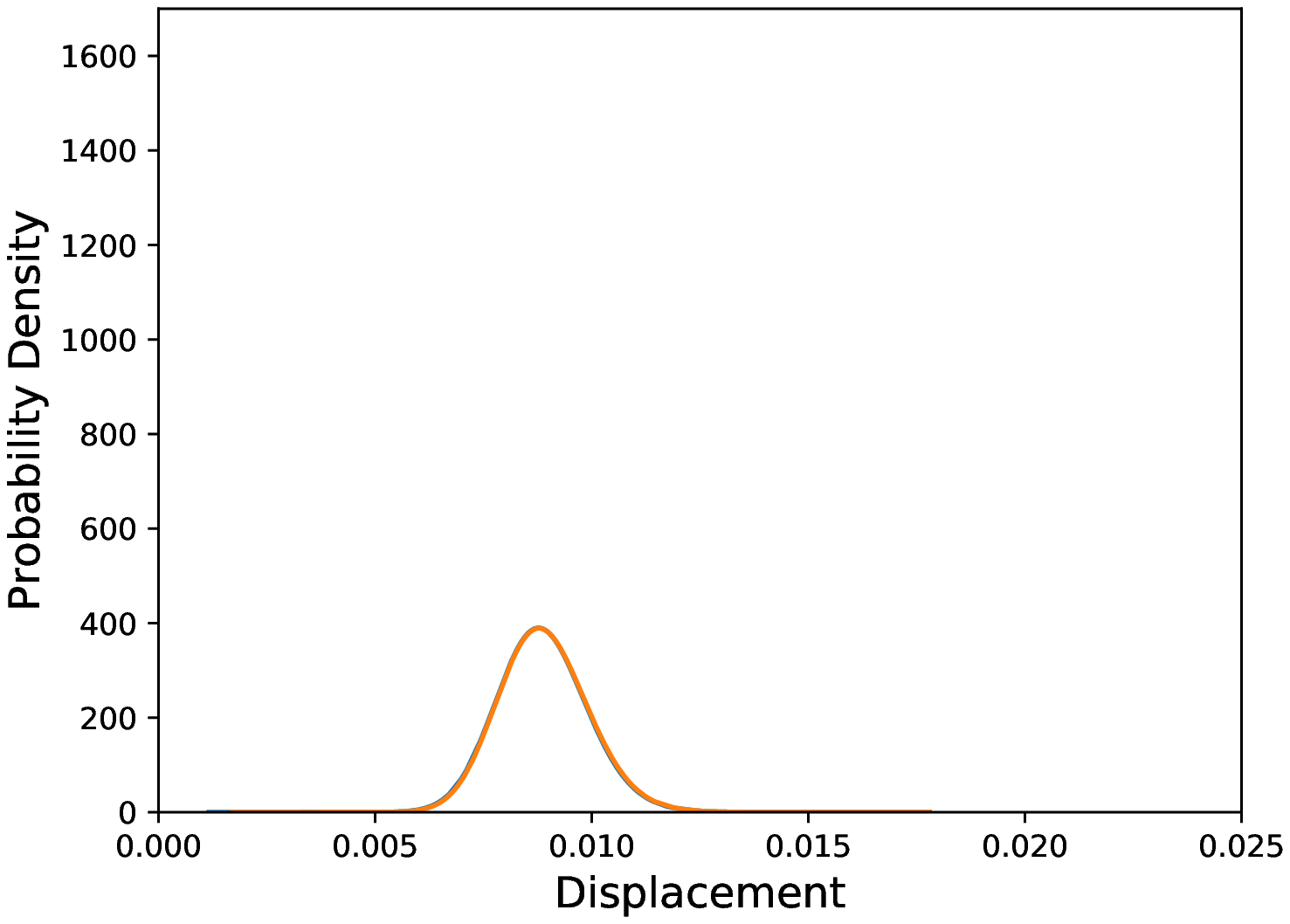}
    \label{fig:sub_c1}
    \caption{}
    \end{subfigure}
    \begin{subfigure}[b]{0.49\linewidth}
    \centering
    \includegraphics[width=0.99\linewidth]{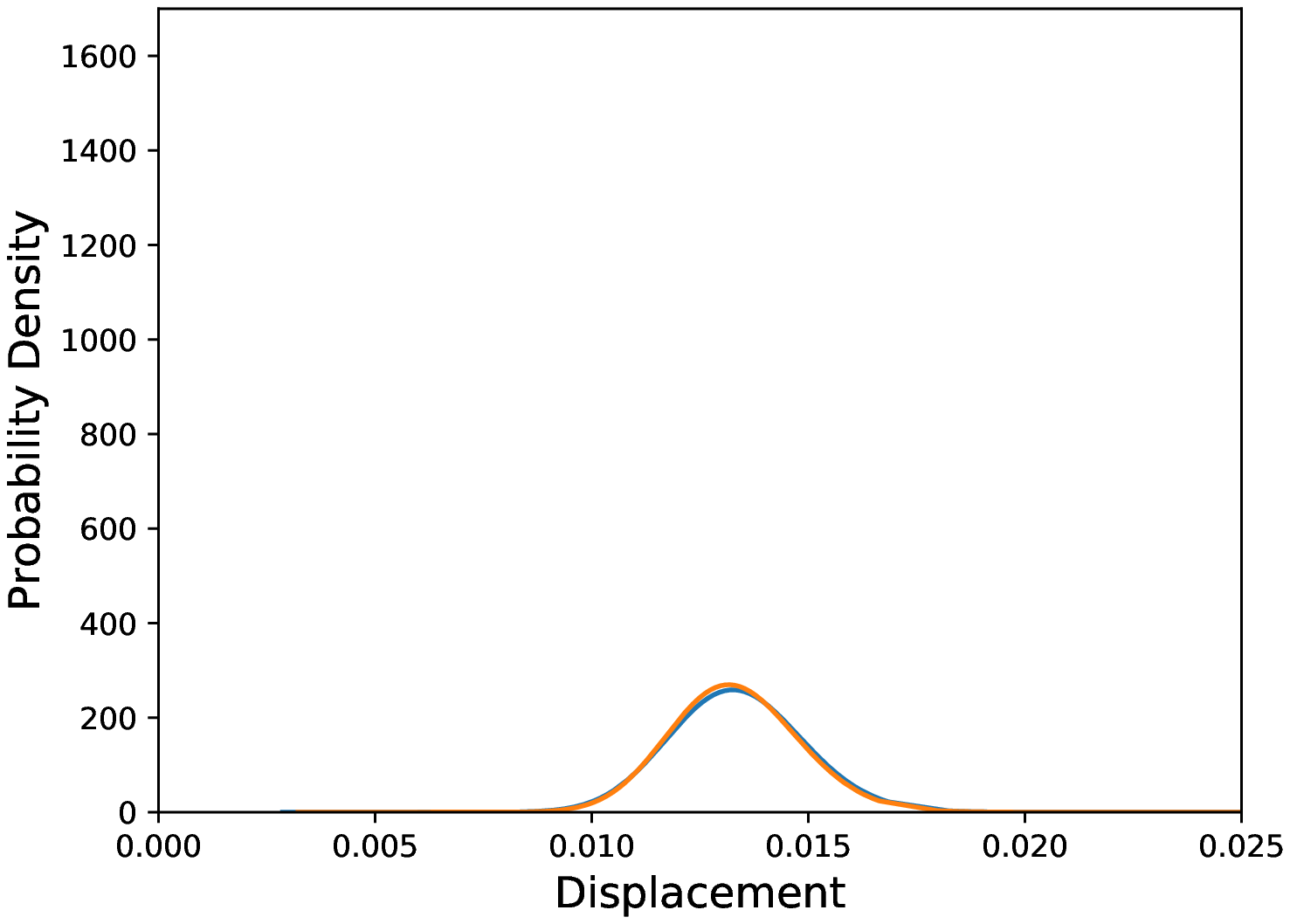}
    \label{fig:sub_d1}
    \caption{}
    \end{subfigure}
    \caption{Distributions of tip displacements corresponding to Monte Carlo samples (orange) and SFEM generated samples (blue) and different load cases, (a) 30 load units, (b) 80 load units, (c) 120 load units and (d) 180 load units.}
    \label{fig:distributions_comparison}
\end{figure}

The calibration procedure followed is simply an exhaustive search in asubset of the three-dimensional parameter space. The search is performed over some logical range of values for the parameters. The range could be defined using engineering insight of the problem. The three parameters are the mean Young's modulus ($\mu_{E}$), the Young's modulus standard deviation ($\sigma_{E}$) and the correlation length ($l_{corr}$). The model was calibrated using a subset (loads $\{10, 40, 70, 100, 130, 160, 200\}$) of the total set of loadcases of the dataset. It is, however, tested in all the cases and some distribution comparisons are shown in Figure \ref{fig:distributions_comparison} for selected loadcases. The overall average KL divergence is $0.0028$. This value means that the model almost perfectly explains the behaviour of the structure and is able to make accurate predictions regarding the distribution of the tip displacements, as evidence, the visual comparisons in Figure \ref{fig:distributions_comparison} are presented.

Of course, the model performs so well because the formulation of the finite elements in the SFE model was exactly the same as the one used to generate the dataset. In real structure situations epistemic uncertainty may be present. In addition, more uncertain parameters may affect the structure, such as humidity. In the latter case, fitting a model with uncertainty imposed in the Young's Modulus, could perform well enough incorporating the uncertainty of the unknown parameters, as uncertainty existing in the stochastic field of the model. In such cases, however, the values of the model parameters from the fitting, will not resemble the real values of the random field of the stochastic quantity.

Taking into account the definitions of $\epsilon$-mirror and $\alpha$-mirror, the SFE model may serve as both types. For the case of an $\epsilon$-mirror, the distance metric $\epsilon$ to be used is the KL divergence of the real data from the SFE model outputs. Considering $\epsilon$ equal to the maximum KL divergence, of the available datasets, between the simulated and the predicted by the SFE model distributions. The maximum KL divergence was $0.0029$ and so, the SFE model can be an $\epsilon$-mirror with $\epsilon \geq 0.0029$ considering engineering judgement or some safety factor.

\begin{figure}[!htbp]
    \centering
    \includegraphics[scale=0.5]{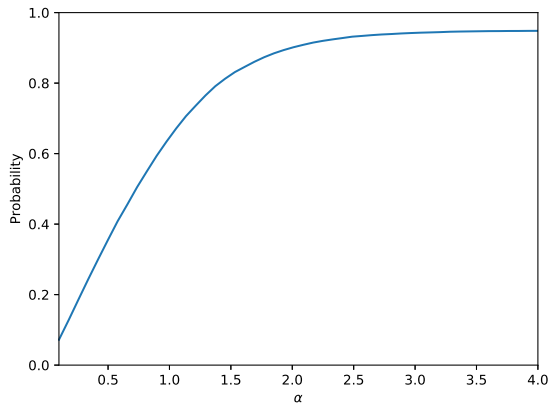}
    \caption{Probability defined in equation (\ref{eq:alpha_mirror_prob}) as a function of the parameter $\alpha$ for the SFE model applied on the linear cantilever case study.}
    \label{fig:alpha_probs_SFEM_linear}
\end{figure}

Regarding the use of the model as an $\alpha$-mirror, the curve $\alpha - p(\alpha)$ from equation (\ref{eq:alpha_mirror_prob}) should be defined. Using only the available data, and various values for the $\alpha$ parameter, Figure \ref{fig:alpha_probs_SFEM_linear} shows the probability $p$ as a function of $\alpha$. For $\alpha = 2$, $90\%$ of the observations fall into the interval defined by equation ($\ref{eq:alpha_mirror_prob}$) while for $\alpha = 3$, $95\%$ of the observations are in the aforementioned interval. A straightforward evaluation of the quality of the model according to this curve is not available. Certainly, as close as this curve is to the corresponding curve of the real data, the better it is. However, this is equivalent (and maybe a more loose evaluation criterion) to the requirement that the distribution of the real and the generated data are similar. The curve represents the potential of the model to explain the structural behaviour and can definitely be used as a tool to perform a probabilistic cost-benefit analysis of the performance of the structure, according to the predictions. Moreover, if one is interested only in defining an interval of the potential quantity of interest, under some environmental conditions, the model should, for some value of $\alpha$, include a sufficient percentage of the observed outcomes.

\section{cGAN as mirrors of a structure}
\label{sec:cGAN_DT}

Since a cGAN is able to generate manifolds of data as a function of a code vector, it fits the desired functionality of a mirror, as described in the current work. Moreover, as stated in \citep*{goodfellow2014generative}, the generator learns, apart from generating data that look real, to generate data close to the distribution of the original data. The same results are expected herein by the use of a cGAN. Two examples are presented below; one refers to the same problem that an SFEM model was used for, in the previous section, and the second is the same problem but with material nonlinearities. Although SFE approaches to nonlinear problems exist \citep*{stefanou2009stochastic}, knowledge about the physics and the source of the nonlinearity is required for their application. The current work is focused on illustrating the convenience of the machine learning approach and its applicability without any knowledge about the nature of the problem (linear or nonlinear) or the source of the uncertainty. Therefore a nonlinear SFE approach is not considered. Nevertheless, as illustrated in the application of the SFE model on the linear structure, if all the underlying physics of the problem matched the nonlinear finite element formulation (this should include the nature of any noise processes), it would outperform any black-box approach, since there would be no epistemic uncertainty. The cGAN method is a completely data-driven method and it is expected to be able to perform regardless of the linearity or otherwise of the underlying problem.

The cGAN also fits the defined context about controlled and uncontrolled environmental parameters, $\underline{e}^{C}_{c}$ and $\underline{e}^{C}_{u}$. There is a direct relationship between the cGAN code and the controlled variables and the noise vector and the uncontrolled variables. Defining such a separation could be crucial when one wants to quantify uncertainty. According to the way the cGAN works, the separation of known and unknown parameters and their effect on the predicted distribution is clearly given by the separation of the input vector into the noise and the code. Continuing, two applications of the cGAN are presented showing the versatility of the algorithm and its ability to perform both in linear and nonlinear structural problems.

\subsection{Application of the cGAN in a linear problem}
\label{sec:cGAN_linear}

Using the same dataset as before, a cGAN was trained. The training procedure followed was a standard neural network cross-validation training procedure. The dataset was split into three subsets: training, validation and testing. The split is performed in order to train according to the first dataset, select as the best model the one that performs best in the validation dataset and confirm that it is able to perform well on data that it has never ``seen'' before, i.e. the testing dataset. The split was made according to the codes/loads. Each load belonged to only one of the three datasets. More specifically, the samples having loads $\{10, 40, 70, 100, 130, 160, 200\}$ were the training dataset and the ones with loads $\{20, 50, 80, 110, 140, 170, 190\}$ and $\{30, 60, 90, 120, 150, 180\}$ were the validation and testing sets respectively.

Both the generator and discriminator are three-layered neural networks here; each has an input layer, a hidden layer and an output layer. The activation function was chosen to be a hyperbolic tangent function, except for the activation of the output layer of the discriminator, which is a sigmoid function in order to map to a probability in the interval $[0, 1]$. The noise vector was ten-dimensional. Different sizes were tested incrementally regarding the noise vector. It was noted that as the size increased, the performance was increased, as was the convergence speed towards the Nash equilibrium. Therefore, the size chosen here was a ten-dimensional noise vector, since increasing it further did not yield notably better results. The code vector was one-dimensional, since the control variable is only the load. Finally, using different hidden layer sizes in the set $[10, 20, 30... 3000]$ and selecting the model with the lowest KL divergence in the validation set, the sizes of the hidden layers that yielded good results were 200 nodes for both networks. Using the same size of hidden layer in both the generator and the discriminator, might conceal a physical meaning, since the generator decodes the noise into the real feature space and the discriminator maps the feature space into some latent code (in its hidden layer) in order to distinguish real and fake samples.

The quantity to be minimised, is the KL divergence between the generated and the acquired dataset distributions. However, the value of the loss function during training does not directly represent this quantity. The KL divergence is used as a model selection criterion and it is calculated between the generated and acquired distributions for the validation dataset every $100$ training epochs. At the end of training, the cGAN instance that had the lowest average KL divergence among the codes of the validation dataset, is selected as the most accurate model and is tested on the testing dataset. In order to define distributions on both the database samples and the cGAN generated ones, kernel density estimates \citep*{Epanechnikov1969} are fitted in both cases. The kernel used in the current work is a Gaussian kernel. Throughout the paper, every KL divergence is calculated for distributions of neural network outputs. These outputs are scaled to the interval $[-1, 1]$ and the bandwidth parameter used to the fitted kernel distributions was in all cases considered equal to $0.1$. That is considered an appropriate value, since the range of the outputs is equal to $2$; therefore, a bandwidth value equal to $\frac{1}{20}$ of the range yields meaningful distributions about the quantities of interest. This could be another training hyperparameter whose optimisation might be the objective of the cross validation procedure \citep*{silverman1981using}, but was not in the current work.

\begin{figure}[!htbp]
    \centering
    \begin{subfigure}[b]{0.49\linewidth}
    \centering
    \includegraphics[width=0.99\linewidth]{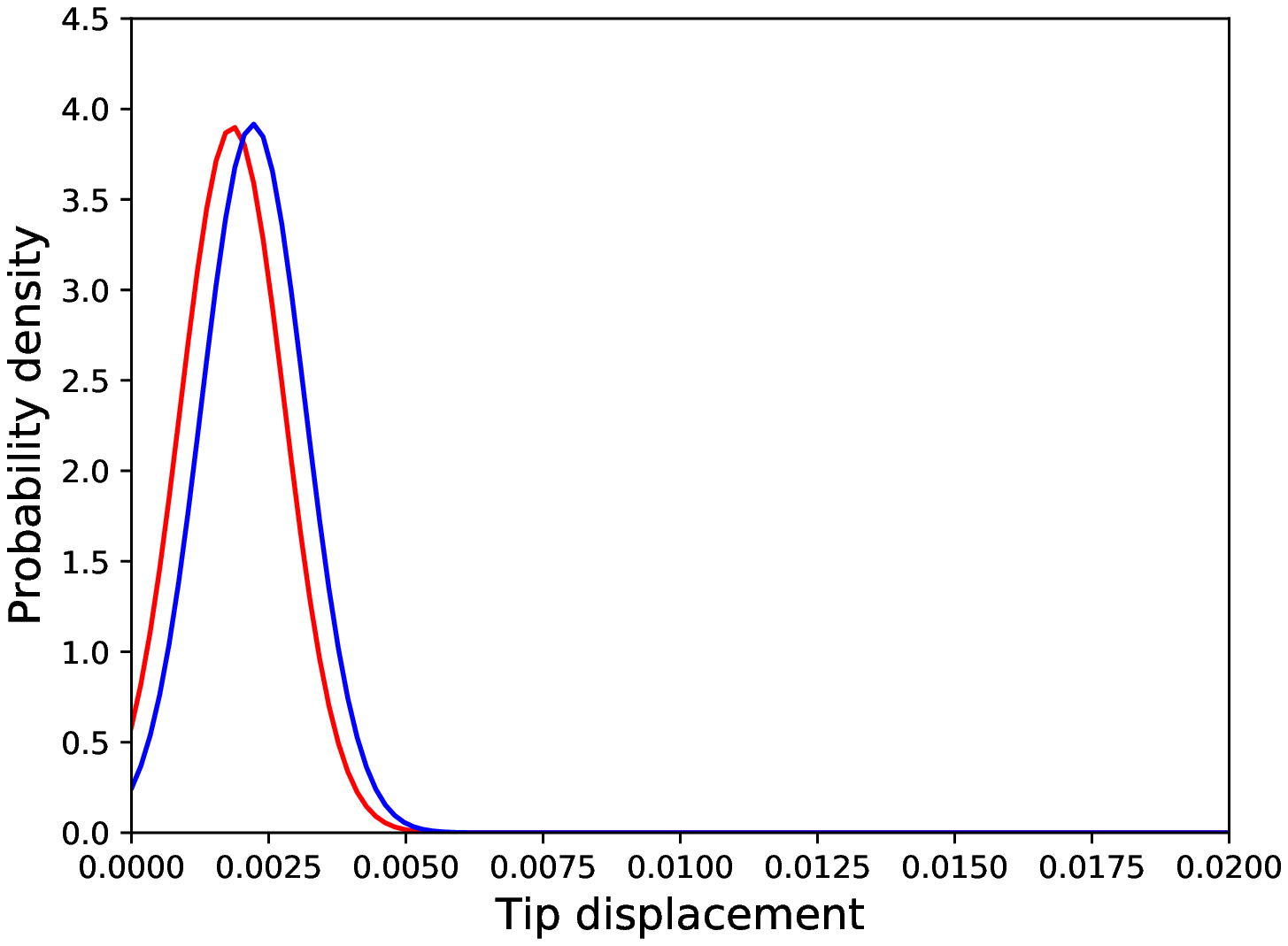}
    \caption{}
    \label{fig:sub_a2}
    \end{subfigure}
    \begin{subfigure}[b]{0.49\linewidth}
    \centering
    \includegraphics[width=0.99\linewidth]{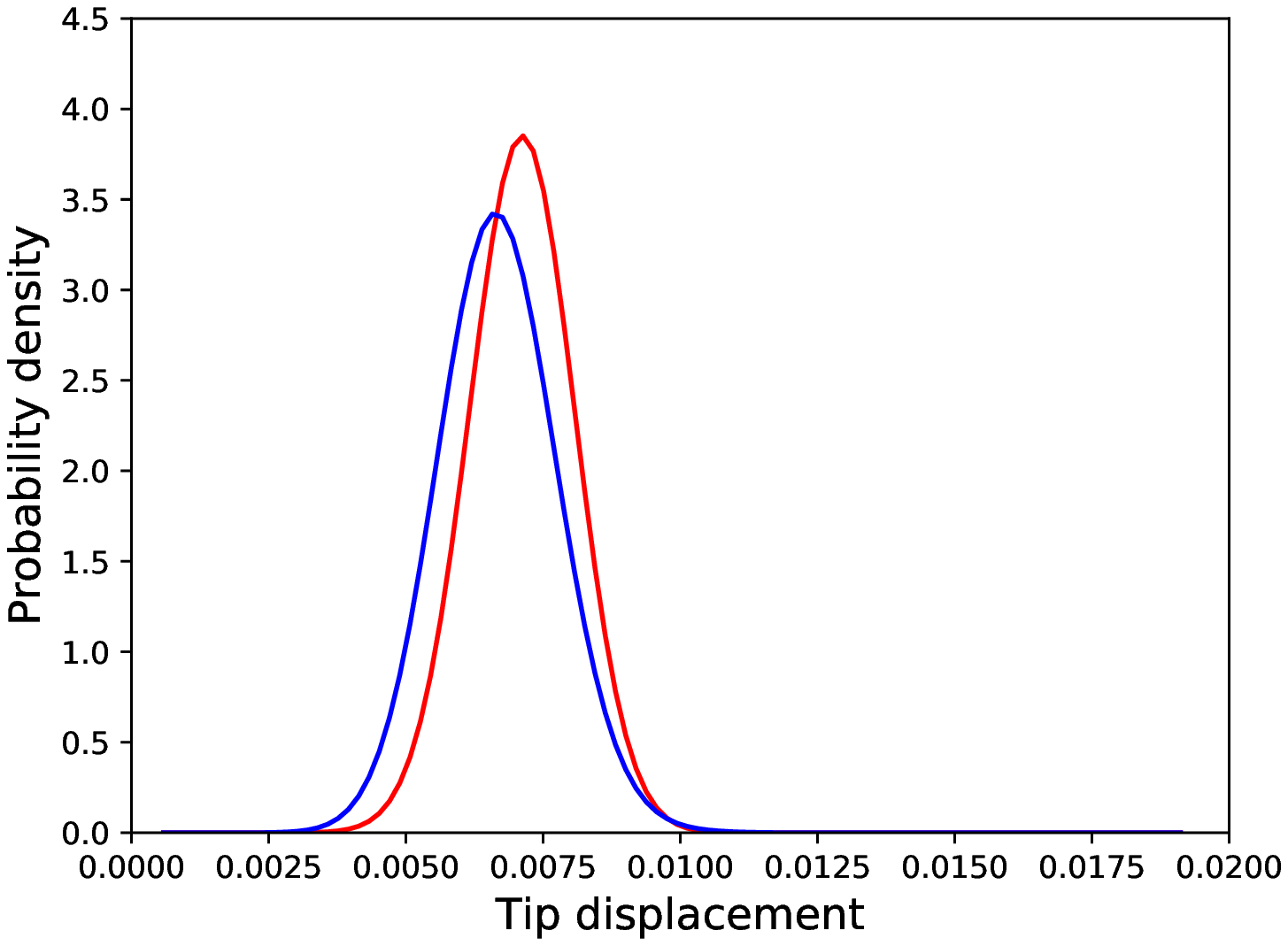}
    \caption{}
    \label{fig:sub_b2}
    \end{subfigure}
    \begin{subfigure}[b]{0.49\linewidth}
    \centering
    \includegraphics[width=0.99\linewidth]{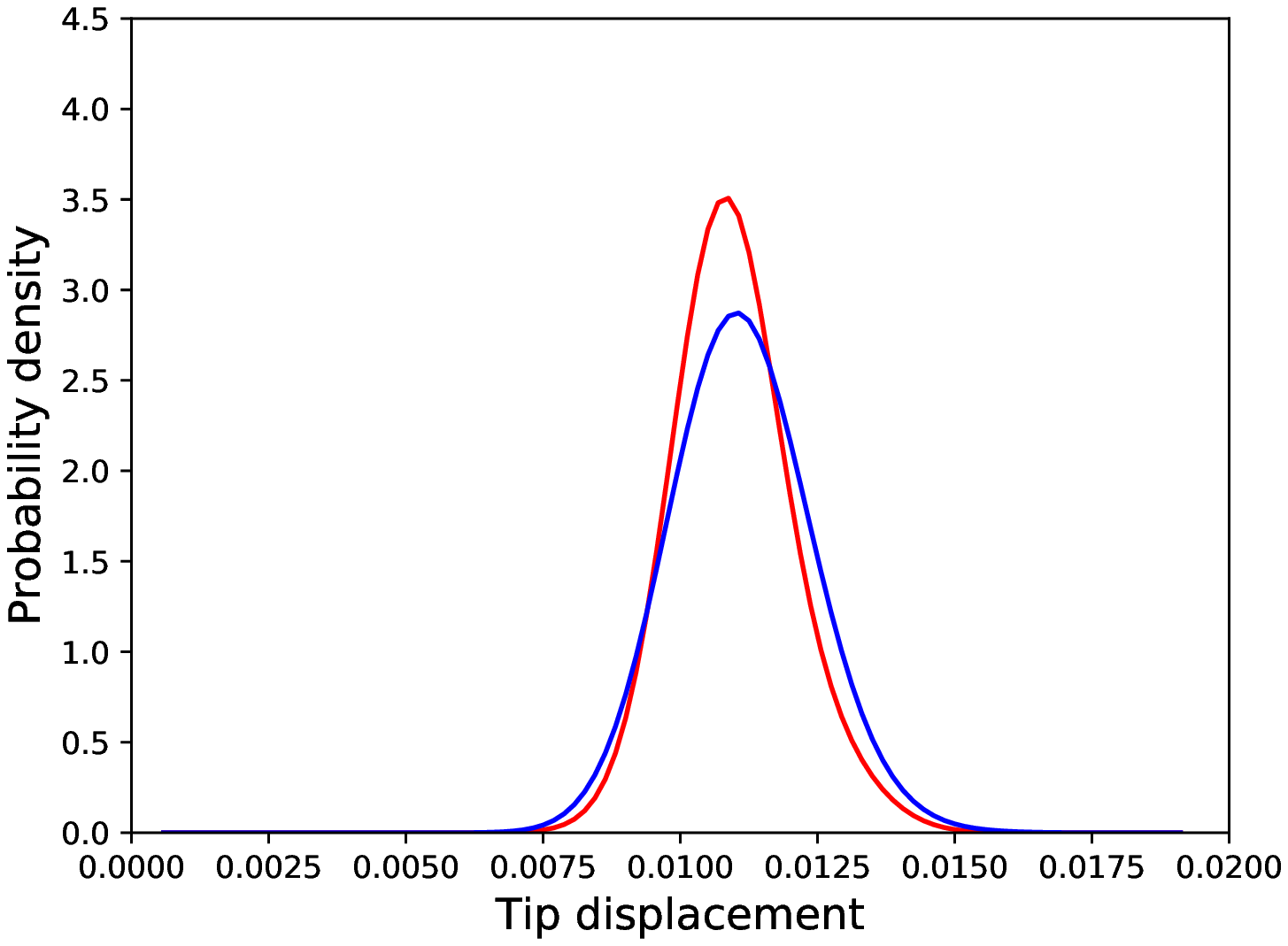}
    \caption{}
    \label{fig:sub_c2}
    \end{subfigure}
    \begin{subfigure}[b]{0.49\linewidth}
    \centering
    \includegraphics[width=0.99\linewidth]{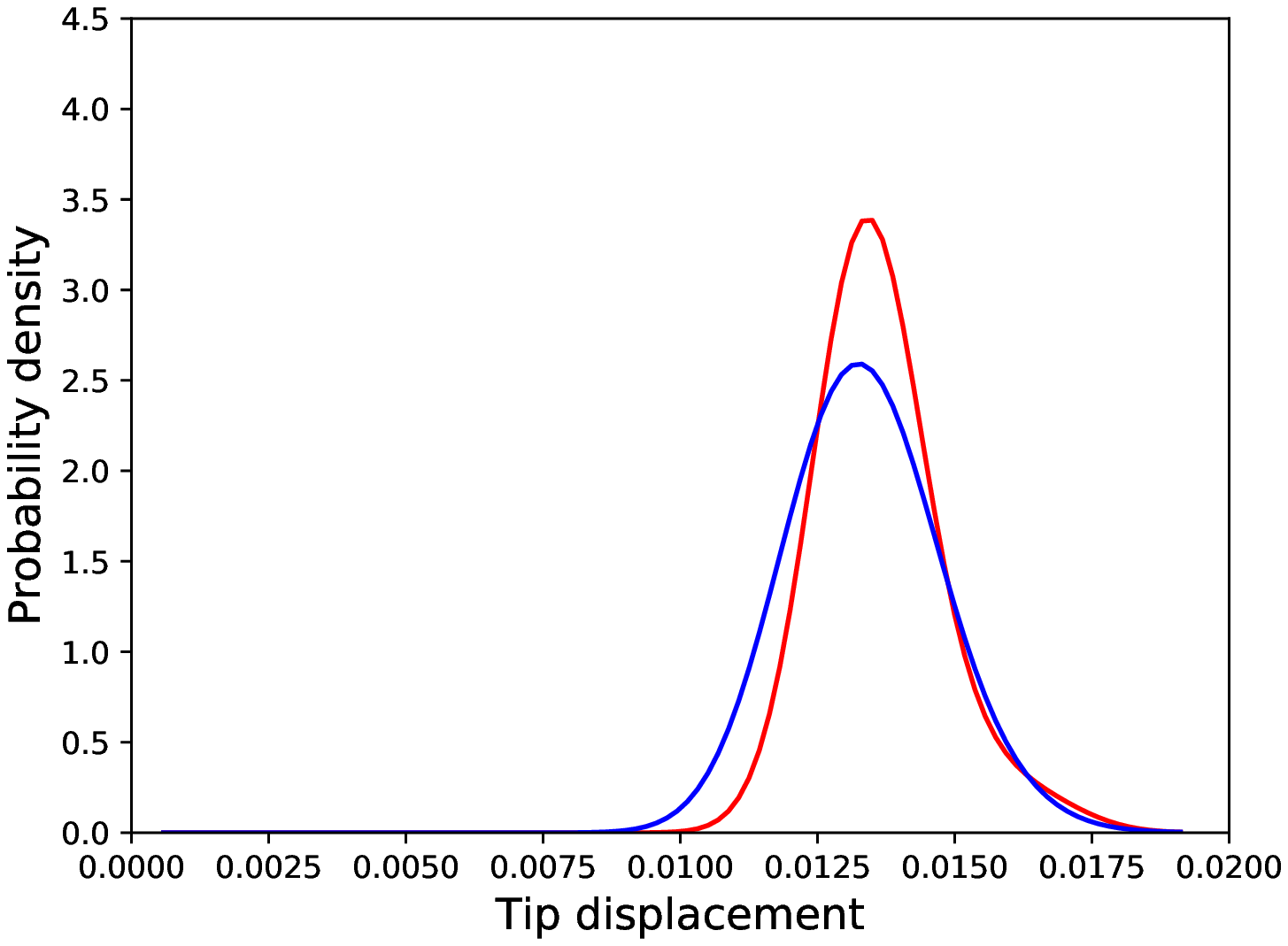}
    \caption{}
    \label{fig:sub_d2}
    \end{subfigure}
    \caption{Distributions of tip displacements corresponding to Monte Carlo samples (blue) and cGAN generated samples (red) regarding the linear problem, different load cases, (a) 30 load units, (b) 90 load units, (c) 150 load units and (d) 180 load units.}
    \label{fig:cGAN_linear_testing_distributions}
\end{figure}

The best average KL divergence, which was achieved by a model whose generator had $3000$ neurons in its hidden layer, was in the validation dataset $0.081$, and in the testing dataset $0.083$. Some of the distributions from the testing dataset are presented in Figure \ref{fig:cGAN_linear_testing_distributions}. It can be seen that the performance is not as good as the performance of the SFEM model, but given that the algorithm is a machine learning algorithm, it is acceptable.

\begin{figure}[!htbp]
    \centering
    \includegraphics[scale=0.5]{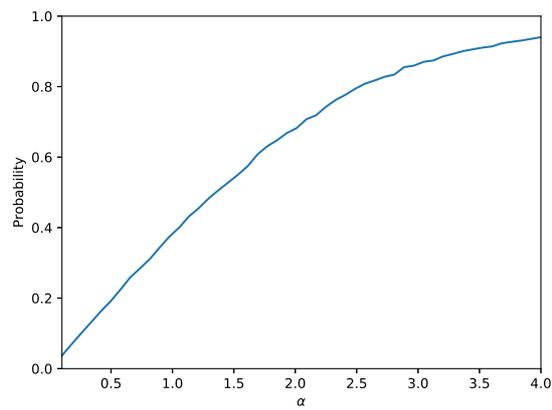}
    \caption{Probability as defined in equation (\ref{eq:alpha_mirror_prob}) as a function of the parameter $\alpha$ for the cGAN model applied on the linear cantilever case study.}
    \label{fig:alpha_probs_cGAN_linear}
\end{figure}

The cGAN model is also tested according to its ability to serve as an $\epsilon$ and an $\alpha$ mirror. As far as its potential use as an $\epsilon$-mirror is concerned, the maximum KL divergence observed on the available testing datasets was $0.168$ and therefore, based only on the data, the model can be considered an $\epsilon$-mirror for $\epsilon = 0.168$ regarding the distribution of the tip displacement of the cantilever. As far as the ability of the model to serve as an $\alpha$-mirror is concerned, similarly to Figure \ref{fig:alpha_probs_SFEM_linear}, the equivalent plot for the cGAN for the linear problem is shown in Figure \ref{fig:alpha_probs_cGAN_linear}.

\subsection{Application of cGAN in a nonlinear problem}
\label{sec:cGAN_nonlin}

\begin{figure}[!htbp]
    \centering
    \includegraphics[width=0.43\textwidth]{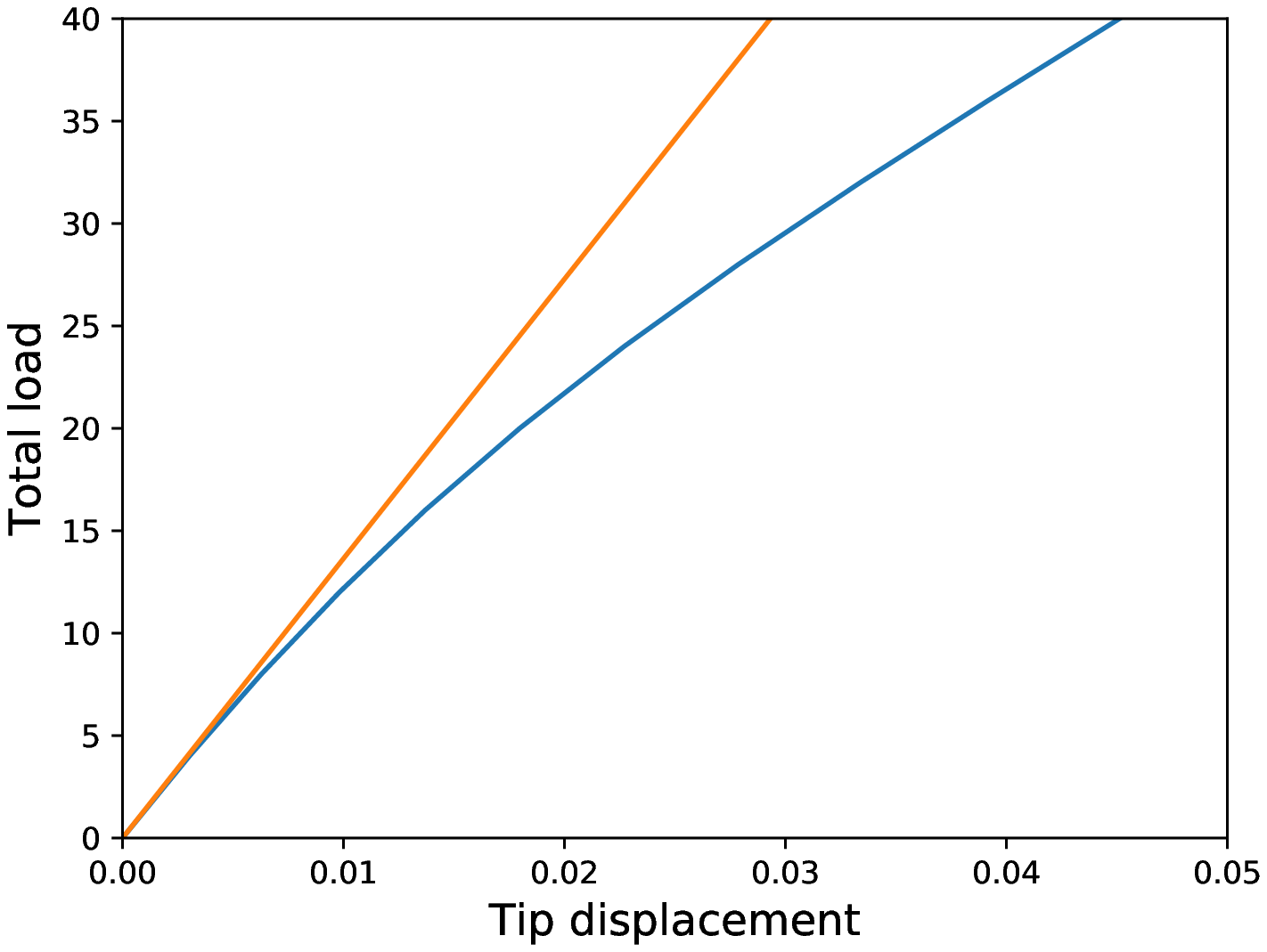}
    \caption{Load curves for the linear (orange) and the nonlinear (blue) material cantilevers.}
    \label{fig:linear_vs_nonlinear_curves}
\end{figure}

The next application for testing the potential of the cGAN in approximating the distribution conditioned on the load, is a similar cantilever but with material nonlinearity. More specifically, a softening material is considered; again, it is a simulated structure. The dimensions of the cantilever are the same. As a comparison, in Figure \ref{fig:linear_vs_nonlinear_curves} the load curves for the linear and the nonlinear structures are shown.

\begin{figure}[!htbp]
    \centering
    \includegraphics[width=0.43\textwidth]{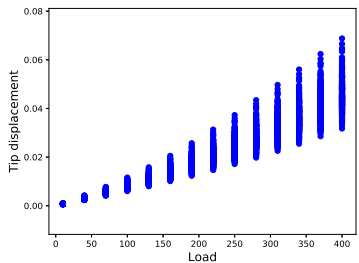}
    \caption{Tip displacement samples generated by the nonlinear cantilever.}
    \label{fig:nonlin_samples}
\end{figure}

In this case, random Young's moduli were sampled from a normal distribution with mean value equal to $2 \times 10^{9}$ and standard deviation $0.1\times 2 \times 10^{9}$. Each nonlinear simulation was performed using a total load of $400$ load units and $40$ loadsteps. After finding the solution of the \textit{Newton-Raphson} iterations, the displacements were stored for every iteration; in this way, every nonlinear simulation provided $40$ tip displacements, one for every load in the set $\{10, 20, 30... , 400\}$. For every load case, $500$ samples were generated. In Figure \ref{fig:nonlin_samples}, samples are shown for different loads in the dataset $\{10, 40, 70... , 400\}$ which is also considered the training dataset for the cross-validation procedure followed (load units $\{20, 50, 80... , 380\}$ comprised the validation dataset and load units $\{30, 60, 90... , 390\}$ the testing dataset).

\begin{figure}[!htbp]
    \centering
    \begin{subfigure}[b]{0.49\linewidth}
    \centering
    \includegraphics[width=0.99\linewidth]{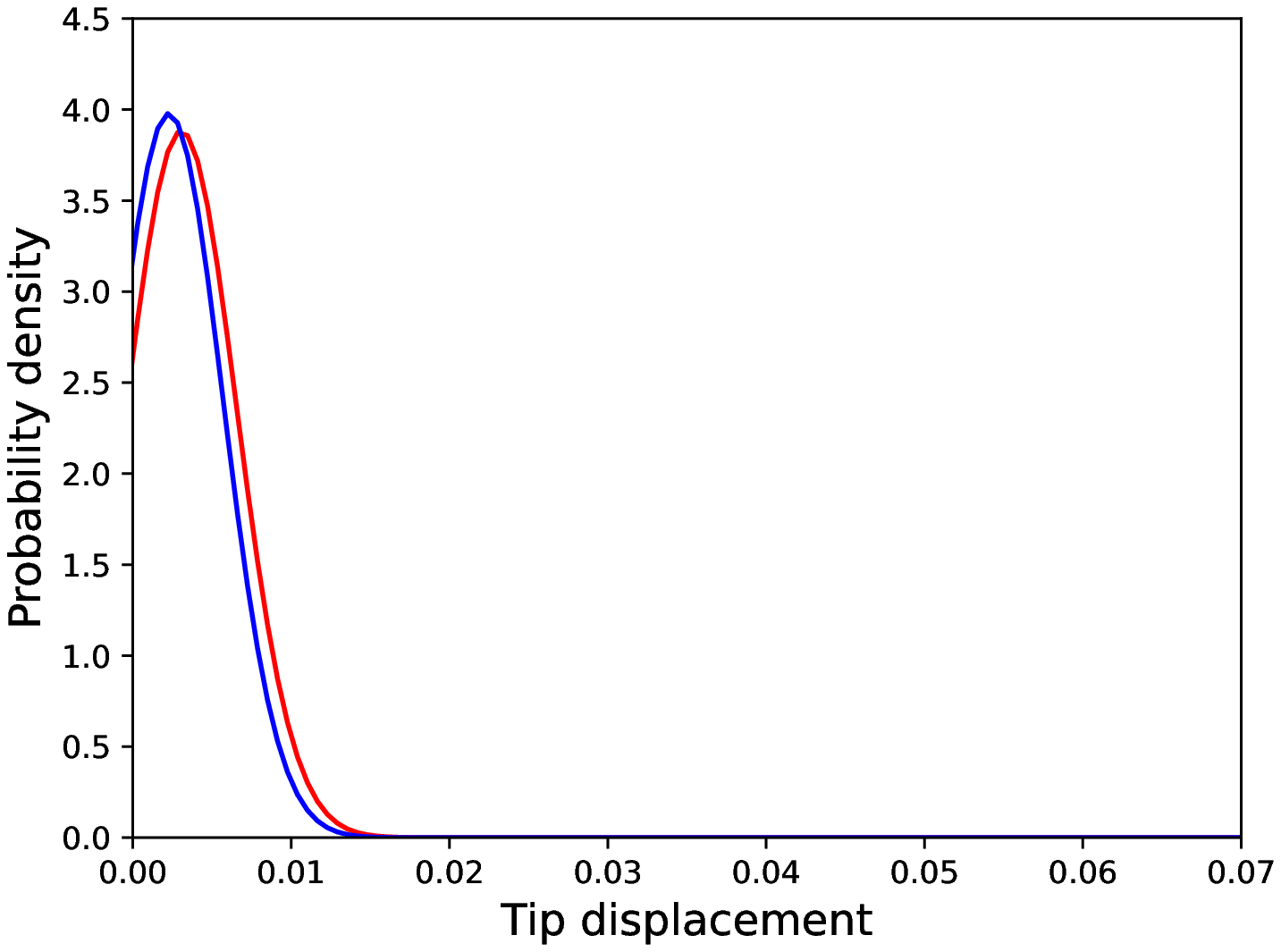}
    \caption{}
    \label{fig:sub_a3}
    \end{subfigure}
    \begin{subfigure}[b]{0.49\linewidth}
    \centering
    \includegraphics[width=0.99\linewidth]{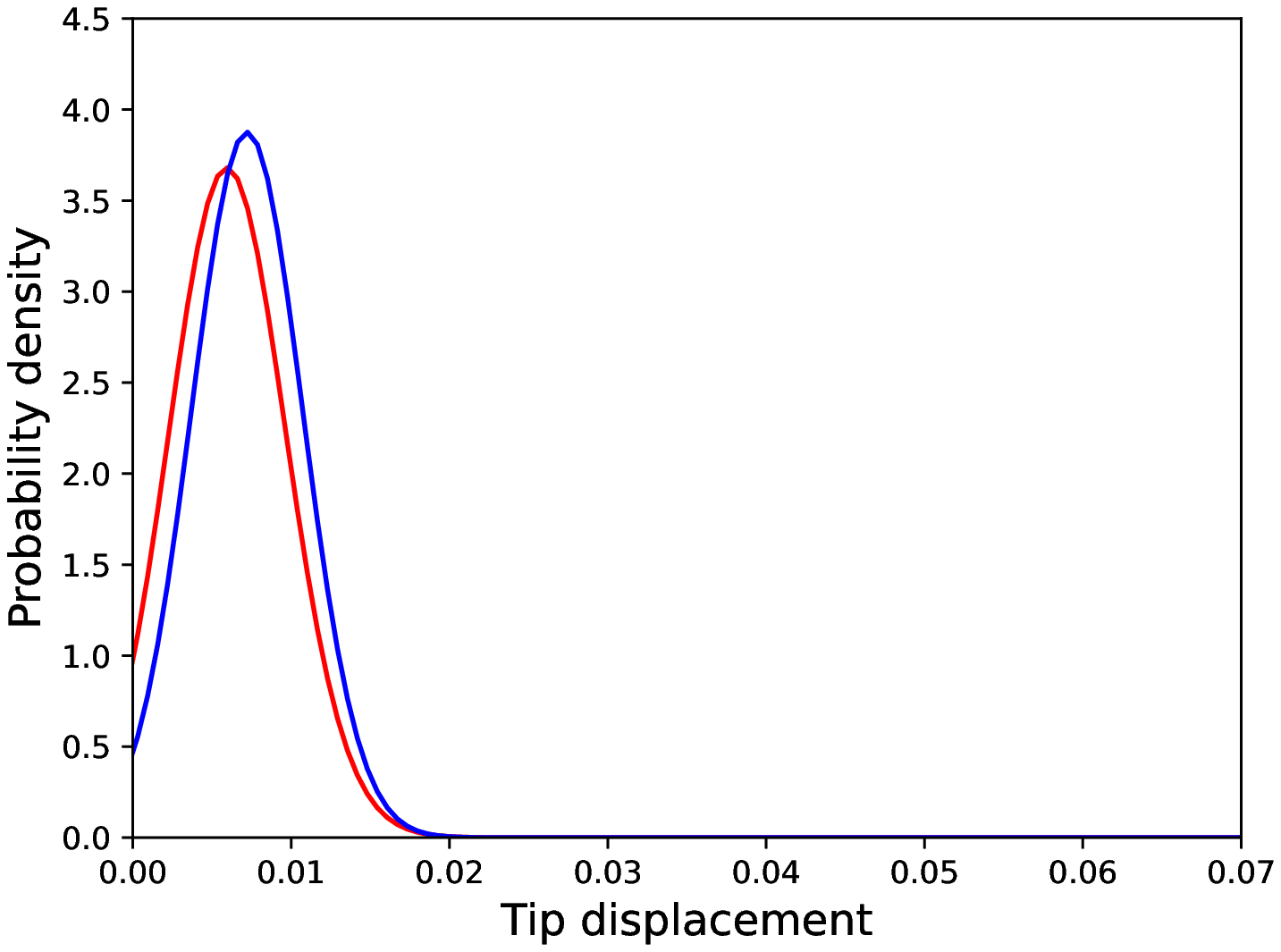}
    \caption{}
    \label{fig:sub_b3}
    \end{subfigure}
    \begin{subfigure}[b]{0.49\linewidth}
    \centering
    \includegraphics[width=0.99\linewidth]{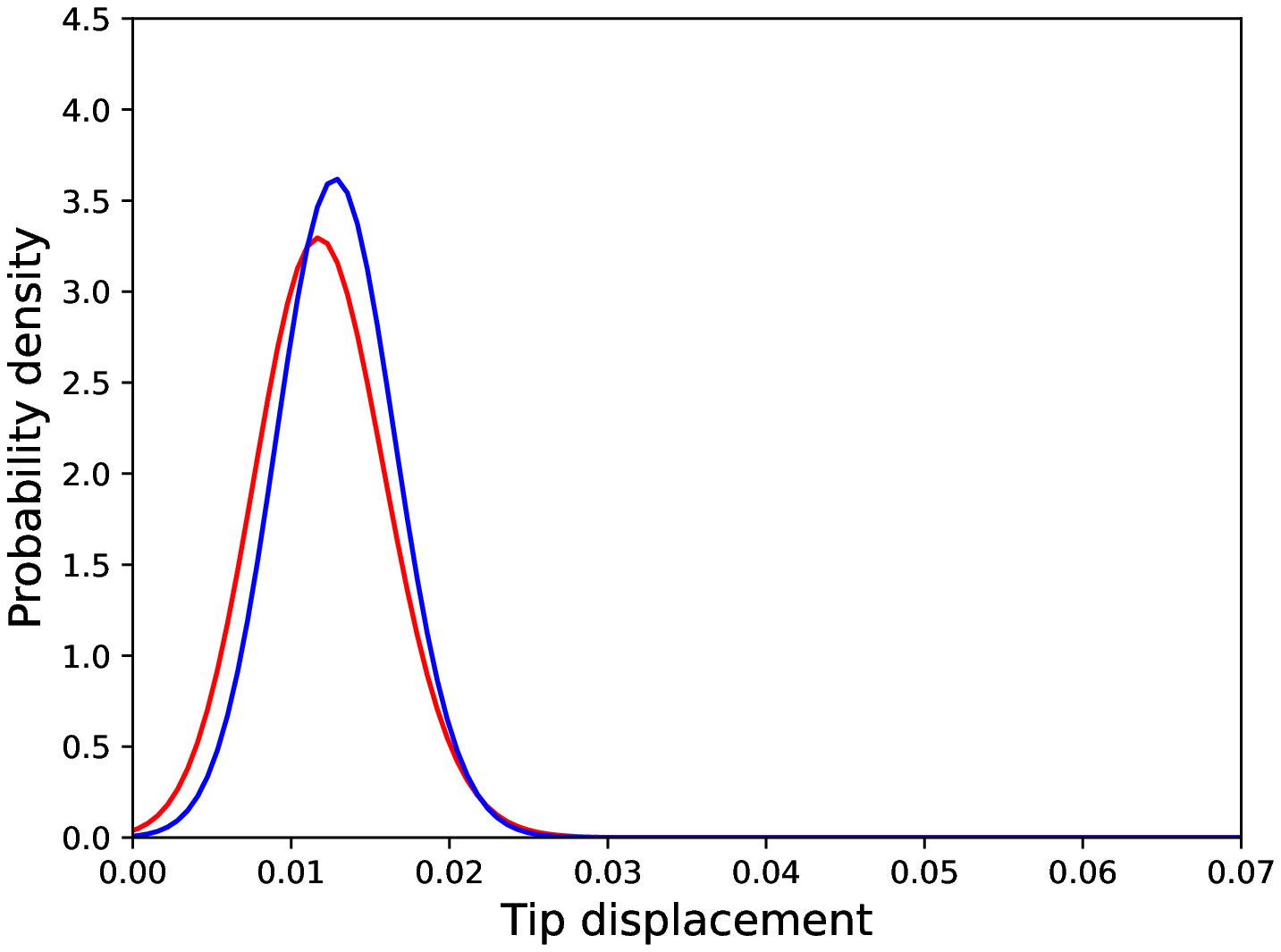}
    \caption{}
    \label{fig:sub_c3}
    \end{subfigure}
    \begin{subfigure}[b]{0.49\linewidth}
    \centering
    \includegraphics[width=0.99\linewidth]{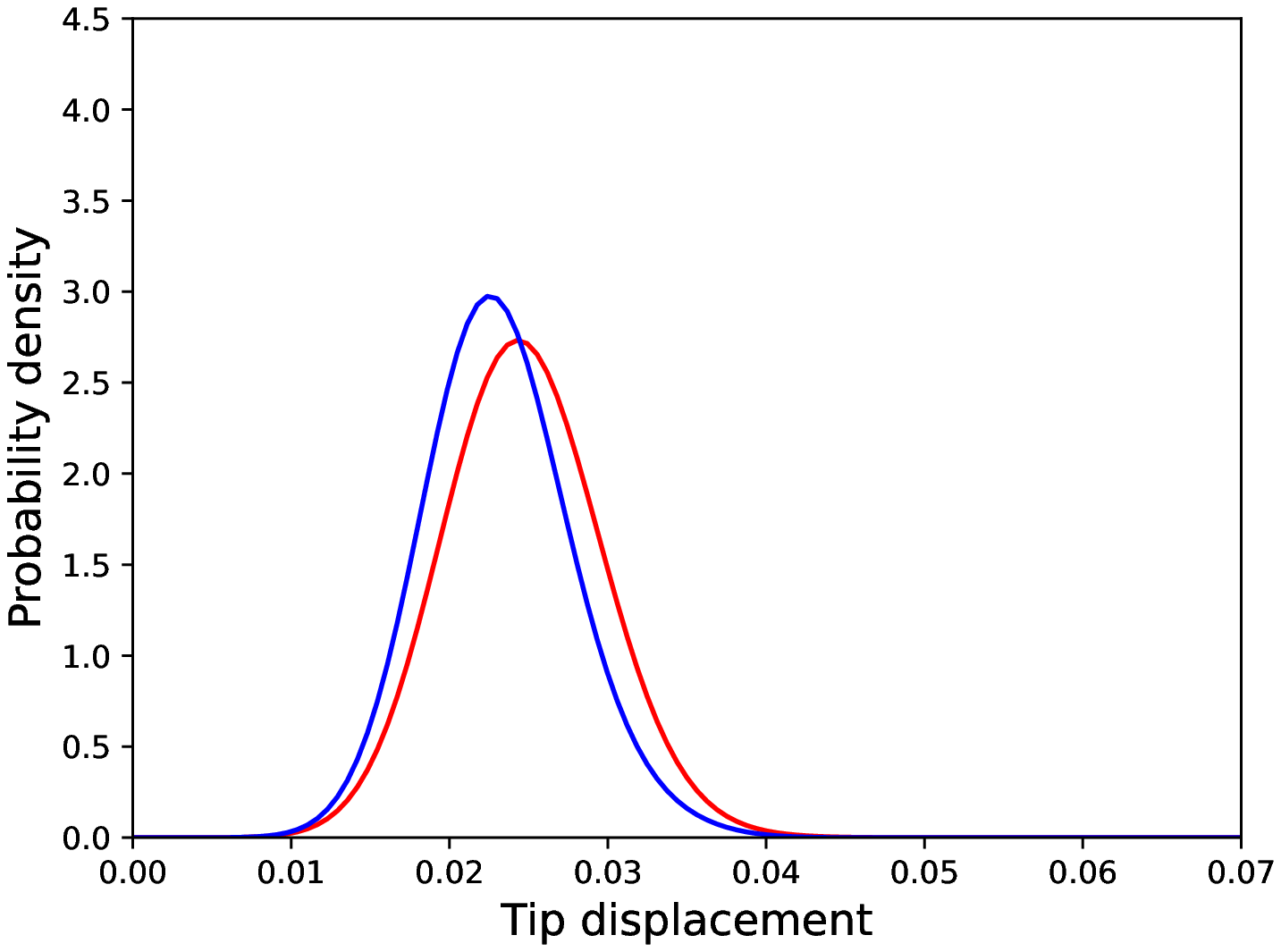}
    \caption{}
    \label{fig:sub_d3}
    \end{subfigure}
    \begin{subfigure}[b]{0.49\linewidth}
    \centering
    \includegraphics[width=0.99\linewidth]{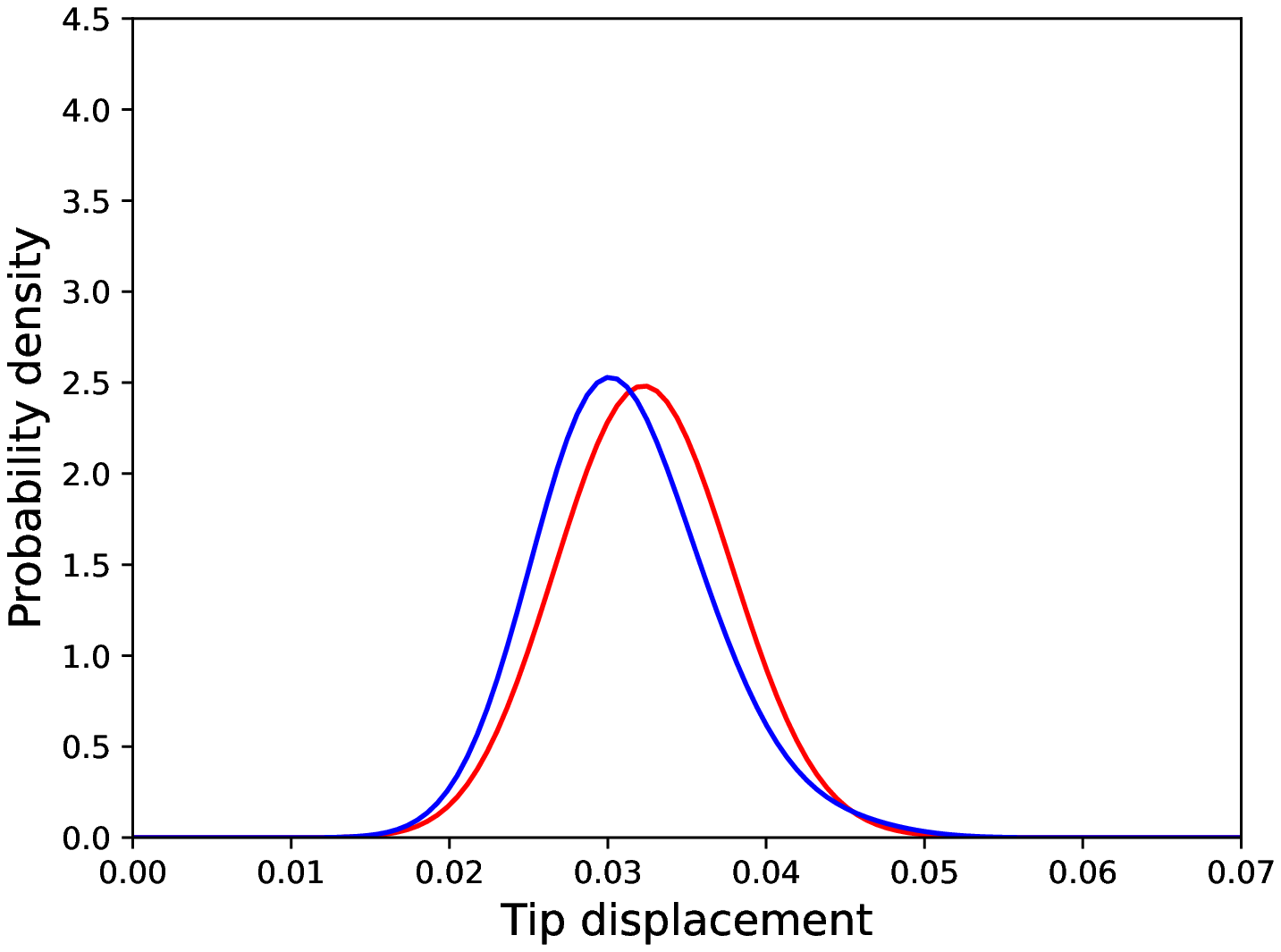}
    \caption{}
    \label{fig:sub_e3}
    \end{subfigure}
    \begin{subfigure}[b]{0.49\linewidth}
    \centering
    \includegraphics[width=0.99\linewidth]{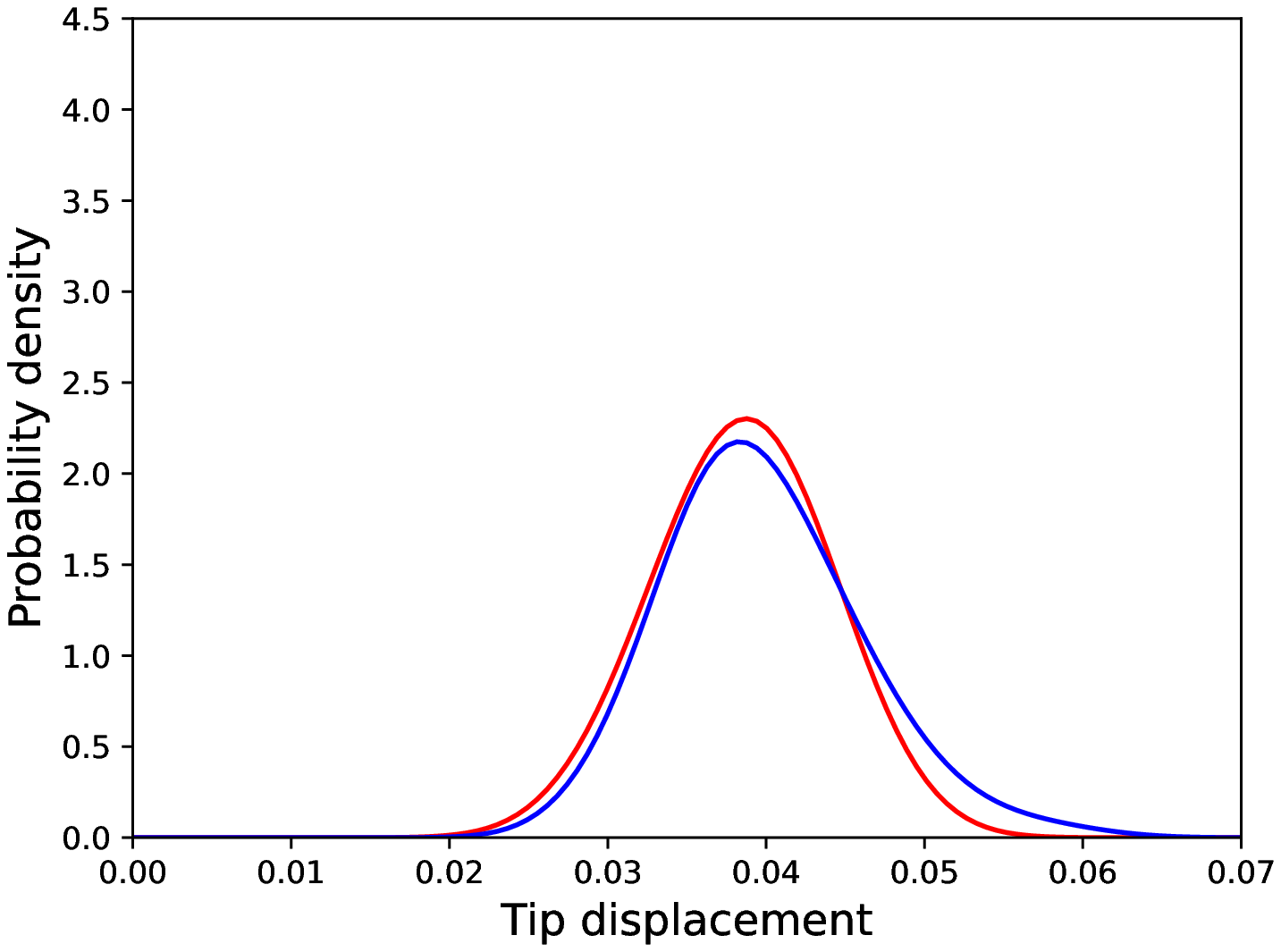}
    \caption{}
    \label{fig:sub_f3}
    \end{subfigure}
    \caption{Distributions of tip displacements corresponding to Monte Carlo samples (blue) and cGAN generated samples (red) regarding the nonlinear problem, for different load cases; (a) 30 load units, (b) 90 load units, (c) 150 load units, (d) 240 load units, (e) 300 load units and (f) 360 load units.}
    \label{fig:cGAN_nonlinear_testing_distributions}
\end{figure}

An identical approach to the linear case was followed except for the sizes considered for the hidden layer of the networks. The sizes tested belonged to the set $\{10, 20, 30, ... 3000\}$. Again, following the same cross-validation procedure, the networks that yielded the best results had $110$ neurons in their hidden layers. The lowest average KL divergence in the validation dataset was found to be $0.045$ and that network yielded KL divergence equal to $0.050$ on the testing dataset. Generated distributions corresponding to codes of the testing dataset, in comparison to the real ones (acquired from the simulated structure) are shown in Figure \ref{fig:cGAN_nonlinear_testing_distributions}. It is observed that as the spread of the distribution along with the load increases, the algorithm has efficiently learnt to generate samples with greater spread.

\begin{figure}[!htbp]
    \centering
    \begin{subfigure}[b]{0.49\linewidth}
    \centering
    \includegraphics[width=0.99\linewidth]{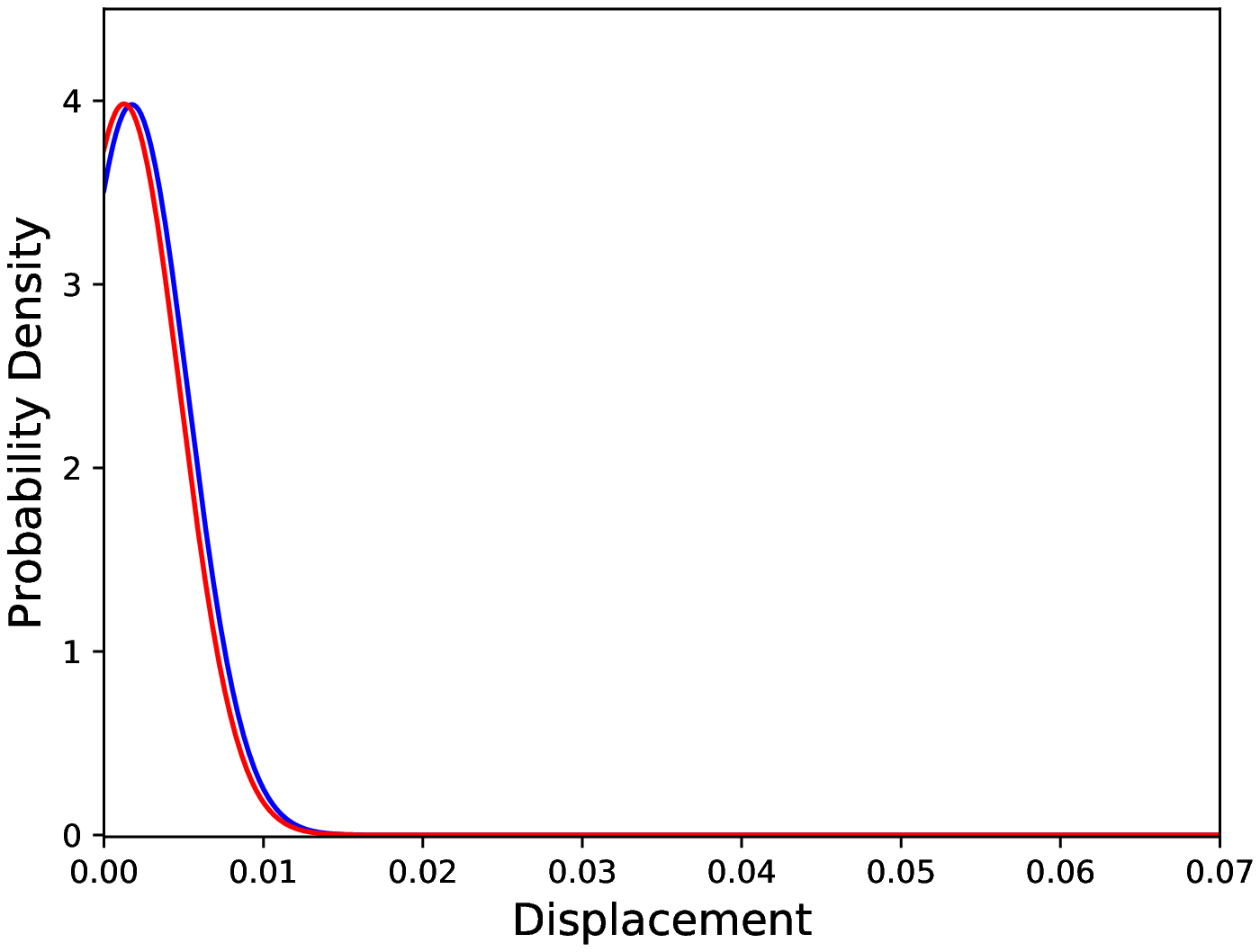}
    \caption{}
    \label{fig:sub_a3}
    \end{subfigure}
    \begin{subfigure}[b]{0.49\linewidth}
    \centering
    \includegraphics[width=0.99\linewidth]{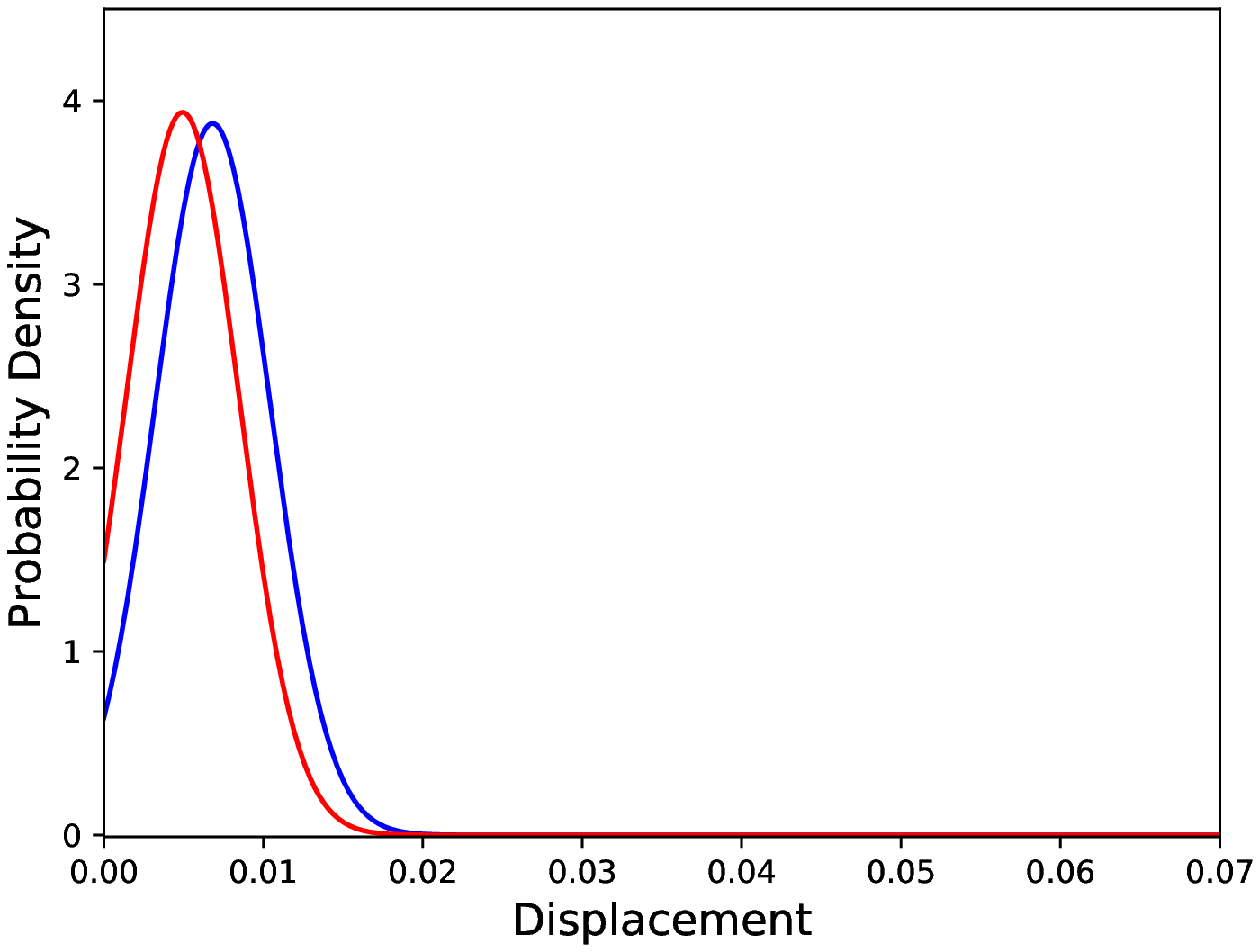}
    \caption{}
    \label{fig:sub_b3}
    \end{subfigure}
    \begin{subfigure}[b]{0.49\linewidth}
    \centering
    \includegraphics[width=0.99\linewidth]{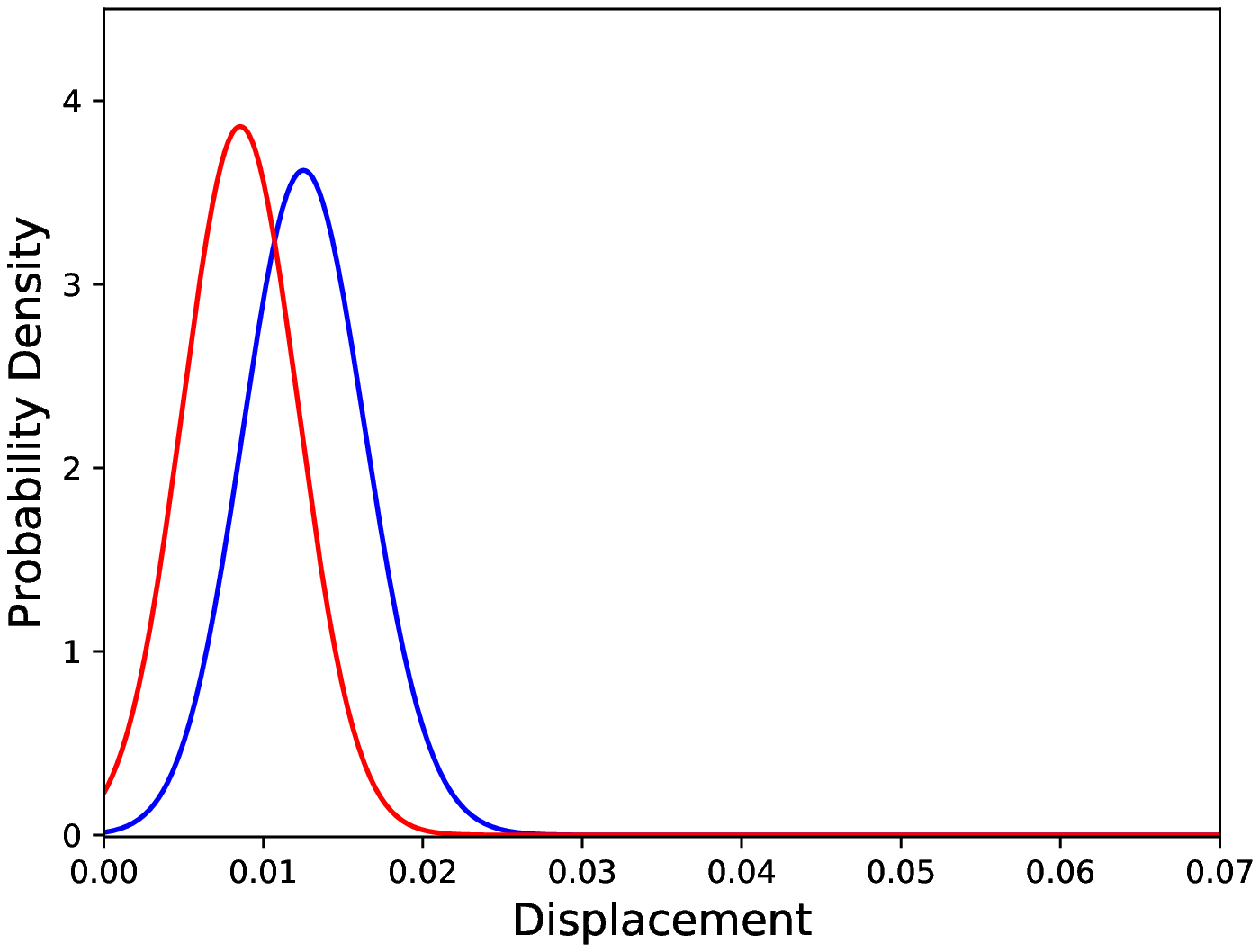}
    \caption{}
    \label{fig:sub_c3}
    \end{subfigure}
    \begin{subfigure}[b]{0.49\linewidth}
    \centering
    \includegraphics[width=0.99\linewidth]{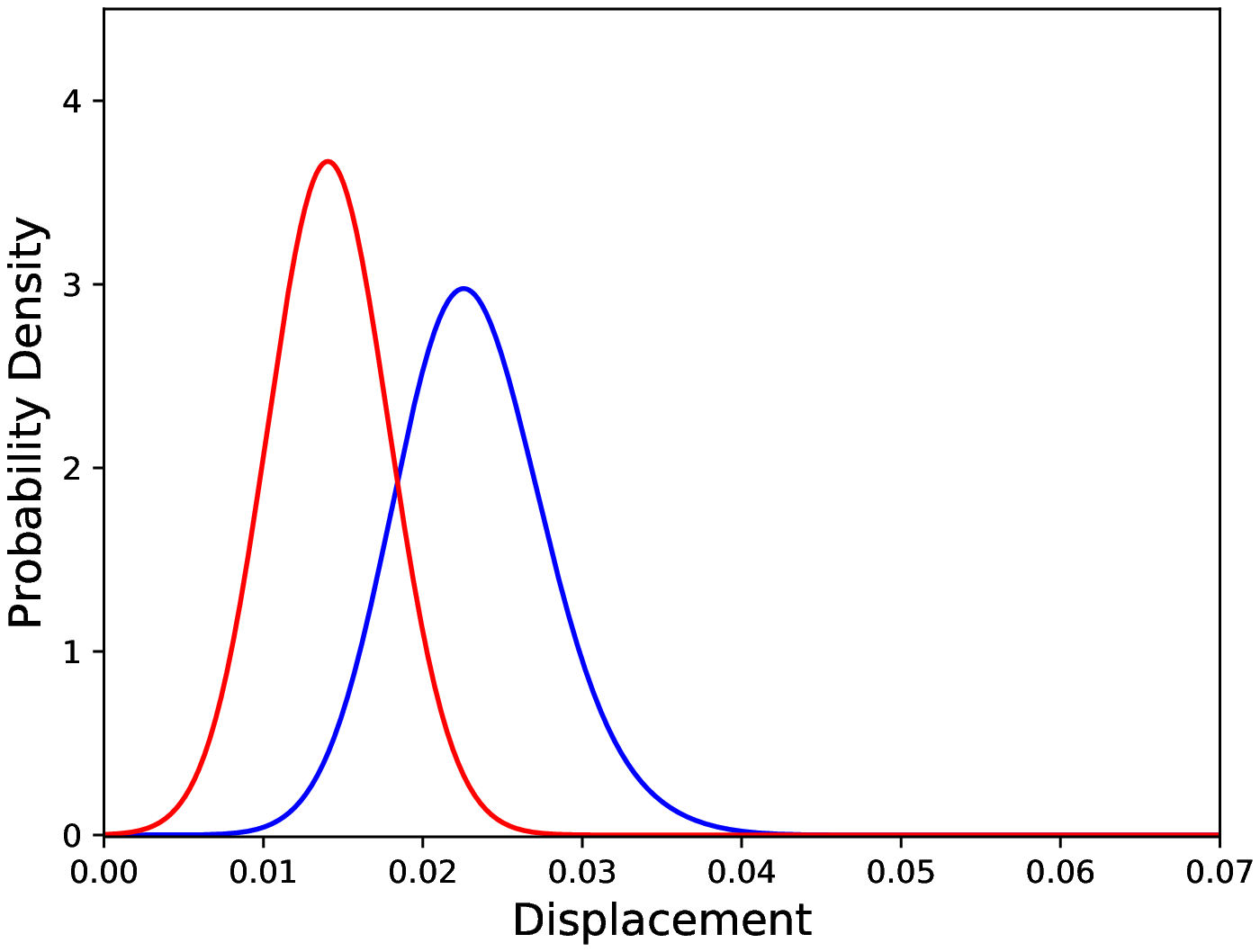}
    \caption{}
    \label{fig:sub_d3}
    \end{subfigure}
    \begin{subfigure}[b]{0.49\linewidth}
    \centering
    \includegraphics[width=0.99\linewidth]{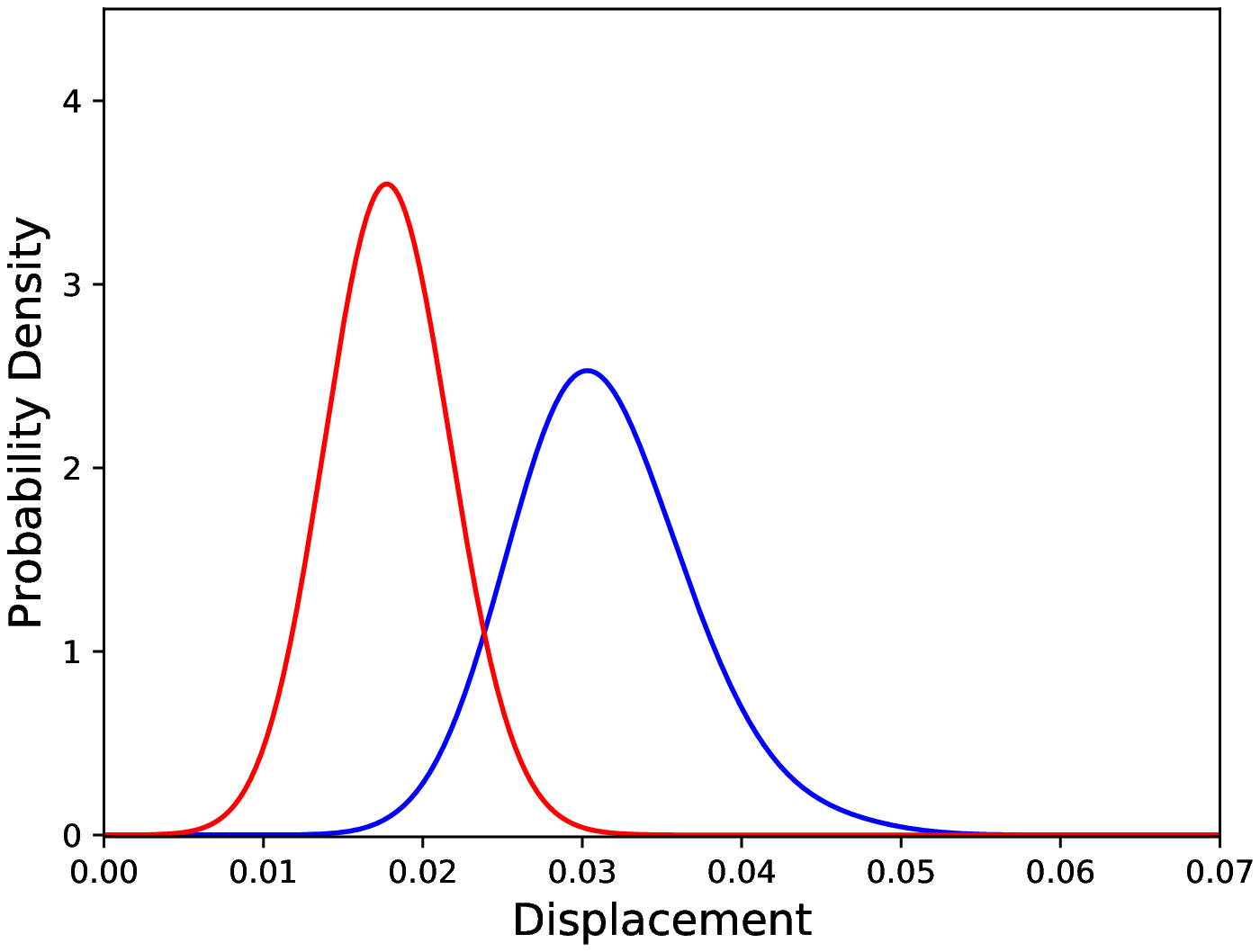}
    \caption{}
    \label{fig:sub_e3}
    \end{subfigure}
    \begin{subfigure}[b]{0.49\linewidth}
    \centering
    \includegraphics[width=0.99\linewidth]{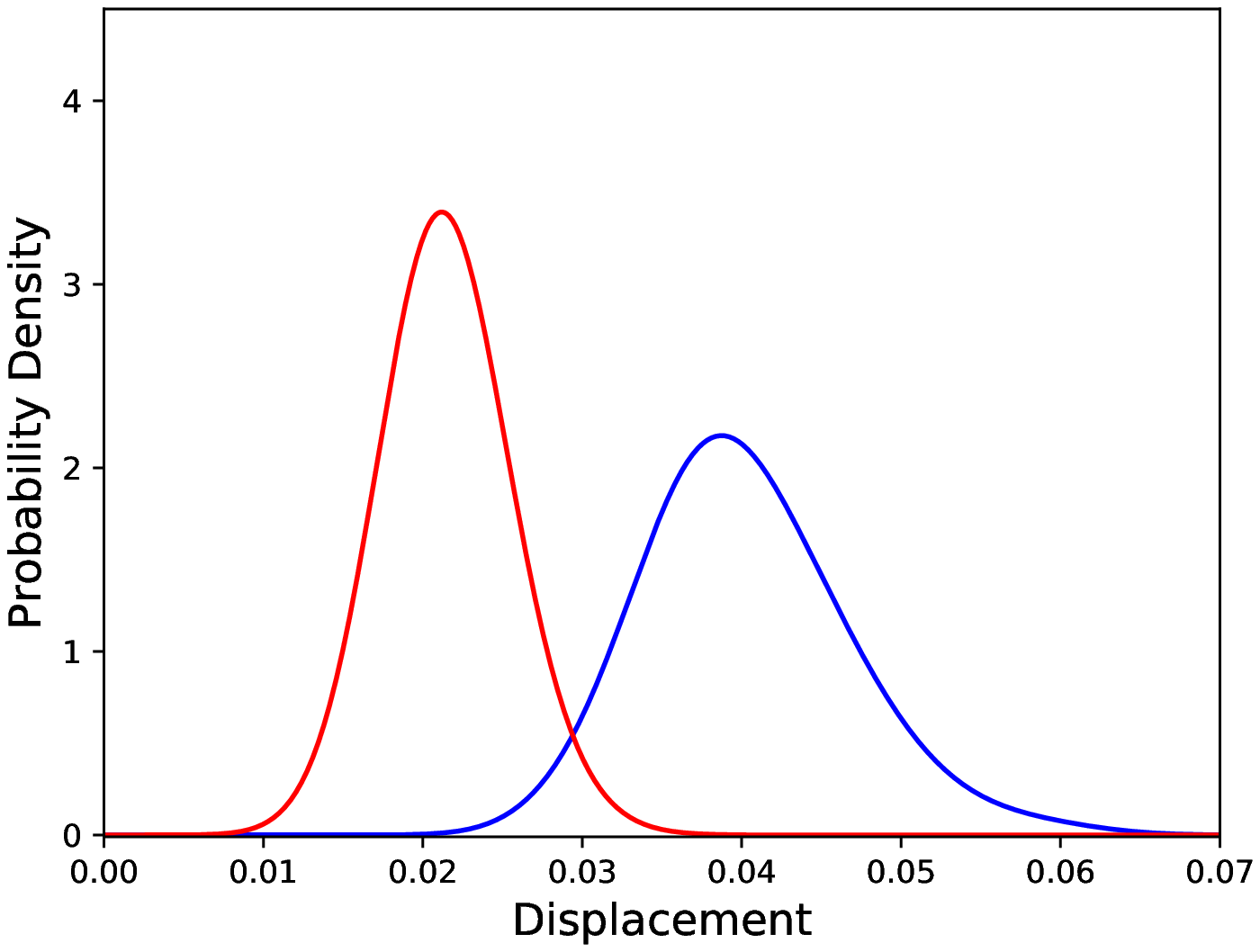}
    \caption{}
    \label{fig:sub_f3}
    \end{subfigure}
    \caption{Distributions of tip displacements corresponding to Monte Carlo samples (blue) and SFEM generated samples (red) regarding the nonlinear problem, for different load cases; (a) 30 load units, (b) 90 load units, (c) 150 load units, (d) 240 load units, (e) 300 load units and (f) 360 load units.}
    \label{fig:SFEM_nonlinear_testing_distributions}
\end{figure}

In order to compare the cGAN approach with the SFE method, a SFE model was calibrated (performing an exhaustive search for the optimal parameters) according to the data of the Monte Carlo simulations of the nonlinear cantilever. Although the linear physics formulation does not suffice in order to accurately explain the behaviour of the nonlinear cantilever, the SFE model could yield accurate enough predictions in order to be used as a mirror. The results for the same loads as in Figure \ref{fig:cGAN_nonlinear_testing_distributions} are shown in Figure \ref{fig:SFEM_nonlinear_testing_distributions}. The results reveal that the SFE model is unable to capture the effect of the nonlinearity on the distributions of the tip displacements; such a result is expected, since nonlinearity is not included in the formulation of the SFE model used. This result is also confirmed by the average KL divergence between the real and predicted distributions in the testing dataset, which was 3.43.

\begin{figure}[!htbp]
    \centering
    \includegraphics[scale=0.5]{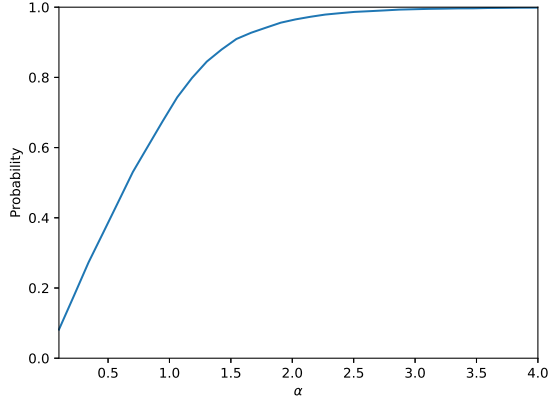}
    \caption{Probability defined in equation (\ref{eq:alpha_mirror_prob}) as a function of the parameter $\alpha$ for the cGAN model applied on the nonlinear cantilever case study.}
    \label{fig:alpha_probs_cGAN_nonlinear}
\end{figure}

Similarly, the cGAN is tested as an $\epsilon$-mirror regarding its ability to predict the distribution of the tip displacement of the cantilever. The maximum value
of the KL divergence on the available testing datasets now is $0.25$. Regarding its use as an $\alpha$-mirror, the plot showing the probability defined in equation (\ref{eq:alpha_mirror_prob}) is shown in Figure \ref{fig:alpha_probs_cGAN_nonlinear}. Although the KL divergence in this later case study was lower than in the application of the cGAN in the linear case, the $\alpha$-curve this time is not as good as the one in Figure \ref{fig:alpha_probs_cGAN_linear}, in the sense that eventually, for larger values of $\alpha$ (for example $\alpha = 4.0$), the cGAN applied in the linear case, is able to include in the desired interval a larger portion of the observed data.

\section{Hybrid model approach}
\label{sec:hybrid}

\subsection{Definition of the hybrid model}
\label{sec:hybrid_definition}

The approaches presented so far were either completely physics-based (SFE model) or data driven, informed partially by the physics of the problem (a cGAN informed by the value of the load). It became clear that a SFE model without the appropriate physics included (such as nonlinear effects) cannot describe efficiently a phenomenon like the tip displacement of the nonlinear cantilever. In real-life applications, insufficient physics in SFE formulation might mean that the users do not know where the uncertainty is coming from (epistemic uncertainty about the aleatory uncertainty). On the other hand, the cGAN was able to capture the effect of the load on the distribution of the tip displacement both in the linear and the nonlinear case without any knowledge about the source of uncertainty. The efficiency of the algorithm lies in its insensitivity to the linearity of the underlying problem.

Naturally, one would prefer a model that includes further understanding of the underlying physics rather than just using the value of the input in the physical system; i.e. the load in the case studies. In order to define such a coupling, a hybrid approach is followed. The SFE model is used as a first estimator of the target distributions and afterwards the cGAN algorithm is applied in order to correct them.  A similar approach for nonlinear modelling has been presented in \citep*{rogers2017grey} and another about learning such model discrepancies has been developed in \citep*{gardner152learning}.

In \citep*{rogers2017grey}, two types of such models are defined. The first is the $A$-type models in which the black-box model is exploited in order to infer the error between the white-box model predictions and the observations. Inference is performed on the error between the white-box model and the real observation, i.e.,
\begin{equation}
    \label{eq:A_type_model}
    \delta(X) = y(X) - f(X)
\end{equation}
where $y$ is the observation, $f$ is the white-box model, $X$ is the input variable to the model and $\delta$ is the error sought to be modelled using a black-box/machine learning model. This type of model is not suitable for the current work, since it contradicts the desired definition of a generative model as a mirror. This is because the generative modelling framework requires unknown parameters affecting the structure, while A-type models requires definition of exact errors for specific inputs on the model. In order to define a training dataset to follow the $A$-type model approach, one would need to specify errors between specific predictions of the white box model and observation. Although, when uncertainty is admitted, that cannot happen, as the output of the white-box model is a function of both a set of controlled and uncontrolled variables; i.e. $f(X) = f(\underline{e}_{c}, \underline{e}_{u})$. Since some variables $\underline{e}_{u}$ are not observable in nature (for example the Young's modulus exact field within the volume of a structure), their exact values cannot be used in order to define such a dataset, which should comprise pairs between model outputs $f(\underline{e}_{c}, \underline{e}_{u})$ and observations $y(\underline{e}_{c}, \underline{e}_{u})$.

As a result the approach to be followed here is to define a $B$-type model. This latter type of hybrid model is based on the idea that the black-box model performs inference using the output of the white-box model and the controlled variables, i.e.,
\begin{equation}
    \label{eq:B_type_model}
    y(X) = g(X, f(X))
\end{equation}
where $g$ is the black-box model. Specifically for the problem addressed here, the cGAN is called to transform the predicted distributions of the SFE model into the correct ones observed in the data (red and blue distributions respectively in Figure \ref{fig:cGAN_nonlinear_testing_distributions}) being also informed by the value of the load/code. Type B models are applicable in the case of generative models, since in order to define the real and the predicted distributions one needs to integrate both sides of equation (\ref{eq:B_type_model}), which can be done without knowledge of the uncontrolled variables because their effect is explained via the generative model $g$. On the contrary, integration of both sides of equation (\ref{eq:A_type_model}) is not feasible without knowledge of the values of the uncontrolled variables for every observation.

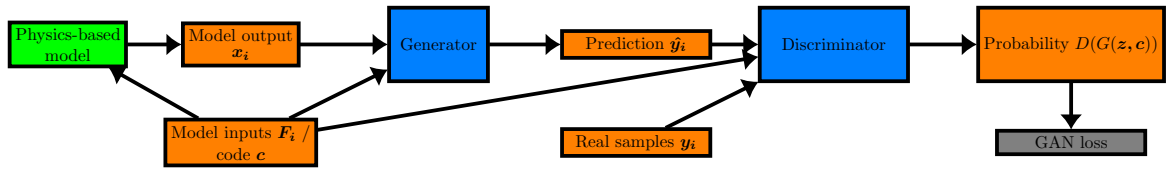
\begin{figure}[!htbp]
    \centering
    \begin{tikzpicture}[scale=0.65, every node/.style={transform shape}]
        \definecolor{blue1}{RGB}{0, 128, 255}

        \node (-1) at (0.0, -2.0) [draw, line width=0.5mm, fill=orange] {\shortstack{Model inputs $\bm{F_{i}}$ /\\ code $\bm{c}$}};

        \node (0) at (-3.5, 0.0) [draw, line width=0.5mm, fill=green] {\shortstack{Physics-based\\model}};

        \draw[-{Latex[width=3mm, length=2mm]}, line width=0.5mm] (-1) to (0);

        \node (1) at (0.0, 0.0) [draw, line width=0.5mm, fill=orange] {\shortstack{Model output\\ $\bm{x_{i}}$}};
        \draw[-{Latex[width=3mm, length=2mm]}, line width=0.5mm] (0) to (1);
        \node (2) at (4.0, 0.0) [draw, line width=0.5mm, fill=blue1, minimum height=1.5cm, minimum width=2cm] {\shortstack{Generator}};
        \draw[-{Latex[width=3mm, length=2mm]}, line width=0.5mm] (1) to (2);

        \node (3) at (8.0, 0.0) [draw, line width=0.5mm, fill=orange, minimum width=3.0cm] {Prediction $\bm{\hat{y_{i}}}$};

        \node (4) at (8.0, -2.0) [draw, line width=0.5mm, fill=orange, minimum width=3.0cm] {Real samples $\bm{y_{i}}$};

        \draw[-{Latex[width=3mm, length=2mm]}, line width=0.5mm] (2) to (3);

        \node (5) at (12.0, 0.0) [draw, line width=0.5mm, fill=blue1, minimum height=1.5cm, minimum width=3.0cm] {Discriminator};

        \draw[-{Latex[width=3mm, length=2mm]}, line width=0.5mm] (3) to (5);
        \draw[-{Latex[width=3mm, length=2mm]}, line width=0.5mm] (4) to (5);

        \node (6) at (16.8, 0.0) [draw, line width=0.5mm, fill=orange, minimum height=1.5cm, minimum width=3.0cm] {Probability $D(G(\bm{z, c}))$};
        \draw[-{Latex[width=3mm, length=2mm]}, line width=0.5mm] (5) to (6);

        \node (8) at (16.8, -2.0) [draw, line width=0.5mm, fill=gray, minimum width=3.0cm] {GAN loss};
        \draw[-{Latex[width=3mm, length=2mm]}, line width=0.5mm] (6) to (8);

        \draw[-{Latex[width=3mm, length=2mm]}, line width=0.5mm] (-1) to (2);
        \draw[-{Latex[width=3mm, length=2mm]}, line width=0.5mm] (-1) to (5);

    \end{tikzpicture}
    \caption{Layout of the SFE-cGAN hybrid model.}
    \label{fig:epist_uncert_estimation_layout}
\end{figure}

The output distributions of the SFE model are used as the latent sampling space of the cGAN. The physics-based model is yielding results based on the ``linear part'' of the physics of the system and the cGAN is called to learn only the ``nonlinear part''. This approach is expected to assist the cGAN training by providing latent variables whose distribution is closer to the real ones. In the completely data-driven case presented earlier, the cGAN simply learnt transformations of the same latent distribution as a function of the code. In the hybrid approach, the latent distribution is also a function of the code, aiming in assisting the training procedure. The layout of the combined model is shown in Figure \ref{fig:epist_uncert_estimation_layout}.

\subsection{Application of the hybrid model}
\label{sec:hybrid_model_application}

\begin{figure}[!htbp]
    \centering
    \begin{subfigure}[b]{0.49\linewidth}
    \centering
    \includegraphics[width=0.99\linewidth]{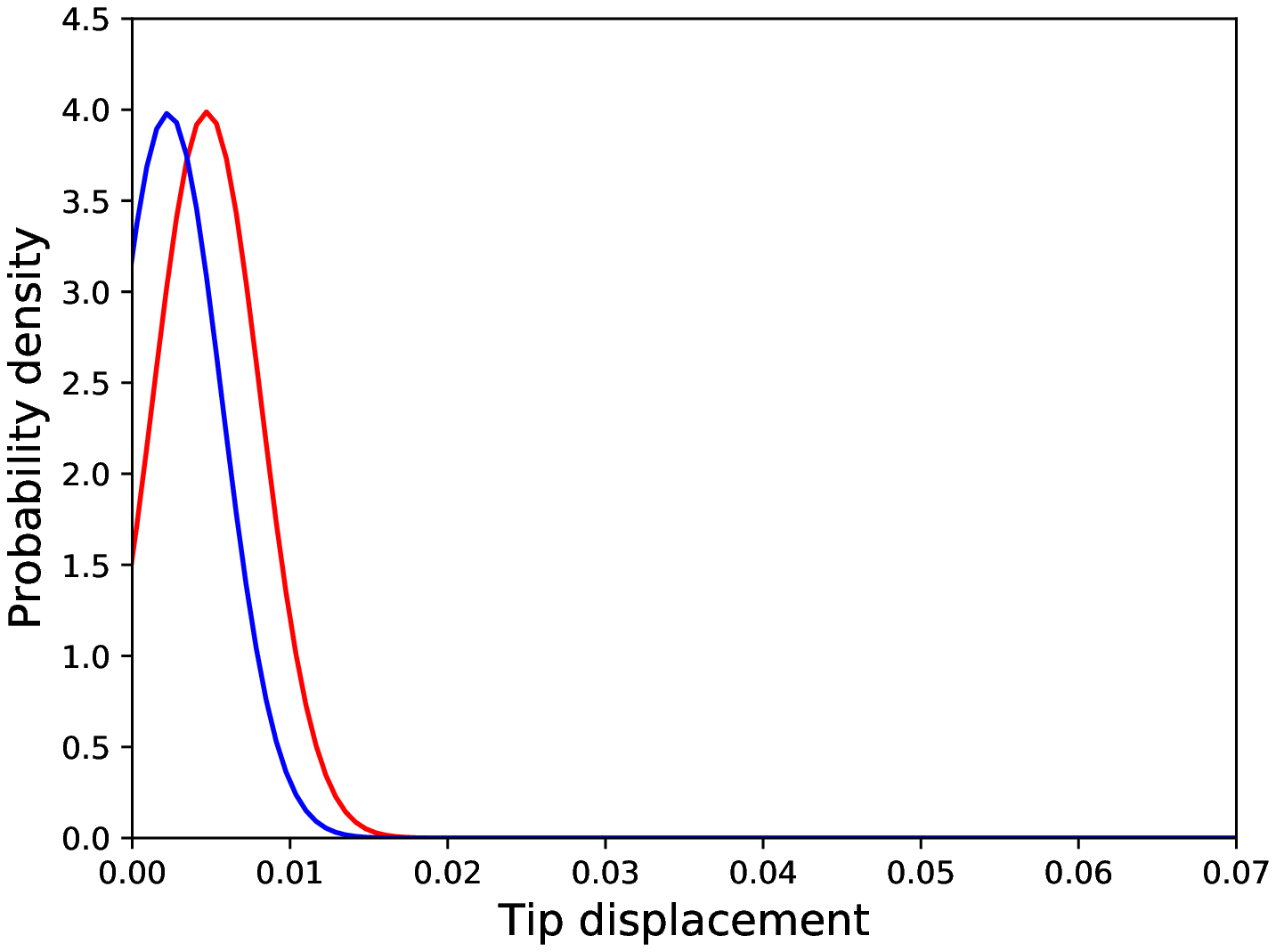}
    \caption{}
    \label{fig:sub_a3}
    \end{subfigure}
    \begin{subfigure}[b]{0.49\linewidth}
    \centering
    \includegraphics[width=0.99\linewidth]{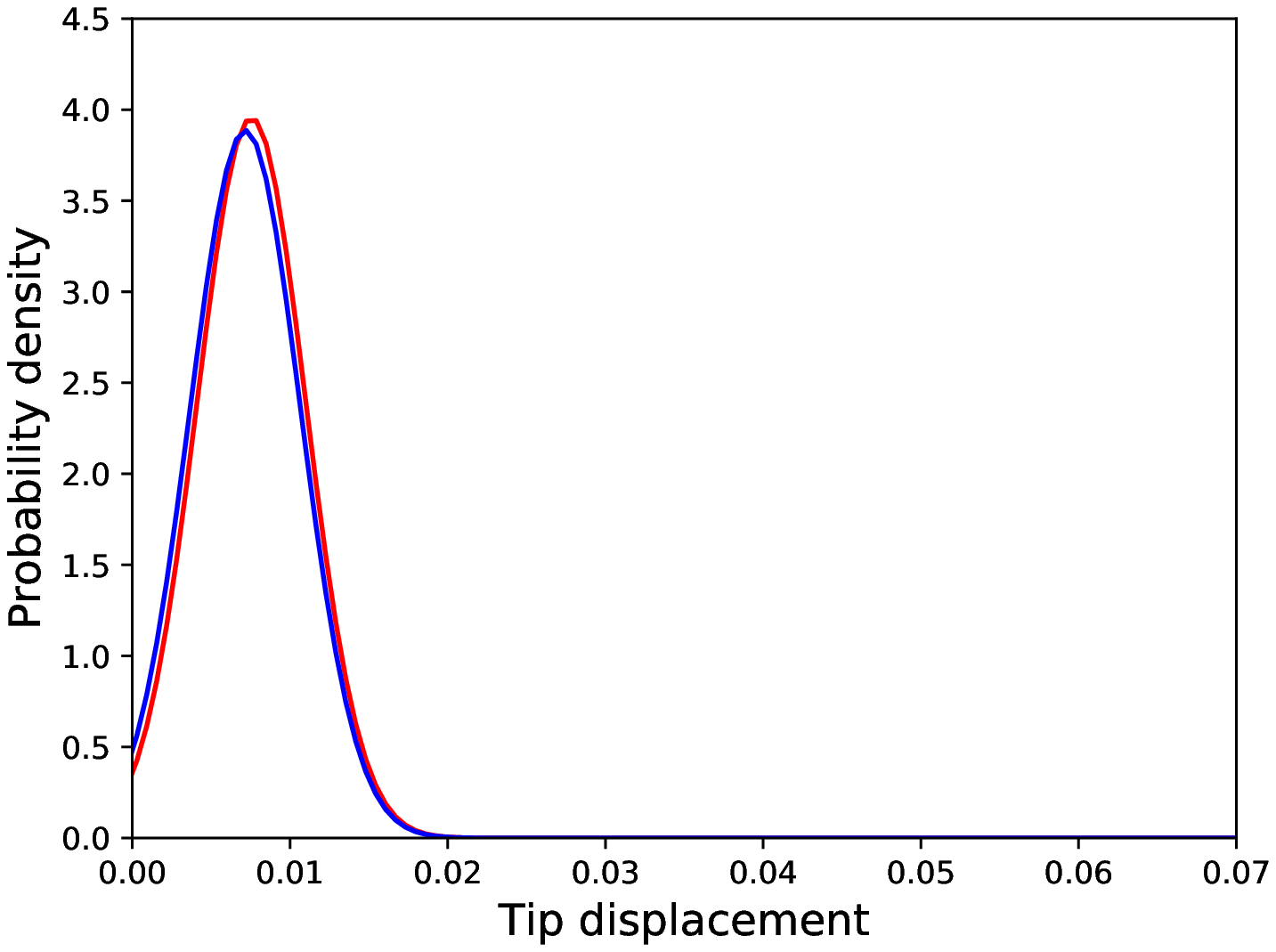}
    \caption{}
    \label{fig:sub_b3}
    \end{subfigure}
    \begin{subfigure}[b]{0.49\linewidth}
    \centering
    \includegraphics[width=0.99\linewidth]{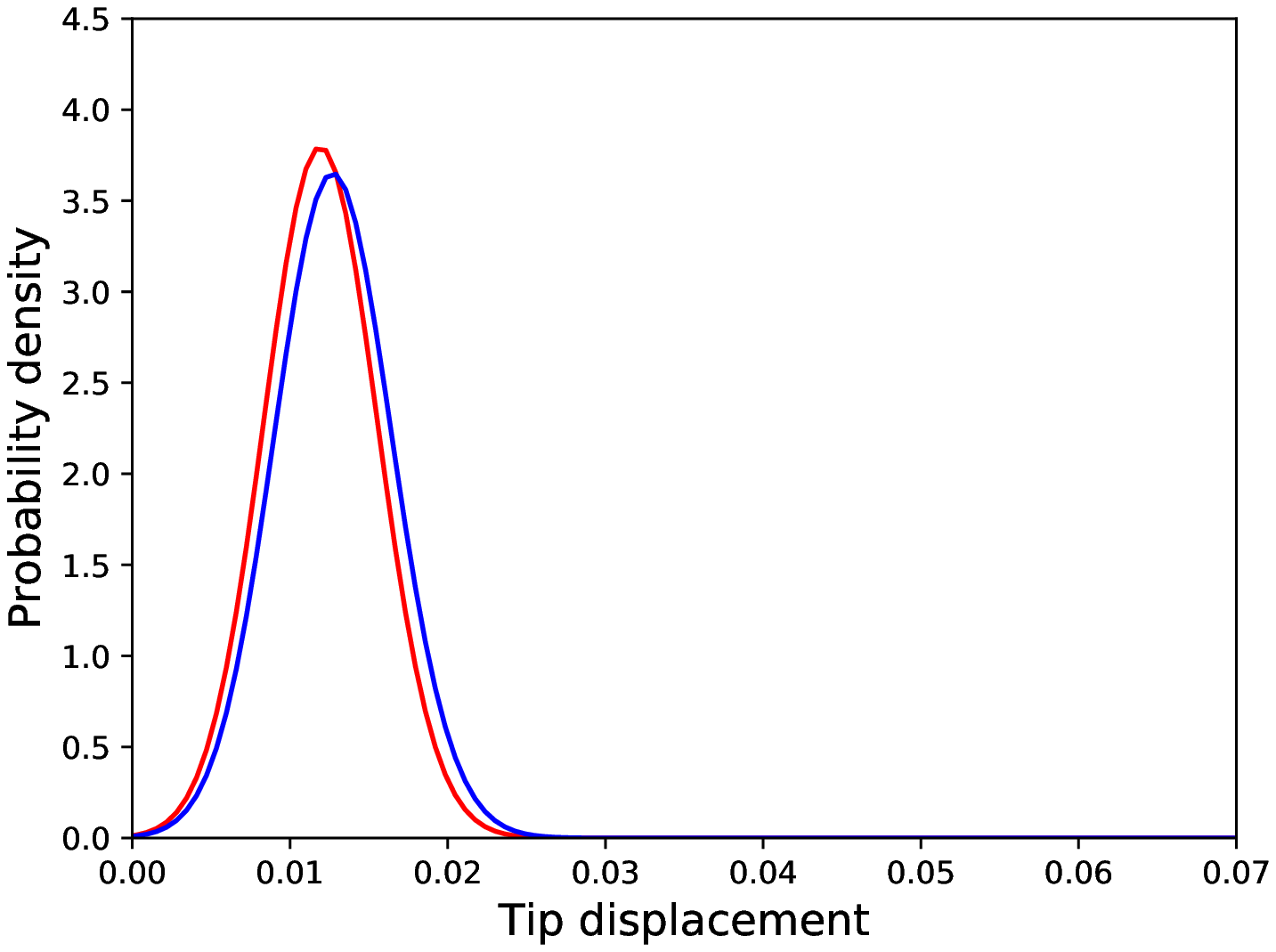}
    \caption{}
    \label{fig:sub_c3}
    \end{subfigure}
    \begin{subfigure}[b]{0.49\linewidth}
    \centering
    \includegraphics[width=0.99\linewidth]{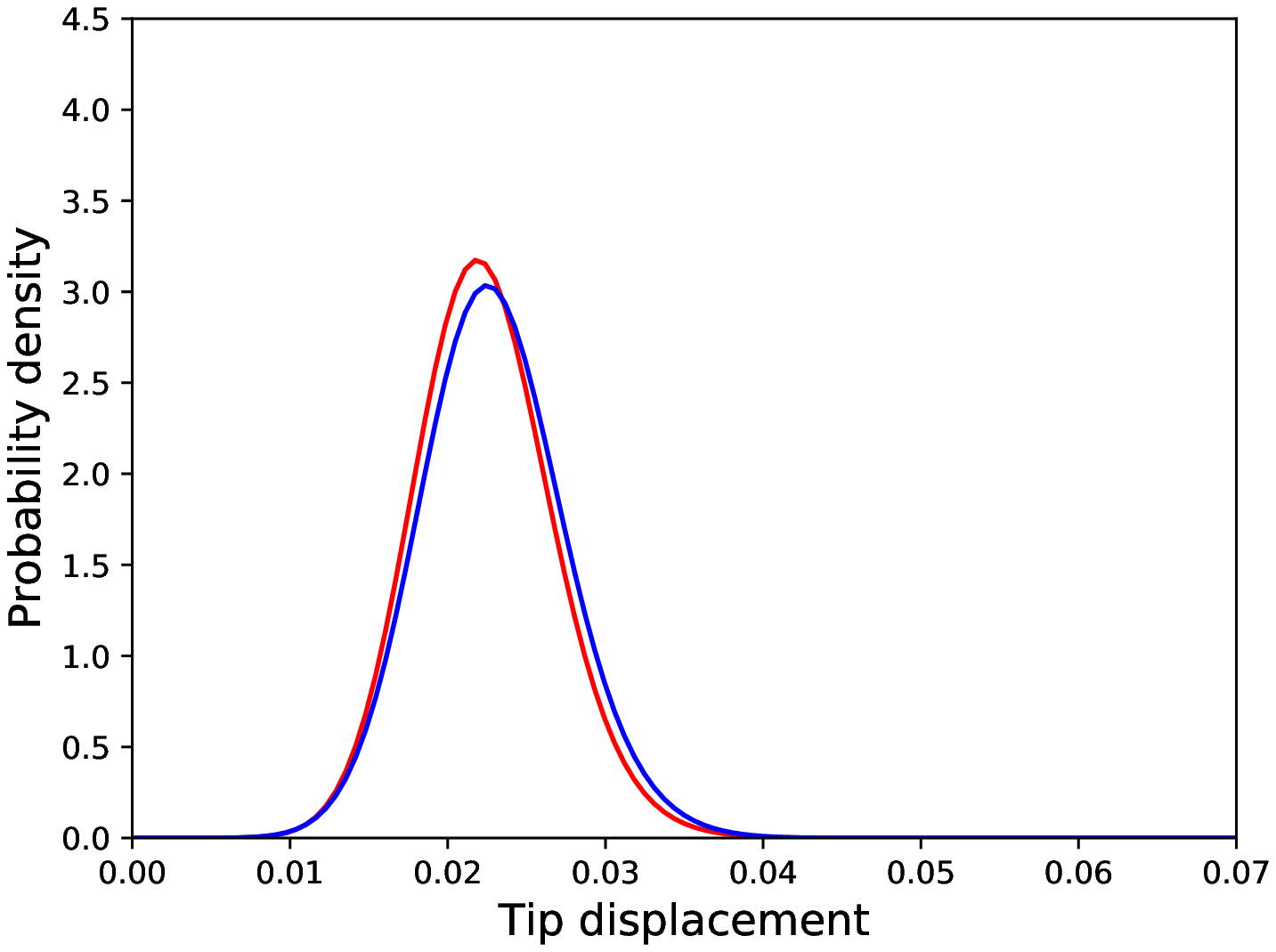}
    \caption{}
    \label{fig:sub_d3}
    \end{subfigure}
    \begin{subfigure}[b]{0.49\linewidth}
    \centering
    \includegraphics[width=0.99\linewidth]{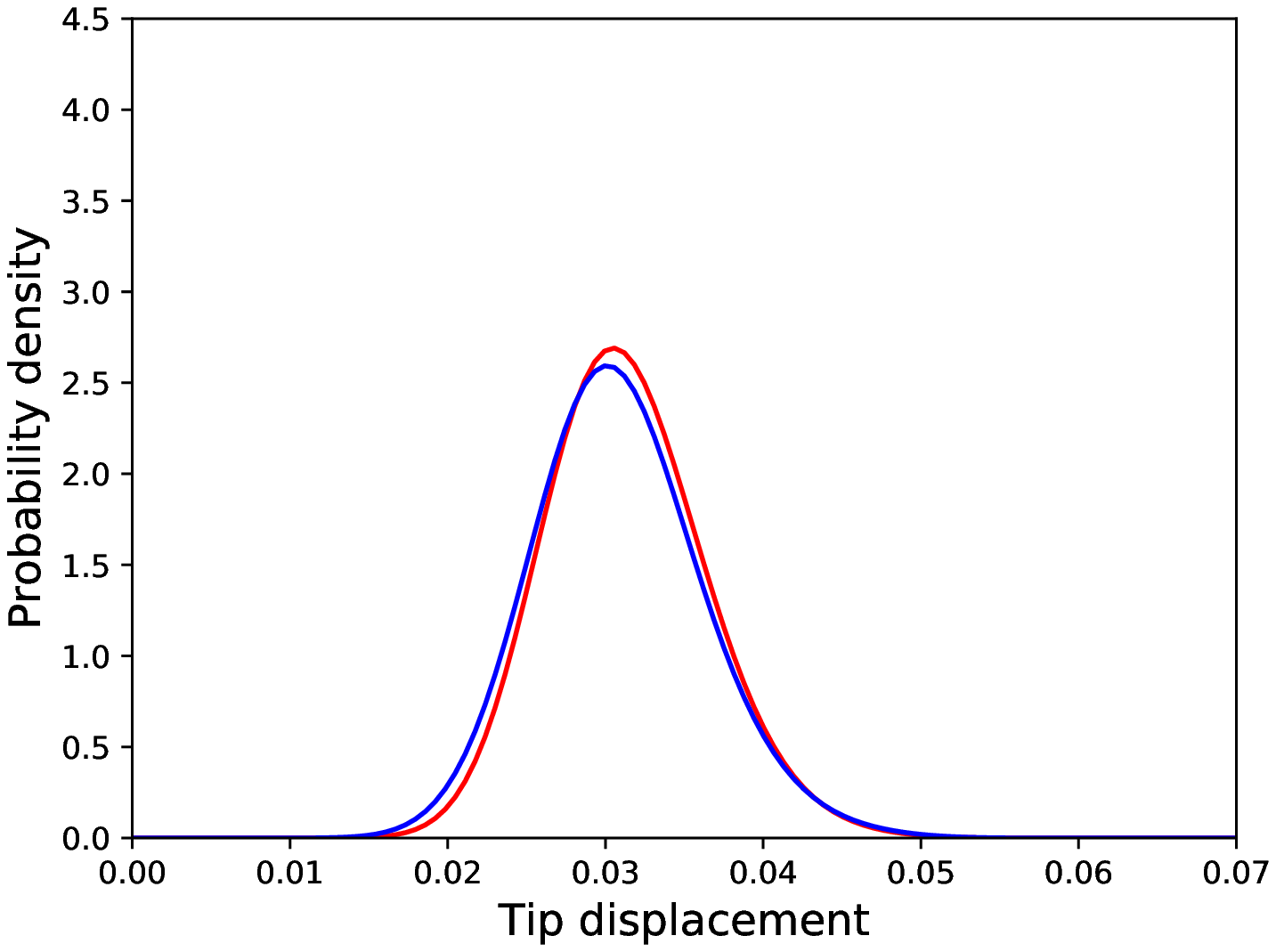}
    \caption{}
    \label{fig:sub_e3}
    \end{subfigure}
    \begin{subfigure}[b]{0.49\linewidth}
    \centering
    \includegraphics[width=0.99\linewidth]{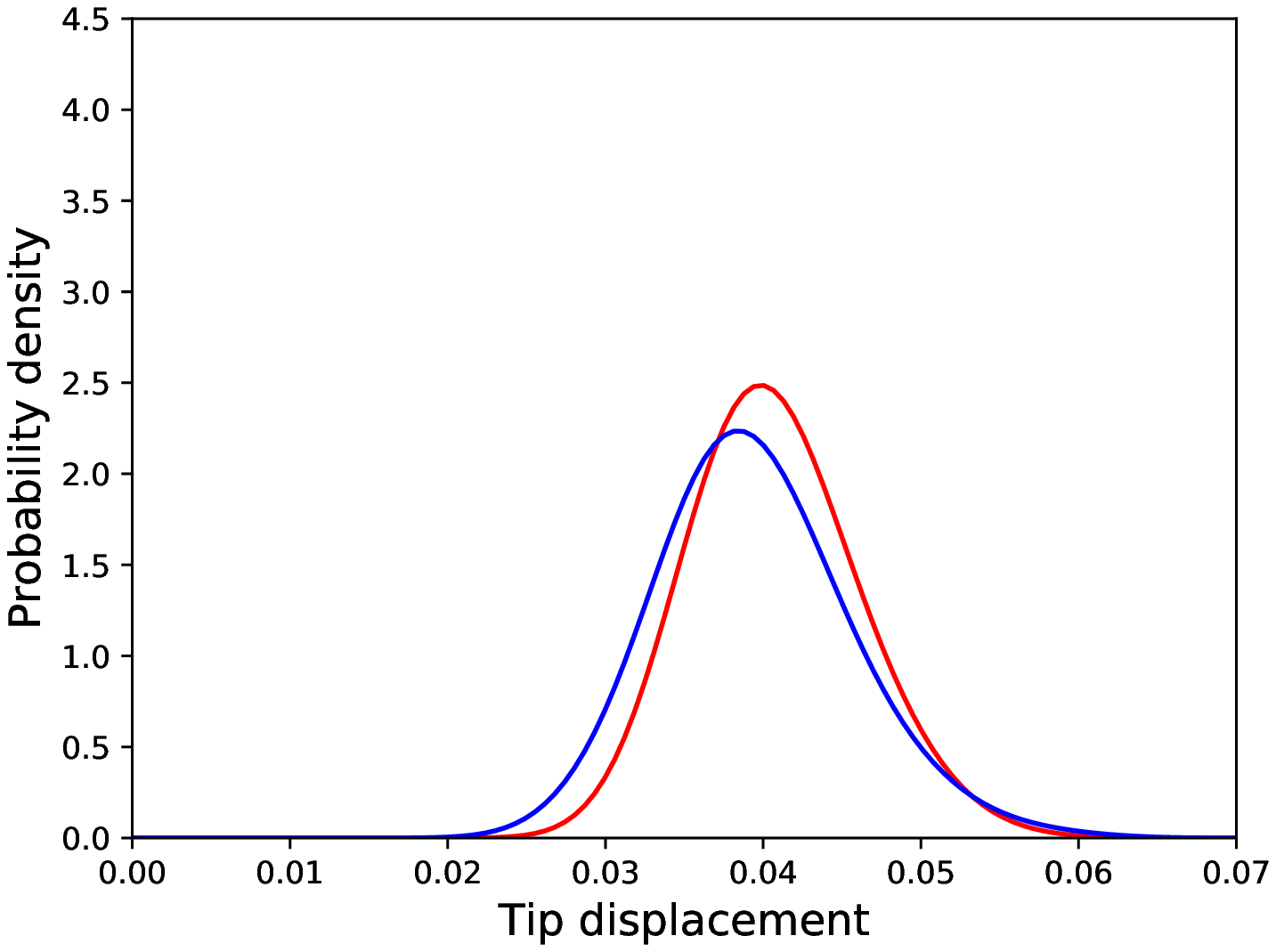}
    \caption{}
    \label{fig:sub_f3}
    \end{subfigure}
    \caption{Distributions of tip displacements corresponding to Monte Carlo samples (blue) and generated samples by the cGAN-SFEM hybrid approach (red) regarding the nonlinear problem, for different load cases; (a) 30 load units, (b) 90 load units, (c) 150 load units, (d) 240 load units, (e) 300 load units and (f) 360 load units.}
    \label{fig:cGAN_SFEM_nonlinear_testing_distributions}
\end{figure}

\begin{figure}[!htbp]
    \centering
    \includegraphics[scale=0.5]{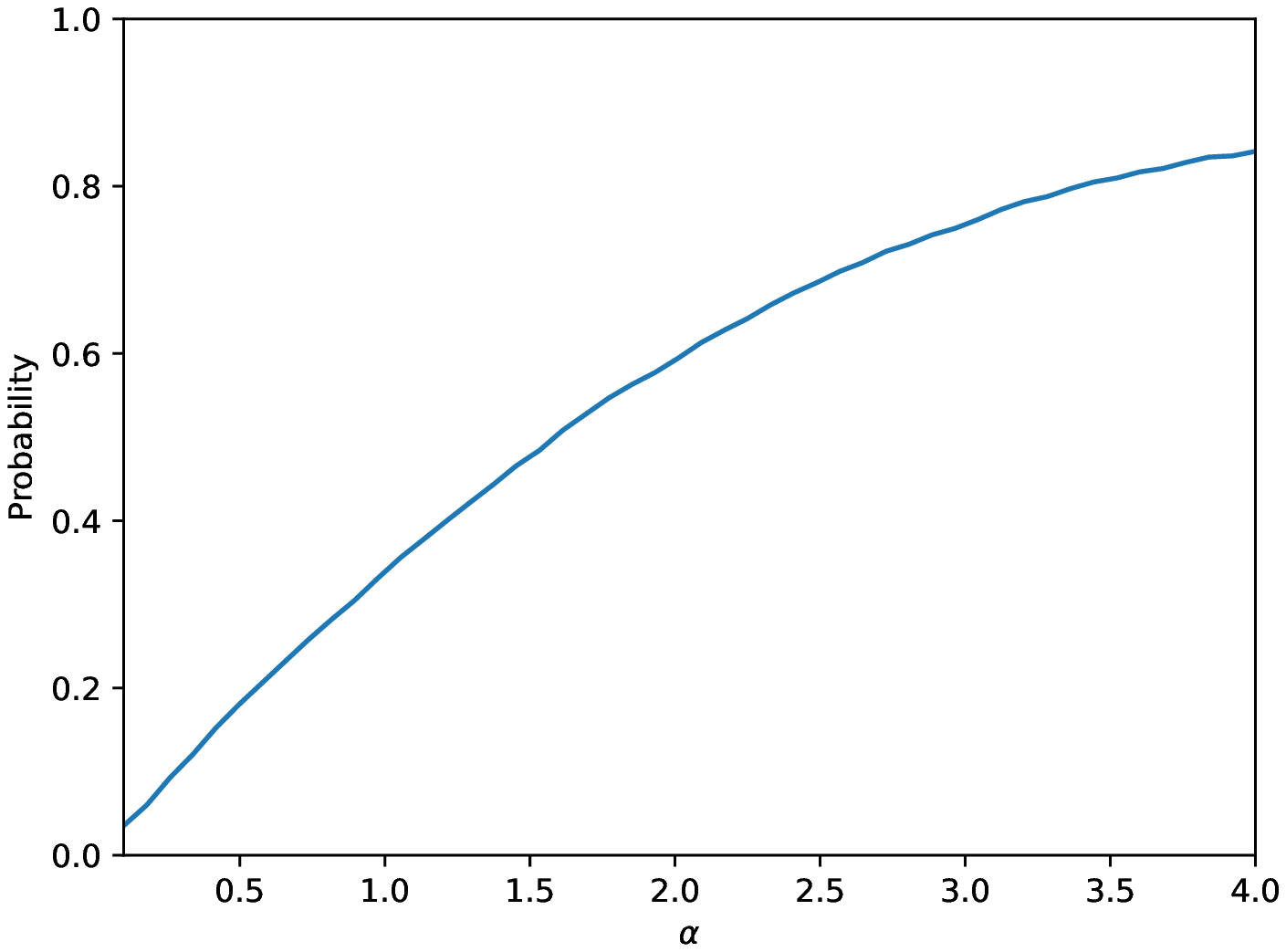}
    \caption{Probability defined in equation (\ref{eq:alpha_mirror_prob}) as a function of the parameter $\alpha$ for the cGAN-SFEM hybrid model applied on the nonlinear cantilever case study.}
    \label{fig:alpha_probs_cGAN_hybrid_nonlinear}
\end{figure}

The same training procedure was followed. The hidden layer sizes considered were in the set $\{10, 20, 30, ... 1500\}$. The model was trained given codes and data of the training dataset. The model that yielded the lowest KL divergence on the validation dataset had $1000$ neurons in its hidden layer and was selected as the best model. The aforementioned model was tested on the testing dataset and the results for a selection of the input loads are presented in Figure \ref{fig:cGAN_SFEM_nonlinear_testing_distributions}. The average KL divergence on the validation dataset and on the testing dataset was $0.049$ and $0.044$ respectively, meaning that the predicted and the real distributions were quite close; the maximum value of KL divergence was $0.22$, which will be the value of $\epsilon$, if the hybrid model is to be considered an $\epsilon$-mirror. Regarding its ability to serve as an $\alpha$-mirror, Figure \ref{fig:alpha_probs_cGAN_hybrid_nonlinear} shows the probability of the observation falling into the interval defined by equation (\ref{eq:alpha_mirror_prob}) as a function of $\alpha$.

\subsection{Extrapolation capability study}
\label{sec:extrapolation_capabilities}

As already mentioned, the greatest advantage of physics-based models is that they can provide accurate predictions even for data coming from a different domain than the one used in order to calibrate them. This advantage is based on the definition of the physics of the model, that fit sufficiently the physical phenomena, which the model is called to explain. On the other hand, data-based models tend to have no such capabilities. If one has trained a machine learning model on a specific domain of the data, use of the model outside this domain (extrapolation) should be avoided.

When a hybrid model is built, inheritance of the extrapolation capabilities of the physics-based part of the model is desired. In the current application, the output of the hybrid model is controlled by the input code to the model and also by the distribution output of the physics-based SFE model. The SFE model's physics do not suffice in order to explain the nonlinear behaviour of the material of the cantilever. Sometimes, even the physics of the underlying problem might not be clear making the use of a white-box model even more unfavorable. However, a part of the physics of the linear SFE model herein applies also in the nonlinear case. Specifically, that higher values of input load correspond to higher values of tip displacements and higher values of standard deviation (as shown in Figure
\ref{fig:distributions_comparison}). Therefore, in the current subsection the potential extrapolation capabilities of the hybrid model are studied and compared with the extrapolation capabilities of the black-box model.

A first step, that can be made in order to increase the extrapolation capabilities of the black-box model (and therefore also of the hybrid model), is to redefine the normalisation scheme of the input and output values of the neural networks used herein. The approach that has been followed so far is to scale the input and the outputs onto the interval $[-1, 1]$. This technique brings all the values within the effective range of the sigmoid functions used (hyperbolic tangent herein) and allows proper gradients to be calculated and the networks to be properly trained using the back-propagation process. However, normalisation on the interval $[-1, 1]$ most probably saturates the sigmoids for the extreme values of the inputs and the outputs and minimises the extrapolation capabilities of the models. To bypass such an issue, in the current section, the scaling is performed having as target the interval $[-0.8, 0.8]$. If one knows that a model might be called to perform outside its training domain, this scaling strategy is a viable option to increase the extrapolation capability of the model.

The initial dataset is split in two in order to compare the extrapolation potential of the hybrid and the black-box models. The first subset includes data corresponding to load units $\{10, 20... 310\}$ and the second subset to load units $\{320, 330... 400\}$. Using the first subset a hybrid model and a black-box model were trained; i.e. the dataset was split in training, validation and testing datasets and the same procedure as before was followed. After training, the models were also tested on the second subset, which is outside the training domain of the models, in order to evaluate their extrapolation capabilities.

\begin{figure}[!htbp]
    \centering
    \begin{subfigure}[b]{0.49\linewidth}
    \centering
    \includegraphics[width=0.99\linewidth]{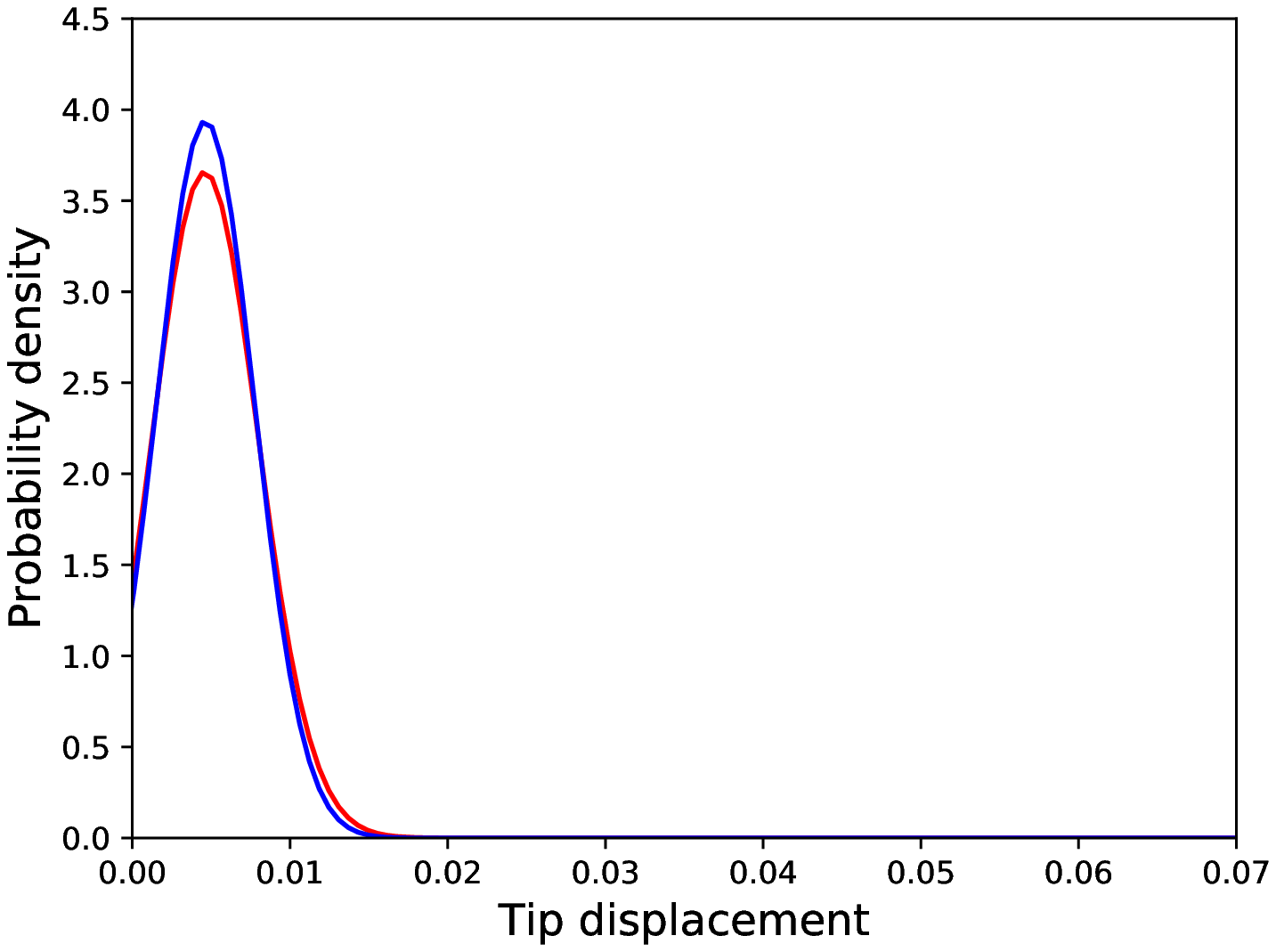}
    \caption{}
    \label{fig:sub_a3}
    \end{subfigure}
    \begin{subfigure}[b]{0.49\linewidth}
    \centering
    \includegraphics[width=0.99\linewidth]{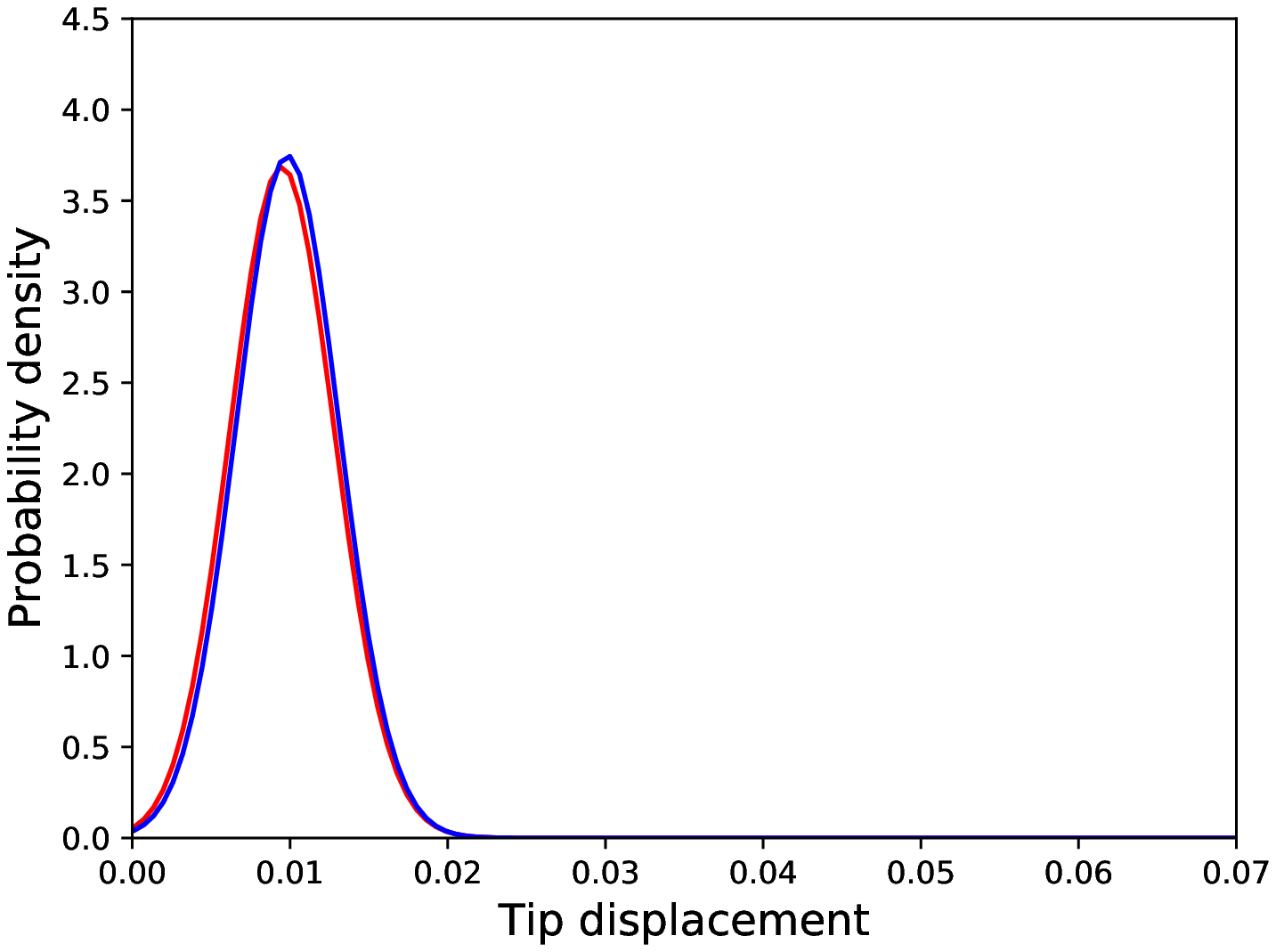}
    \caption{}
    \label{fig:sub_b3}
    \end{subfigure}
    \begin{subfigure}[b]{0.49\linewidth}
    \centering
    \includegraphics[width=0.99\linewidth]{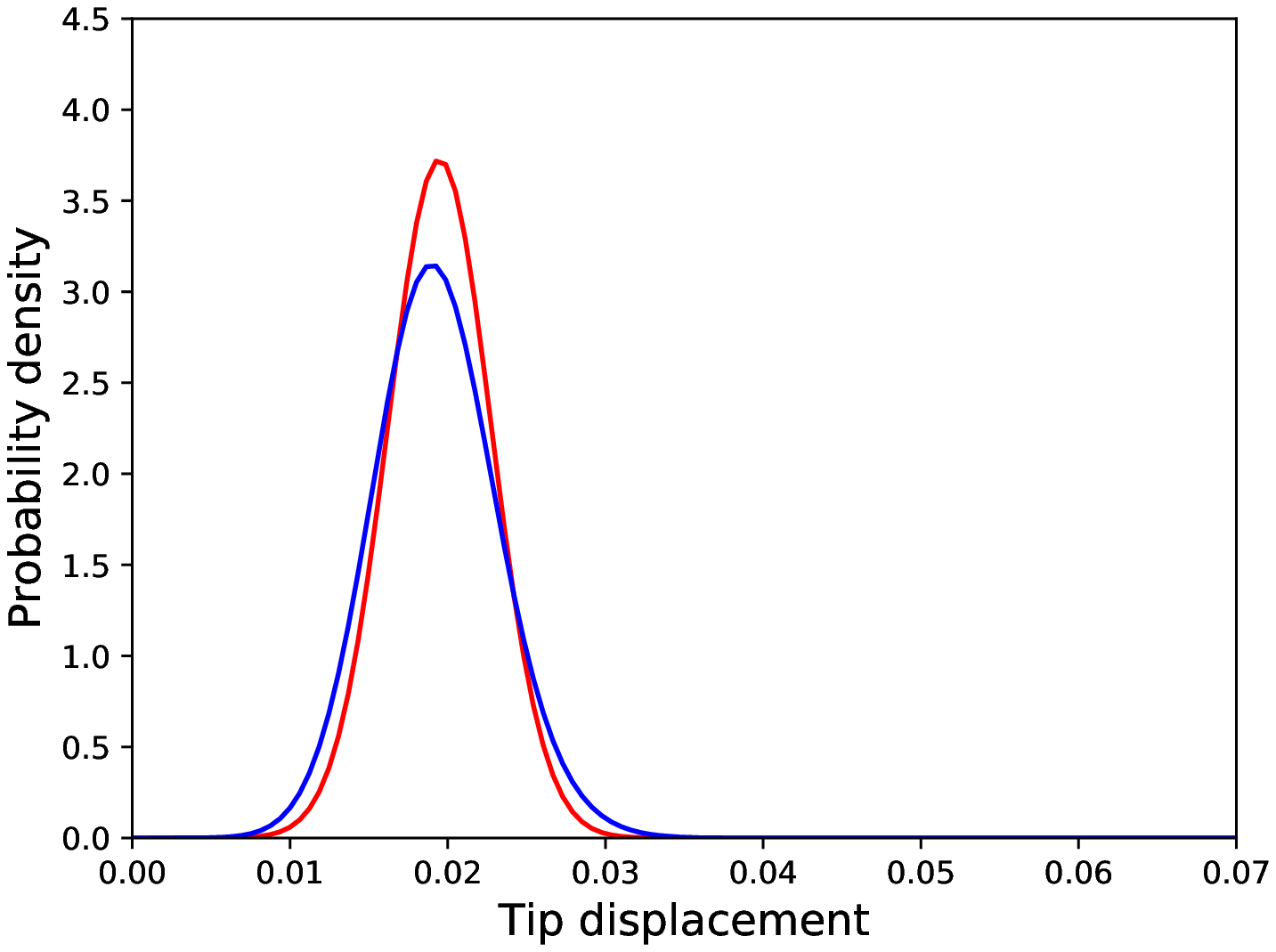}
    \caption{}
    \label{fig:sub_c3}
    \end{subfigure}
    \begin{subfigure}[b]{0.49\linewidth}
    \centering
    \includegraphics[width=0.99\linewidth]{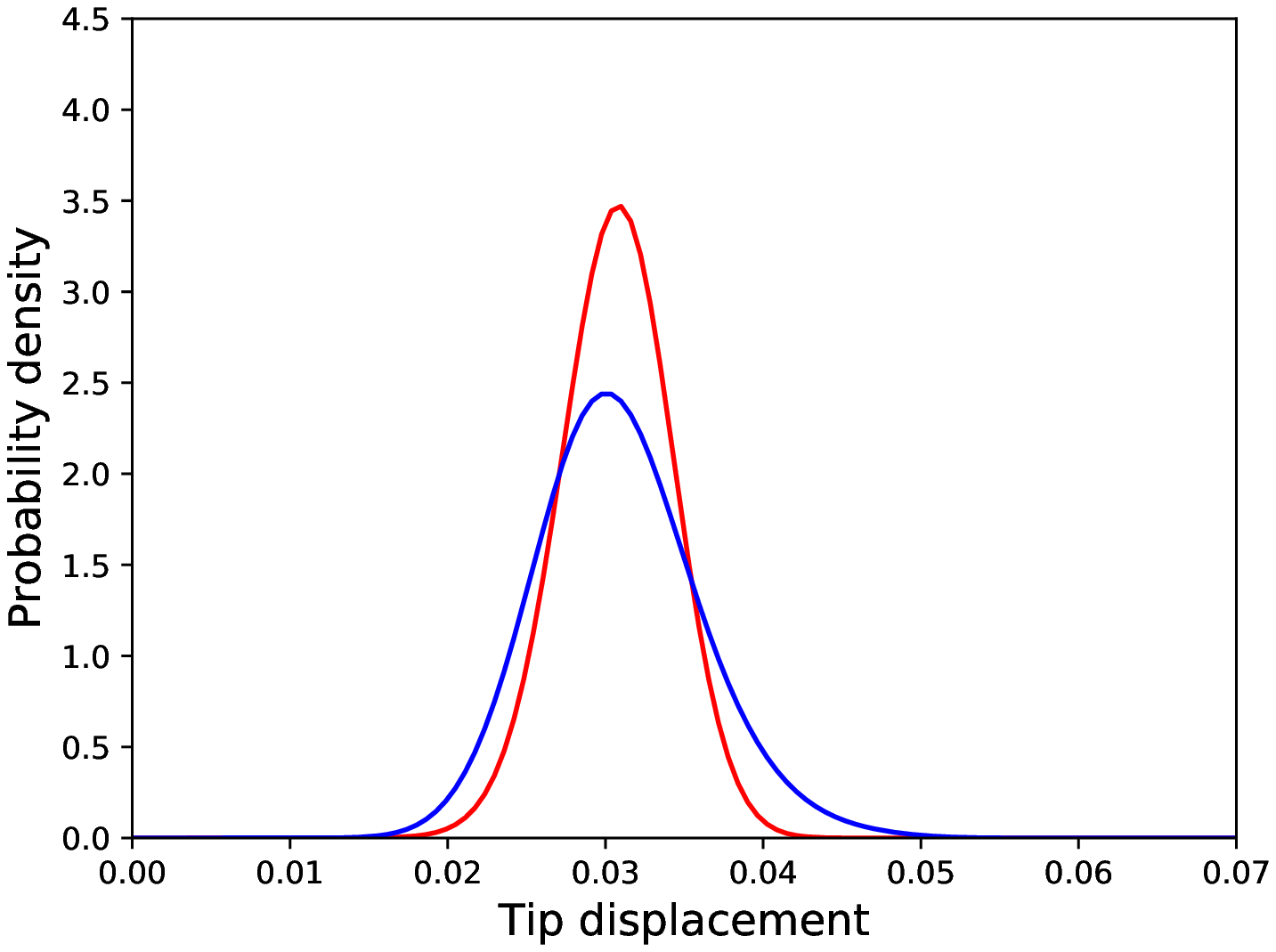}
    \caption{}
    \label{fig:sub_d3}
    \end{subfigure}
    \caption{Distributions of tip displacements corresponding to Monte Carlo samples (blue) and generated samples by the cGAN model (red) regarding the nonlinear problem, trained according to the reduced dataset for different load cases; (a) 60 load units, (b) 120 load units, (c) 210 load units, (d) 300 load units.}
    \label{fig:cGAN_small_dataset_distribs}
\end{figure}

\begin{figure}[!htbp]
    \centering
    \begin{subfigure}[b]{0.49\linewidth}
    \centering
    \includegraphics[width=0.99\linewidth]{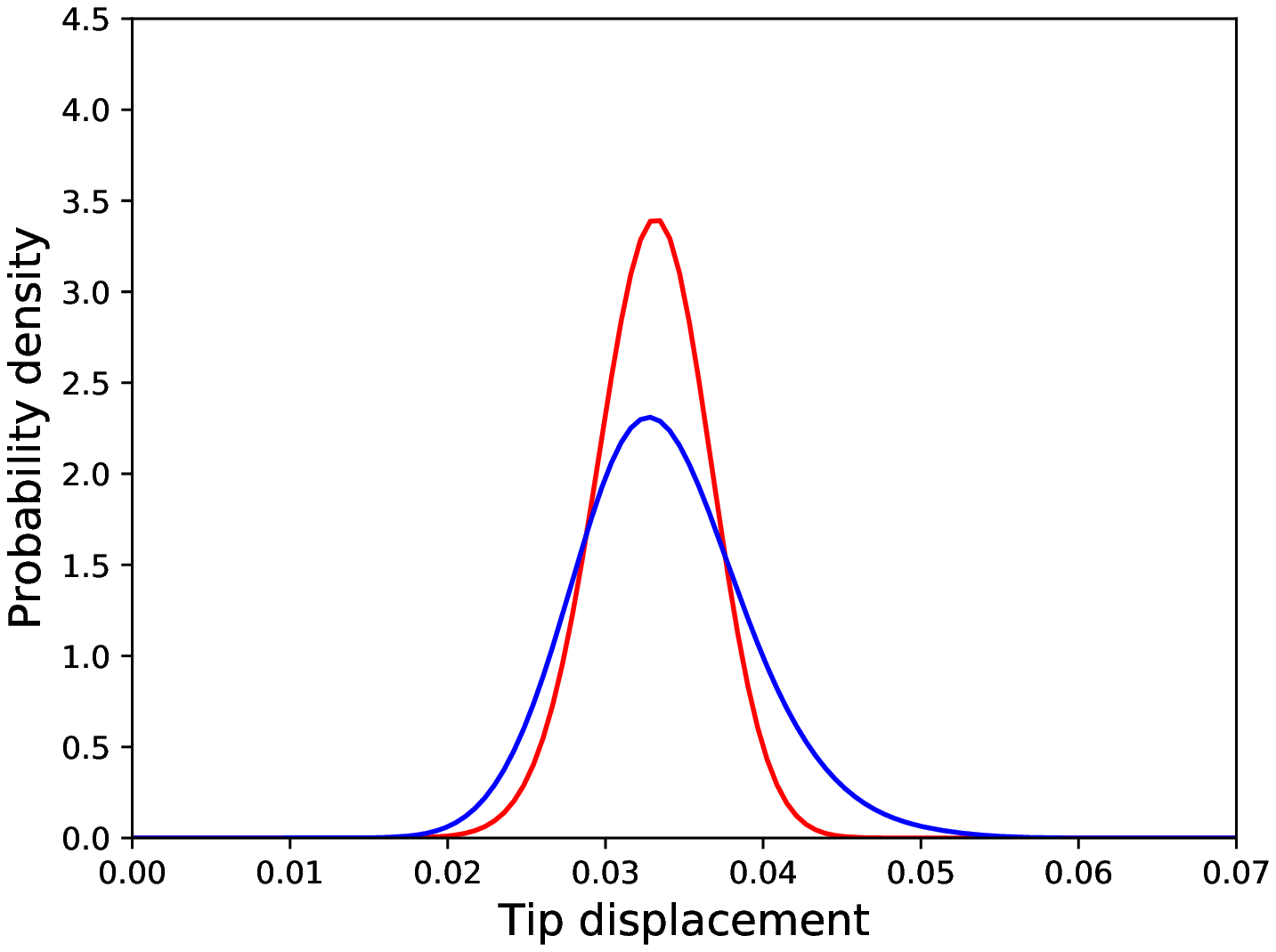}
    \caption{}
    \label{fig:sub_a3}
    \end{subfigure}
    \begin{subfigure}[b]{0.49\linewidth}
    \centering
    \includegraphics[width=0.99\linewidth]{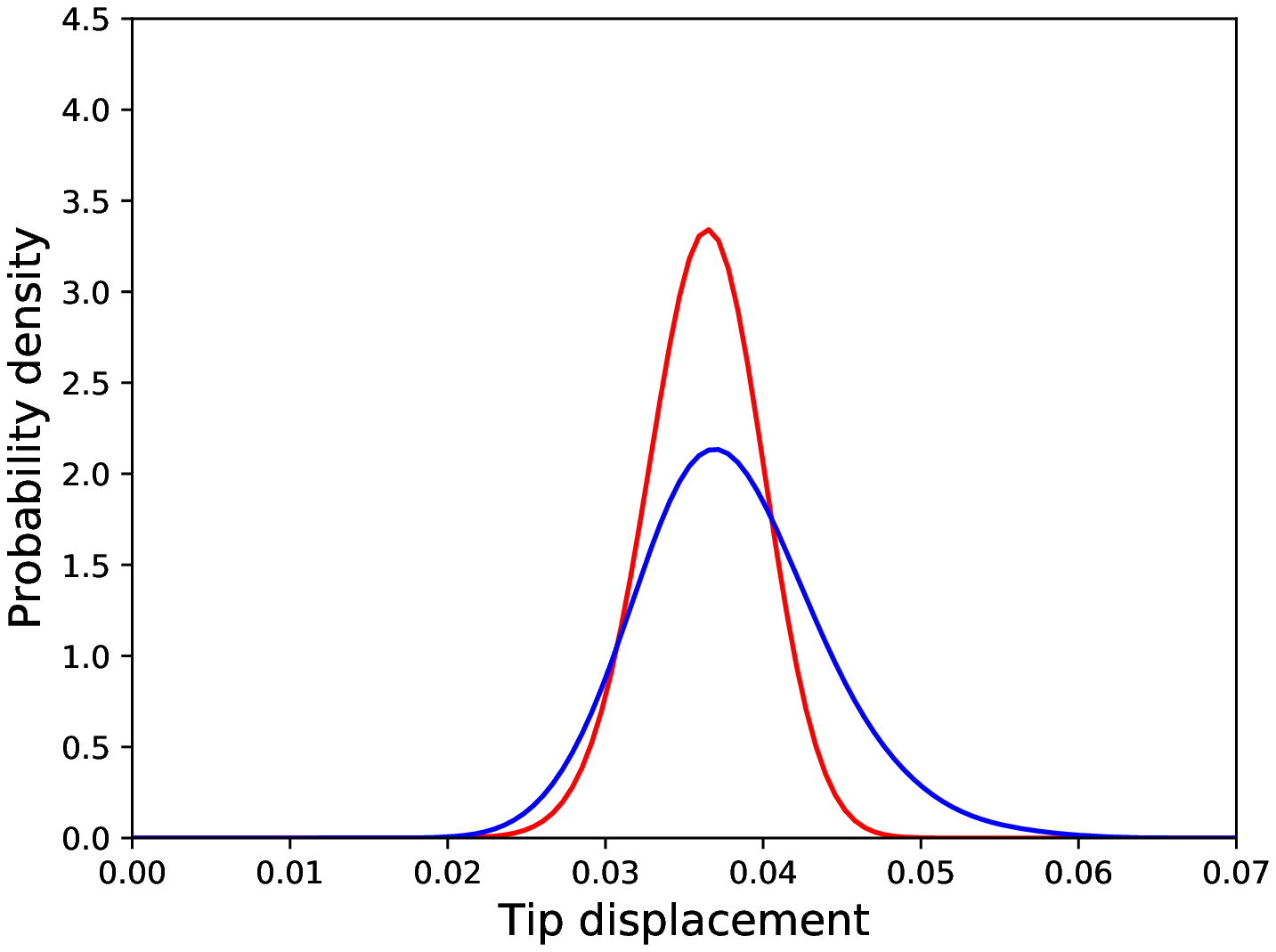}
    \caption{}
    \label{fig:sub_b3}
    \end{subfigure}
    \begin{subfigure}[b]{0.49\linewidth}
    \centering
    \includegraphics[width=0.99\linewidth]{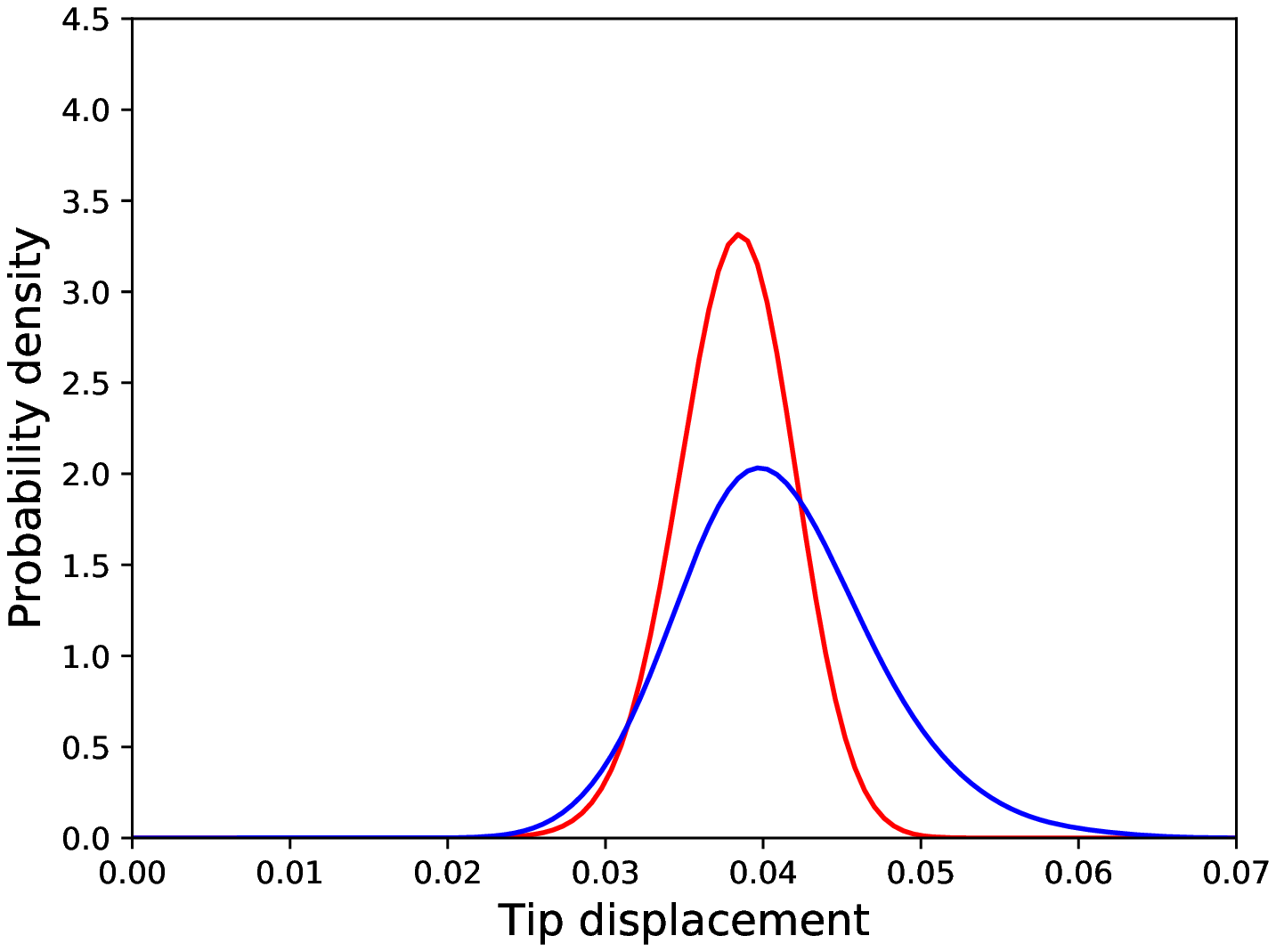}
    \caption{}
    \label{fig:sub_c3}
    \end{subfigure}
    \begin{subfigure}[b]{0.49\linewidth}
    \centering
    \includegraphics[width=0.99\linewidth]{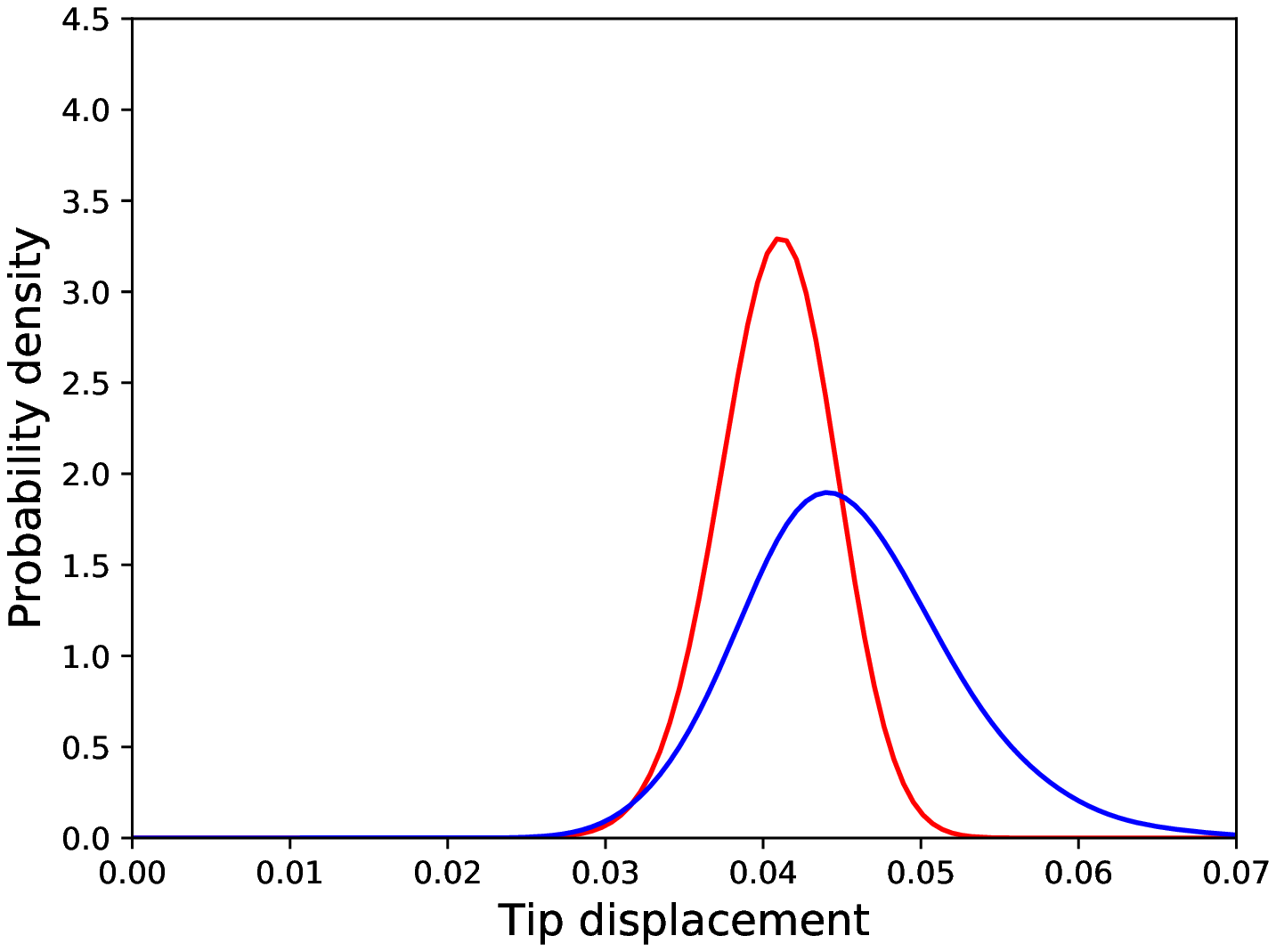}
    \caption{}
    \label{fig:sub_d3}
    \end{subfigure}
    \caption{Distributions of tip displacements corresponding to Monte Carlo samples (blue) and generated samples by the cGAN model (red) regarding the `extrapolation dataset' for the nonlinear problem for different load cases; (a) 320 load units, (b) 350 load units, (c) 370 load units, (d) 400 load units.}
    \label{fig:extrapolation_cGAN}
\end{figure}

The best black-box model had in terms of KL divergence on the validation dataset had $1000$ neurons in its hidden layer. It yielded average KL divergence equal to $0.048$ on the validation dataset and $0.054$ on the testing dataset. A distribution comparison between the predicted and the real distributions for loads of the testing dataset is shown in Figure \ref{fig:cGAN_small_dataset_distribs}. The results are quite close to the ones observed in the application on the complete dataset. In order to see how the model extrapolates, the same model was tested on the second subset of loads (load units $320, 330... 400$), which is outside the training domain. Some distributions are shown in Figure \ref{fig:extrapolation_cGAN} and the average KL divergence for the `extrapolation dataset' is equal to $0.77$. It is clear that outside the training domain, the model does not perform well and as the load increases, the distribution moves only slightly towards higher values of displacements.

\begin{figure}[!htbp]
    \centering
    \begin{subfigure}[b]{0.49\linewidth}
    \centering
    \includegraphics[width=0.99\linewidth]{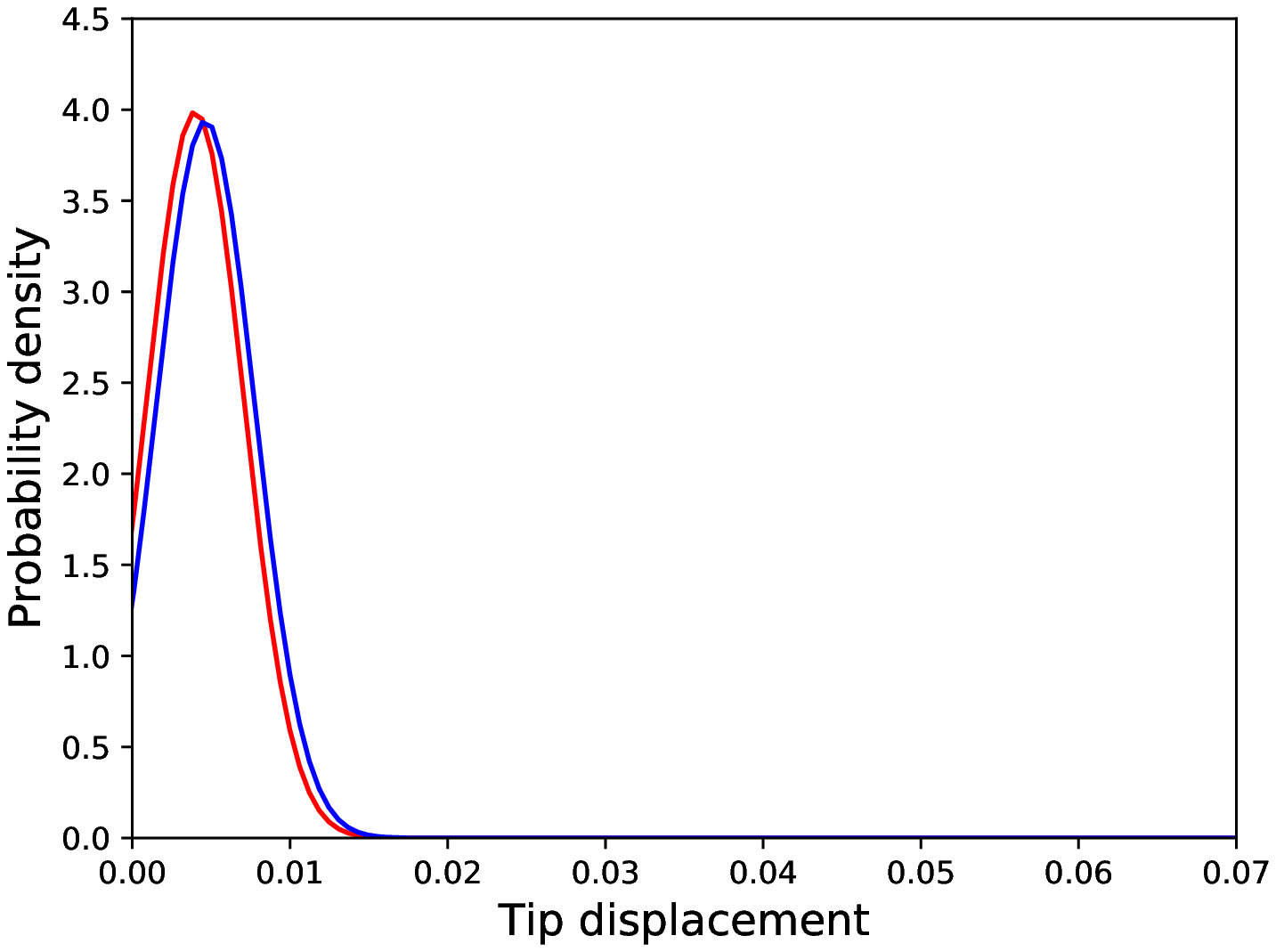}
    \caption{}
    \label{fig:sub_a3}
    \end{subfigure}
    \begin{subfigure}[b]{0.49\linewidth}
    \centering
    \includegraphics[width=0.99\linewidth]{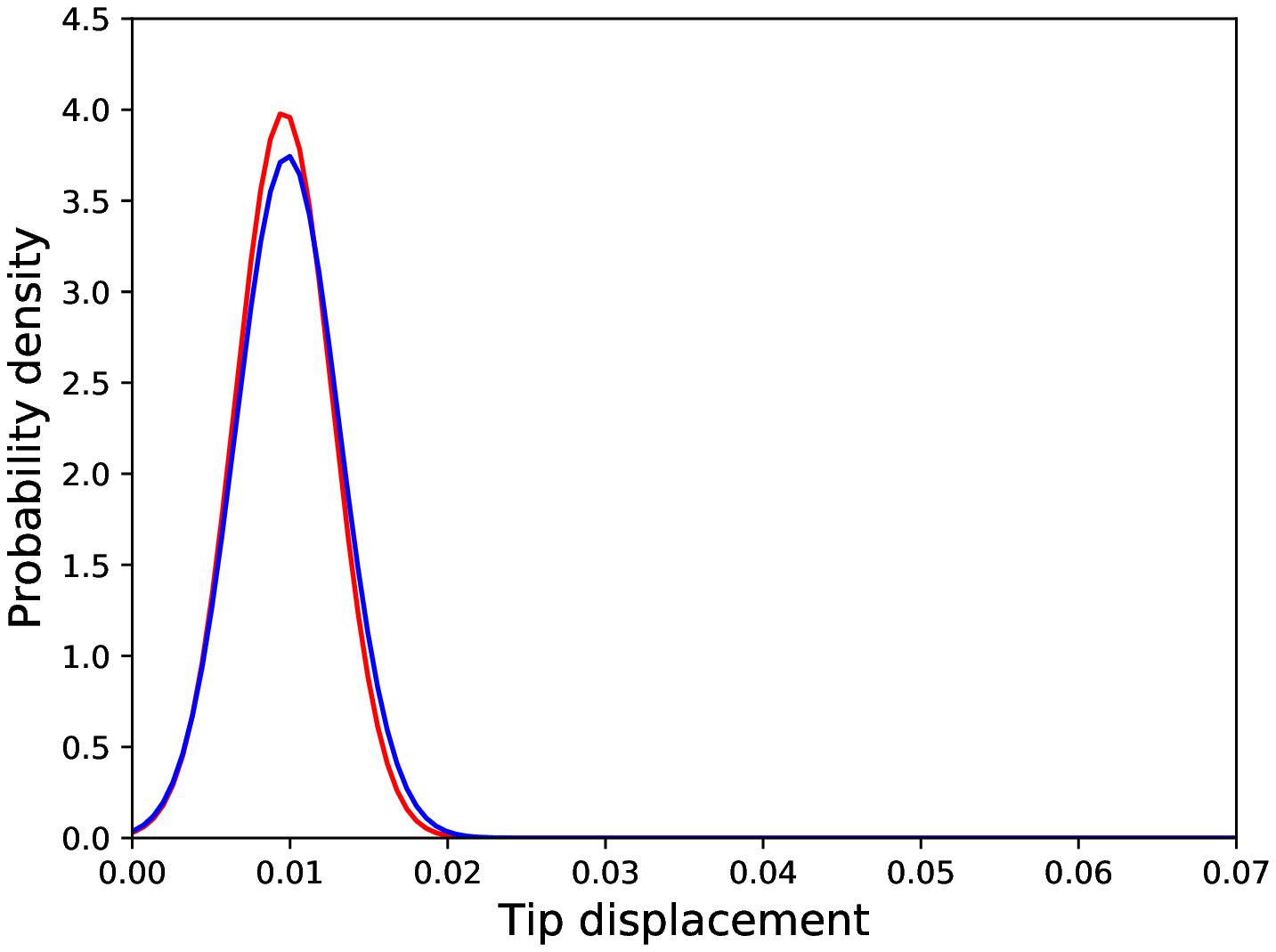}
    \caption{}
    \label{fig:sub_b3}
    \end{subfigure}
    \begin{subfigure}[b]{0.49\linewidth}
    \centering
    \includegraphics[width=0.99\linewidth]{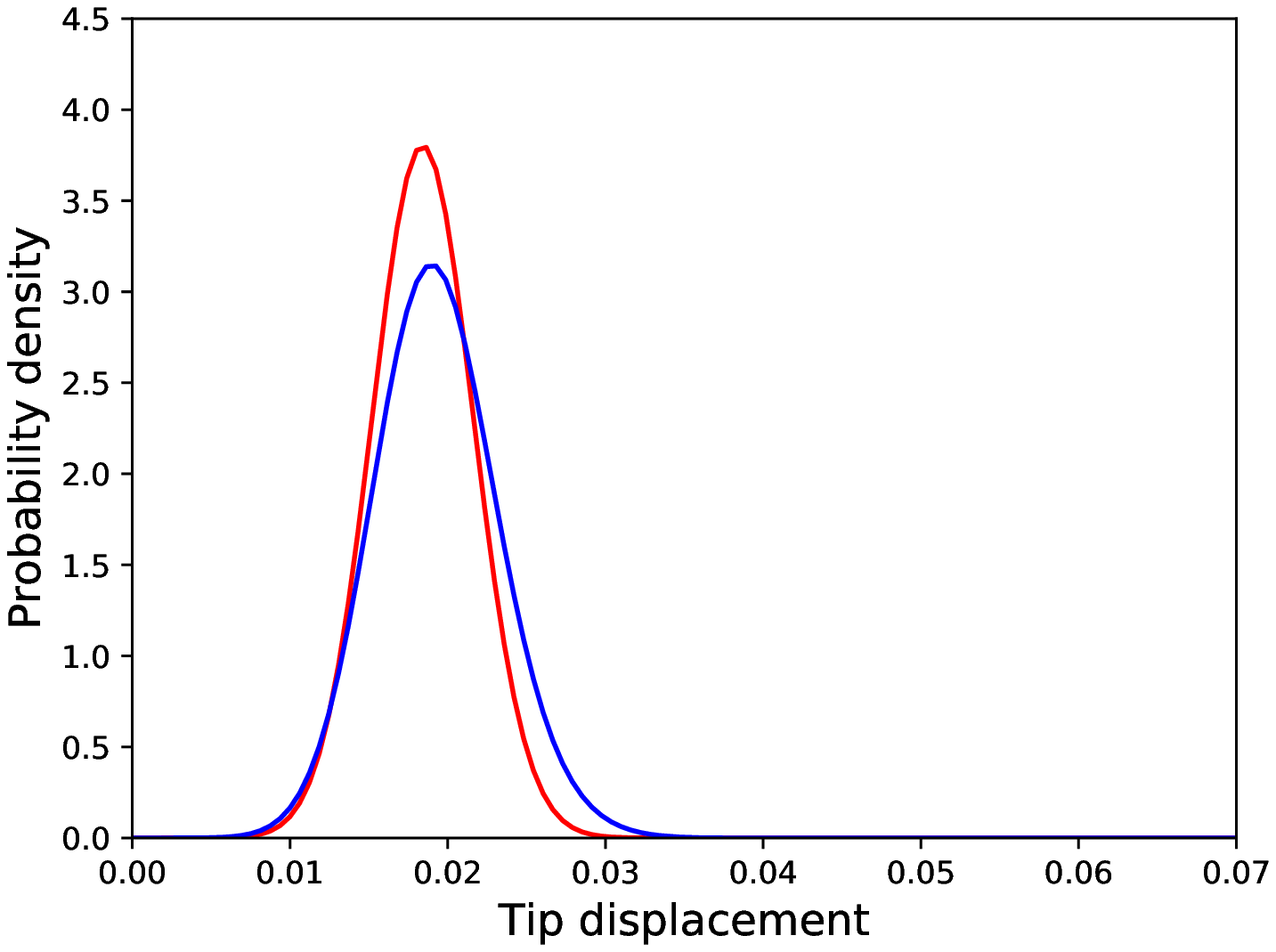}
    \caption{}
    \label{fig:sub_c3}
    \end{subfigure}
    \begin{subfigure}[b]{0.49\linewidth}
    \centering
    \includegraphics[width=0.99\linewidth]{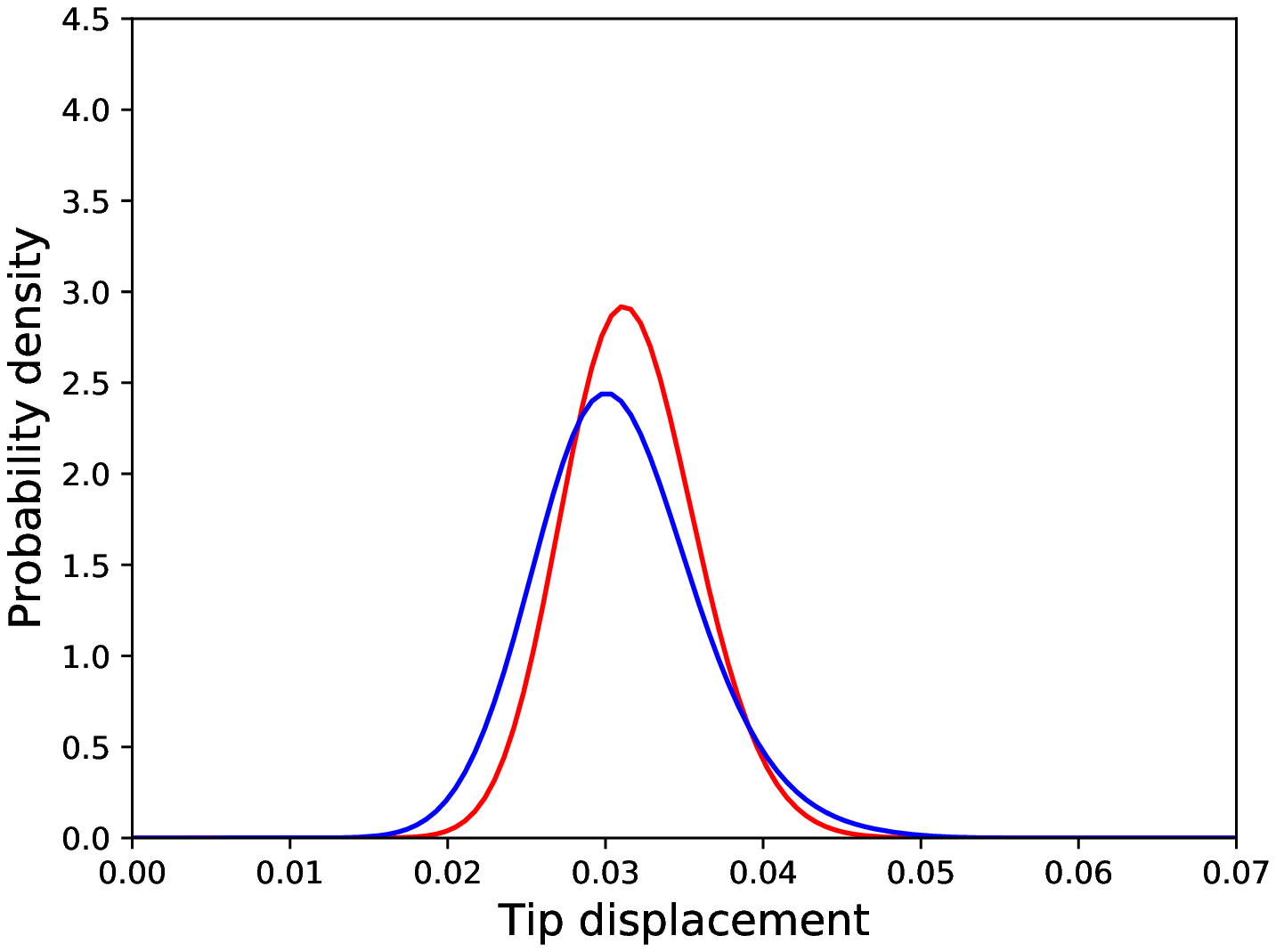}
    \caption{}
    \label{fig:sub_d3}
    \end{subfigure}
    \caption{Distributions of tip displacements corresponding to Monte Carlo samples (blue) and generated samples by the hybrid model (red) regarding the nonlinear problem, trained according to the reduced dataset for different load cases; (a) 60 load units, (b) 120 load units, (c) 210 load units, (d) 300 load units.}
    \label{fig:hybrid_small_dataset_distribs}
\end{figure}

\begin{figure}[!htbp]
    \centering
    \begin{subfigure}[b]{0.49\linewidth}
    \centering
    \includegraphics[width=0.99\linewidth]{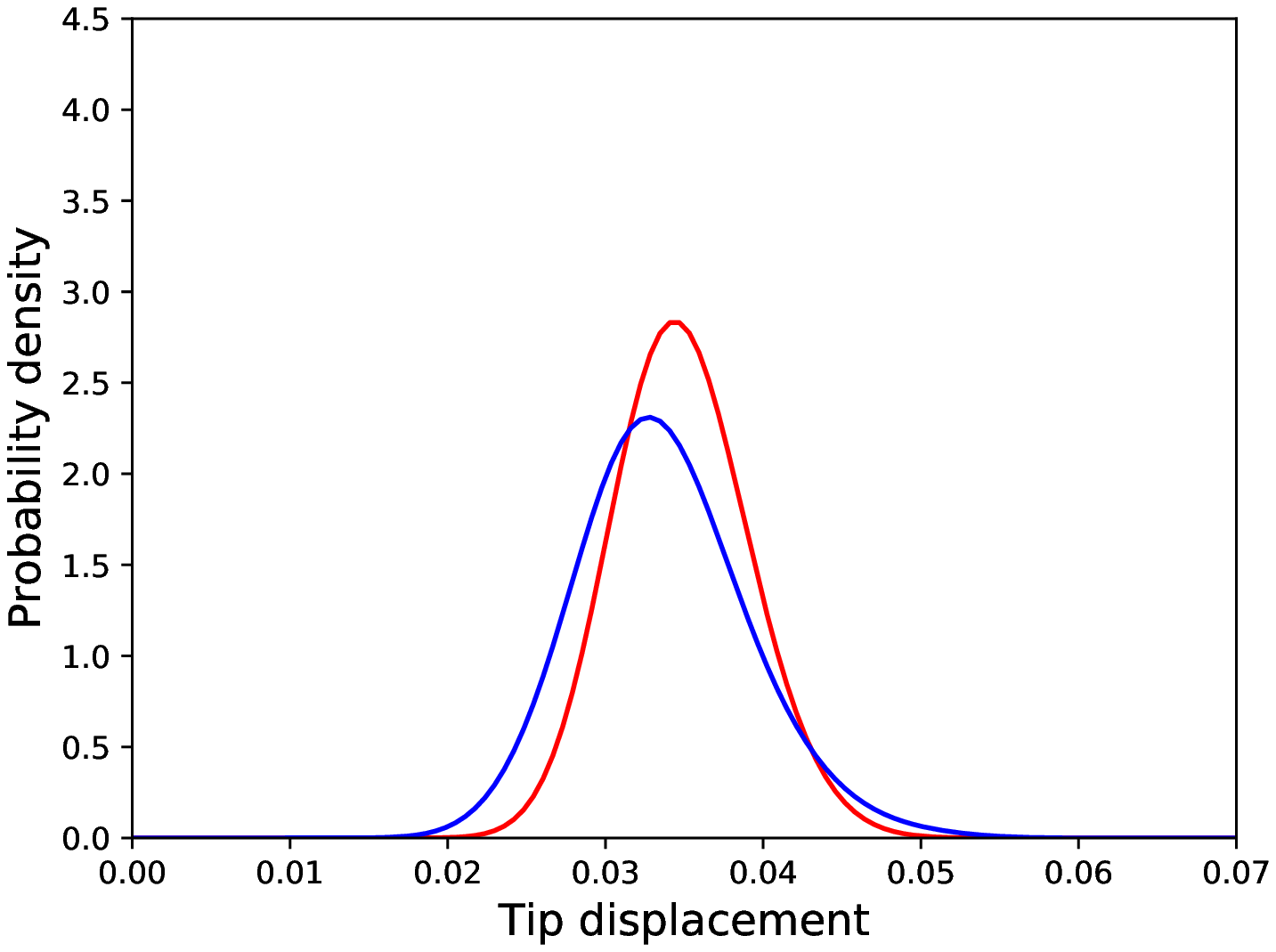}
    \caption{}
    \label{fig:sub_a3}
    \end{subfigure}
    \begin{subfigure}[b]{0.49\linewidth}
    \centering
    \includegraphics[width=0.99\linewidth]{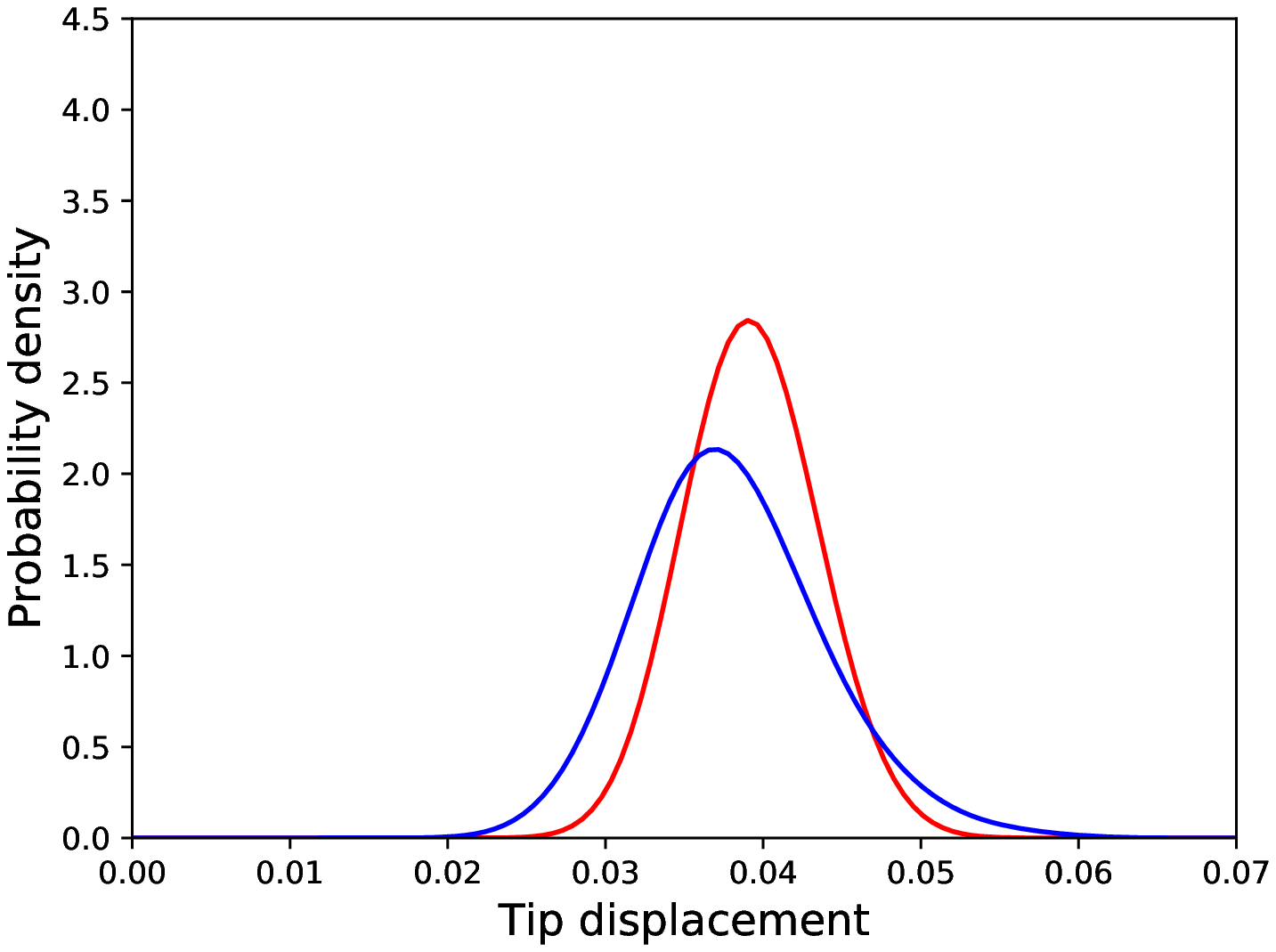}
    \caption{}
    \label{fig:sub_b3}
    \end{subfigure}
    \begin{subfigure}[b]{0.49\linewidth}
    \centering
    \includegraphics[width=0.99\linewidth]{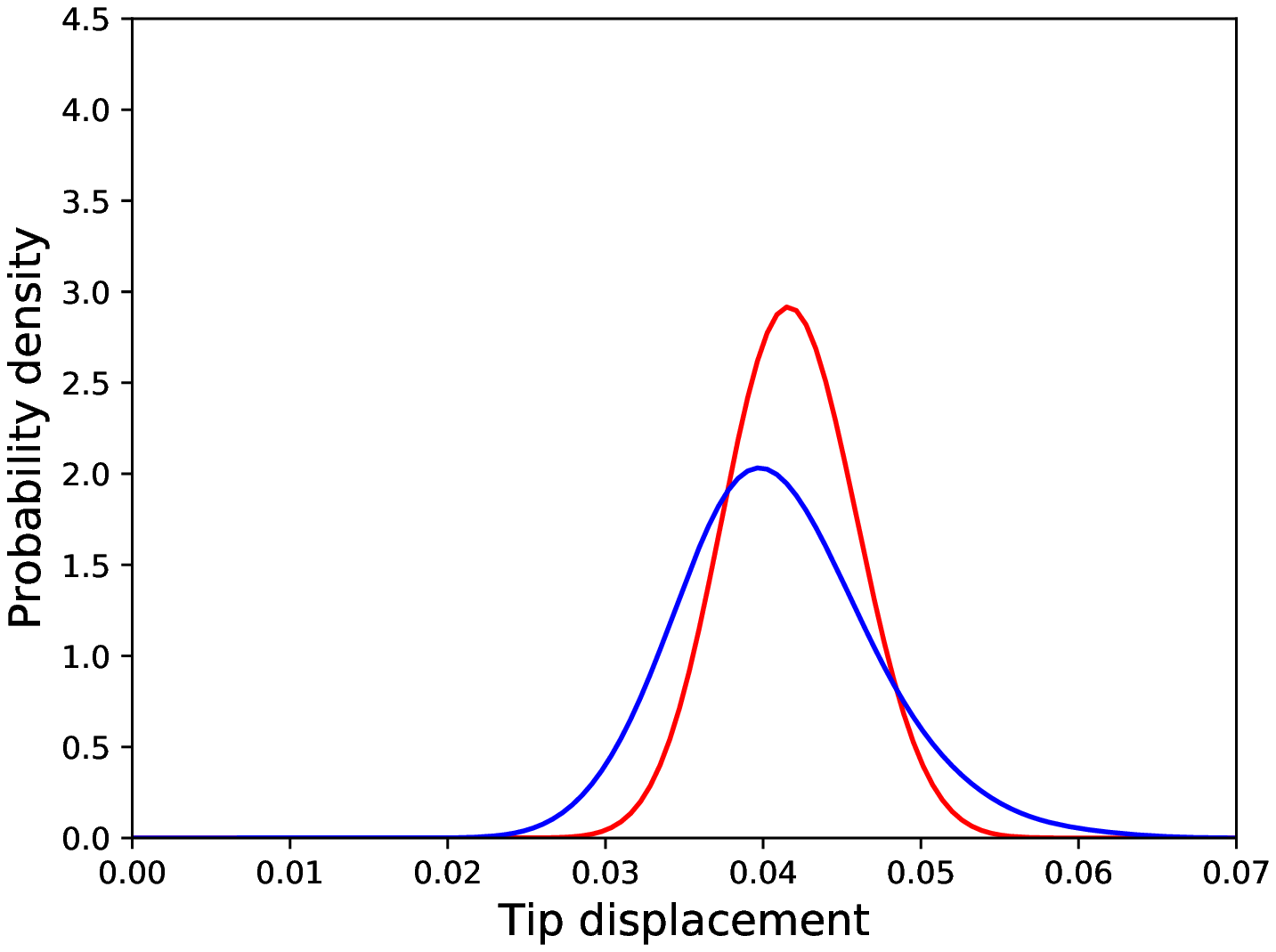}
    \caption{}
    \label{fig:sub_c3}
    \end{subfigure}
    \begin{subfigure}[b]{0.49\linewidth}
    \centering
    \includegraphics[width=0.99\linewidth]{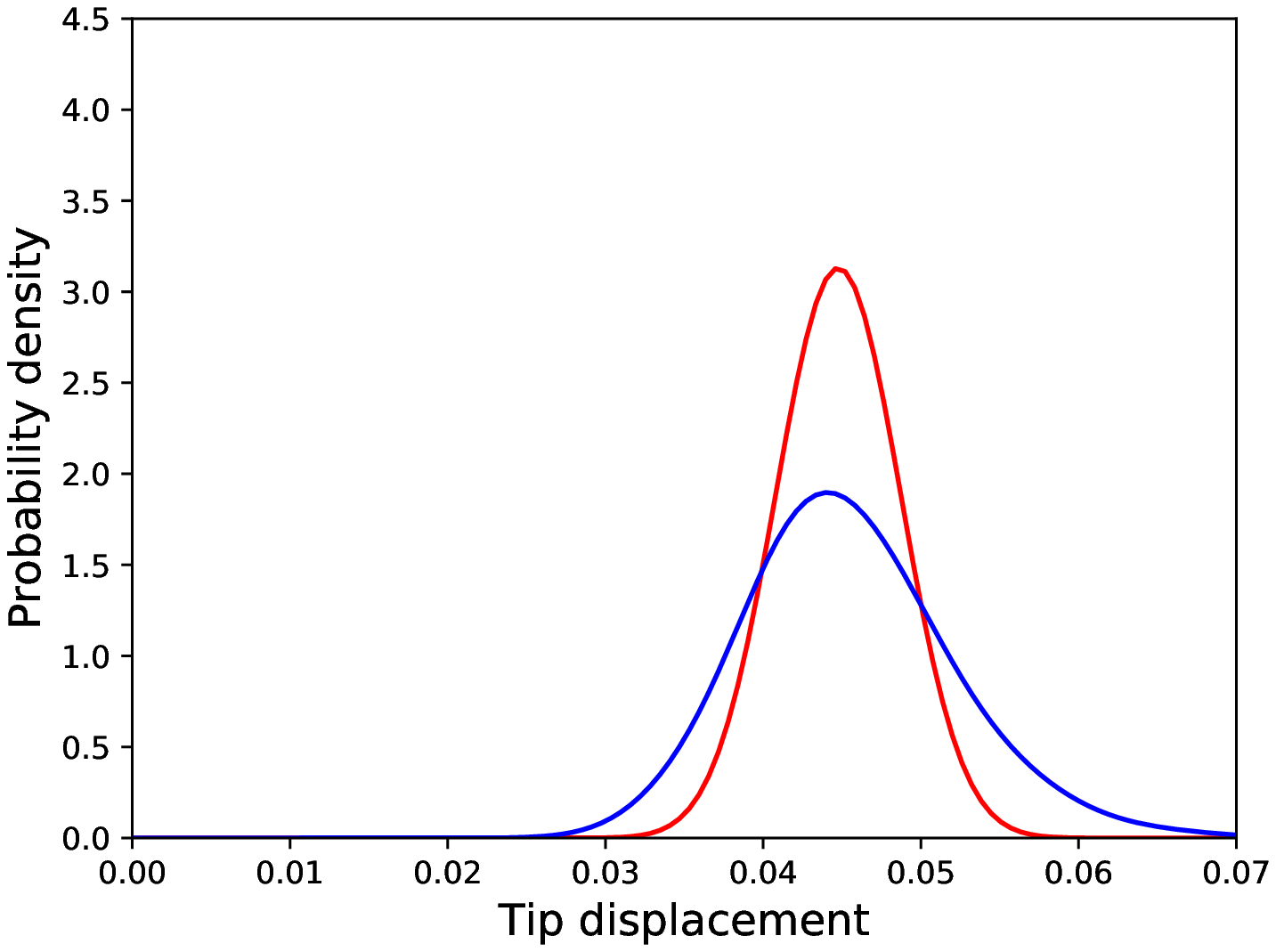}
    \caption{}
    \label{fig:sub_d3}
    \end{subfigure}
    \caption{Distributions of tip displacements corresponding to Monte Carlo samples (blue) and generated samples by the hybrid model (red) regarding the `extrapolation dataset' for the nonlinear problem for different load cases; (a) 320 load units, (b) 350 load units, (c) 370 load units, (d) 400 load units.}
    \label{fig:extrapolation_hybrid}
\end{figure}

The hybrid model is tested in exactly the same way. The model with the minimum KL divergence on the validation dataset had $1500$ neurons in its hidden layer. This model yielded KL divergence equal to $0.034$ on the validation dataset and equal to $0.038$ on the testing dataset. Samples of the predicted distributions in comparison with the real ones are shown in Figure
\ref{fig:hybrid_small_dataset_distribs}. Consequently, the same model was tested on the `extrapolation dataset'. Similar distribution comparison figures are shown in Figure \ref{fig:extrapolation_hybrid}. It is clear that the performance this time is better than in the case of the black-box model and this is confirmed by the average KL divergence observed on the second dataset, which is equal to $0.288$. One can see that the part of the physics which is correctly incorporated in the physics-based model (higher load values correspond to higher displacements), has affected positively the performance of the hybrid model on data outside its training domain.

\section{Discussion and conclusions}
\label{sec:discussion}
In the current work, models were developed that serve as mirrors of a specific structure. The models were chosen to be generative models in an attempt to take into account various uncertainties that might be present during the modelling procedure. A physics-based method, a data-driven method and a hybrid approach were presented along with their results and performance according to the chosen metric (KL-divergence).

SFEM is the physics-based method, which, calibrated according to acquired data from a structure, was tested as a potential mirror model of a beam structure. As for every physics-based method, if the model's physics formulation fits exactly the physics of the problem, the model is able to outperform any other method and have almost perfect accuracy in predicting the behaviour of the structure it describes. This situation is also the case for the SFE model in the concept described in this work. The SFE model was calibrated using a exhaustive search (within some range of values for each parameter) in the parameter space according to acquired data and the generated distributions almost perfectly fit the ones corresponding to the measured distributions.

Even though such a model outperforms every other method, if there is epistemic uncertainty, i.e. the model does not describe fully the physics or one does not know which parameter exactly is stochastic or the form of the stochasticity, it would not perform as well as in the previous case. In such cases, a data-driven method such as the one described herein should be employed. The nonlinear problem described is one such case.

Machine learning approaches could prove even more useful in cases where a physical phenomenon cannot be modelled by finite elements or any other such physics-based method. Situations like this are when one has no knowledge about how an environmental parameter affects the structure. For example, temperature and stiffness reduction (or increase) are usually assumed to have a linear relationship. If that does not stand, a finite element model, in which stiffness is reduced or increased linearly according to temperature, would not suffice regarding generation of accurate predictions.

Within the current framework, a cGAN was considered as a mirror of a simulated structure with material nonlinearities. The algorithm was able to perform well enough to be an $\epsilon$-mirror of the structure for values of $\epsilon$ calculated by the data. The algorithm fits the framework of controlled and uncontrolled environmental variables and is able to incorporate within its formulation any uncertainties that might exist. It can be also considered an $\alpha$-mirror, given different probabilities, for the observation to fall into the interval, as a function of $\alpha$. The latter type of mirror models may serve as a conservative aspect of the digital mirror model, since it provides an interval, within which any observation should fall into with probability $P(\alpha)$. The $\alpha$-mirror approach is irrelevant to the shape of the distribution and is only informed by the mean value and the standard deviation of the samples generated by the model.

Finally, a hybrid approach of the SFEM and cGAN is presented. The approach is based on using the cGAN to correct the predicted-by-the-SFE-model probability density functions. The hybrid model is able to perform slightly better than the cGAN in the nonlinear case. The advantages of the method are considered the information imposed into the model by the physics-based SFE method and the versatility of the cGAN algorithm in being able to perform regardless the nonlinearity of the problem. It appears to be a viable strategy for such applications, since it is further based on the physics of the problem, and also provides a framework of correcting the predictions of a physics-based generative model when unknown and even immeasurable parameters affect the result. Moreover, the presented results reveal that such models have greater extrapolation capabilities. Even if the physics-based model used does not completely incorporate the underlying physics of the problem, the model can still be informed by the valid parts of the captured physics and, that way, outperform black-box models in terms of accuracy in situations outside the training domain of the model.

\paragraph{Funding Statement}
This project has received funding from the European Union’s Horizon 2020 research and innovation programme under the Marie Skłodowska-Curie grant agreement No 764547. KW would like to thank the UK Engineering and Physical Sciences Research Council (EPSRC) for an Established Career Fellowship (EP/R003645/1). DW would like to acknowledge the support of EPSRC grand EP/R006768/1.

\paragraph{Competing Interests}
None

\paragraph{Ethical Standards}
The research meets all ethical guidelines, including adherence to the legal requirements of the study country.

\paragraph{Data Availability Statement}
The data used in the applications can be found in \url{https://drive.google.com/drive/folders/1Ykv_lC1SkRbWtWFoSE_PWK2zd1CgyZqx?usp=sharing} and the code to recreate the results of the methods described in the current work can be found in the GitHub repository: \url{https://github.com/GiorgTsial/cGANS_DT}.

\paragraph{Author Contributions}

Conceptualization, G.T., D.J.W., N.D., K.W.; methodology, G.T., D.J.W., N.D., K.W.; investigation, G.T.; validation, G.T.; visualisation, G.T.; software, G.T.; writing-original draft G.T.; writing - review and editing D.J.W., N.D., K.W.; funding acquisition, D.J.W., N.D., K.W.; project administration D.J.W., N.D., K.W.;supervision D.J.W., K.W.

\bibliographystyle{apalike}
\bibliography{cGANs_SFEM}

\section*{Appendix}
\label{sec:appendix}

\subsection*{Stochastic finite elements}
\label{sec:SFEM}

Finite element method (FEM) models \citep*{bathe2006finite} have been a very powerful and useful tool to numerically solve differential equations which describe mechanical systems. FEM transforms a continuous problem and a continuous differential equation into a discrete system of equations. Solutions are calculated only for a discrete number of points. Solutions for intermediate points are calculated using interpolation functions called shape functions. The continuous static differential equation most commonly used in FE models is given by,
\begin{equation}
    \label{eq:static_diff_eq}
    \int_{V} \sigma \epsilon dV = \int_{V} f^{V} dV + \int_{S_{e}} f^{S} dS
\end{equation}
where the LHS is the internal potential energy of a body, where $\sigma$ is stress, $\epsilon$ is strain and $V$ is the volume of the body of interest, the RHS is the potential work of the forces applied on the body, which is the integral over the volume of all the volume forces ($f^{V}$) plus the integral over the surface of the surface forces ($f^{S}$). Using a finite element formulation and minimising the total potential energy or the difference between the two sides of equation, one gets,
\begin{equation}
    \label{eq:static_fem_eq}
    [K]\{U\} = \{F\}
\end{equation}
where $[K]$ is the stiffness matrix of the structure for a specific meshing scheme applied, $\{U\}$ is the displacement vector of the nodal displacements and $\{F\}$ is the equivalent nodal force vector to the total applied forces on the body.

For dynamic problems, inertia forces are introduced into the RHS of equation (\ref{eq:static_diff_eq}) resulting in,
\begin{equation}
    \label{eq:dynamic_diff_eq}
    \int_{V} \sigma \epsilon dV = \int_{V} f^{V} dV + \int_{S_{e}} f^{S} dS - \int_{V} \rho \ddot{u} dV - \int_{V} c \dot{u} dV
\end{equation}
where $\rho$ refers to the mass density function of the body, $\ddot{u}$ is the acceleration, $c$ is the damping parameter and $\dot{u}$ is the velocity at every point. This equation holds for every time instant $t$ of the simulation. Once again, following the FEM formulation and defining a discretisation of the body, the system of equations are,
\begin{equation}
    \label{eq:dynamic_fem_eq}
    [M]\{\ddot{U}\}(t) + [C]\{\dot{U}\}(t) + [K]\{U\}(t) = \{F\}(t)
\end{equation}
where $[M]$, $[C]$, $[K]$ are the mass, damping and stiffness matrices and $\{\ddot{U}\}$, $\{\dot{U}\}$, $\{U\}$ the nodal accelerations, velocities and displacement vectors respectively.

The matrices in equations (\ref{eq:static_fem_eq}) and (\ref{eq:dynamic_fem_eq}) often are calculated assuming deterministic structural parameters, e.g. Young's modulus ($E$), Poisson's ration ($\nu$), mass density ($\rho$) etc. However, these parameters are quite often \textit{not} deterministic. Especially in composites, such parameters are almost certainly random. Young's modulus might vary within the volume of the body one tries to analyse using FEM. These variations are not just discrete variables, almost certainly they are \textit{stochastic processes} \citep*{papoulis2002probability}. A stochastic process has a correlation function that defines how values over some distance (spatial or temporal) are correlated. Furthermore, every point has a mean value and a variance defined by functions $\mu(x)$ and $\sigma^2(x)$ respectively. If the two functions are constant everywhere over the space where the process is defined, then it is called a \textit{stationary} process.

\begin{figure*}
    \centering
    \begin{tikzpicture}
        \node[] (A) at (0.0, 0.0) {};
        \node[] (B) at (5.0, 0.0) {};
        \node[] (C) at (5.0, 0.6) {};
        \node[] (D) at (0.0, 0.6) {};

        \draw[line width=0.6mm] (0.0, 0.0) -- (5.0, 0.0);
        \draw[line width=0.6mm] (5.0, 0.0) -- (5.0, 0.6);
        \draw[line width=0.6mm] (5.0, 0.6) -- (0.0, 0.6);
        \draw[line width=0.6mm] (0.0, 0.6) -- (0.0, 0.0);

        \draw[-{>[scale=2.5, length=2, width=3]}, line width=0.5mm] (0.0, 1.0) to (0.0, 4.0);

        \node[] (E) at (-0.4, 3.6) {E};
        \node[] () at (-0.4, 2.5) {$10^{9}$};

        \draw[line width=0.3mm] (-0.1, 2.5) -- (0.1, 2.5);
        \draw[line width=0.3mm] (-0.0, 2.5) -- (5.0, 2.5);

        \draw[line width=0.3mm, red] (0.0, 2.8) to[out=-10,in=-150] (2.0, 3.5) to[out=30,in=140] (5.0, 2.0);

        \draw[line width=0.3mm, blue] (0.0, 1.9) to[out=40,in=-130] (3.0, 2.3) to[out=50,in=220] (5.0, 3.0);

        \draw[line width=0.2mm] (0.0, 0.0) -- (-0.1, -0.1);
        \draw[line width=0.2mm] (0.0, 0.1) -- (-0.1, -0.0);
        \draw[line width=0.2mm] (0.0, 0.2) -- (-0.1,  0.1);
        \draw[line width=0.2mm] (0.0, 0.3) -- (-0.1,  0.2);
        \draw[line width=0.2mm] (0.0, 0.4) -- (-0.1,  0.3);
        \draw[line width=0.2mm] (0.0, 0.5) -- (-0.1,  0.4);
        \draw[line width=0.2mm] (0.0, 0.6) -- (-0.1,  0.5);
    \end{tikzpicture}
    \caption{Cantilever beam with constant Young's modulus (black line) or spatially varying (red and blue lines).}
    \label{fig:cantilever_varying_E}
\end{figure*}
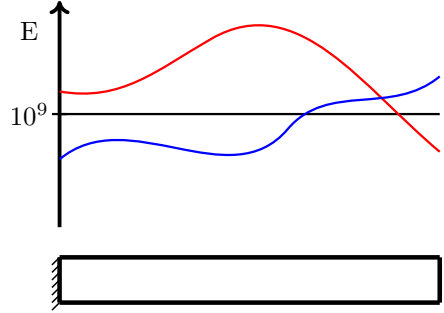

An example of a deterministic consideration of Young's modulus and a stochastic Young's modulus for a cantilever beam problem would look like the functions shown in Figure \ref{fig:cantilever_varying_E}. The black line represents how conventional FEM is applied assuming a constant and deterministic value for structural parameters, while the red and blue lines show samples drawn from a stochastic process. A way to address problems like this, in general, would be to sample from the stochastic process and follow a Monte Carlo scheme. This approach would require solving a large number of deterministic FEM problems, performing a sensitivity analysis and post-processing the results in order to infer the statistics of the quantities of interest. In this case, the quantity of interest could be the displacement of the tip and a probability distribution would be defined over the potential values of this displacement.

The SFEM already mentioned, is a quite popular means of propagating uncertainty from material properties and randomness in the excitation forces into the response characteristics of a structure. In contrast to a Monte Carlo approach, SFEM infers the distribution of interest as a function of a set of discrete normally-distributed variables. In order to explain the SFEM formulation of a problem, first a method to decompose the random field is needed. The expansion method used herein is the Karhunen-Loève expansion \citep*{sudret2000stochastic, loeve1977elementary}.

\subsection*{Karhunen-Loève (KL) expansion}

The KL expansion is based on the spectral (i.e. eigenvalue) decomposition of the autocovariance function of the given random field. Given a random field $H(x)$ and its autocovariance function $C_{HH}(x, x')$, any realisation of the field $H(x)$ is expanded over a basis of deterministic functions, defined by the eigenvalue problem \citep*{sudret2000stochastic},

\begin{equation}
    \int_{\Omega}C_{HH}(x, x')\phi(x')d\Omega_{x'} = \lambda_{i}\phi_{i}(x)
    \label{eq:eigenval_problem}
\end{equation}
where the kernel $C_{HH}(x, x')$ is a kernel autocovariance function, bounded, symmetric and positive definite and $\Omega_{x'}$ is the total space of $x'$. The set of eigenvalues $\lambda_{i}$ and eigenfunctions/eigenvectors $\{\phi_{i}\}$ form a complete basis to express every realisation of the field as,

\begin{equation}
    H(x, \theta) = \mu(x) + \sum_{i=1}^{\infty}\sqrt{\lambda_{i}}\xi_{i}(\theta)\phi_{i}(x)
    \label{eq:realisations_expanded}
\end{equation}
where $\xi_{i}(\theta)$ are the coordinates of the realisation which are independent random variables and $\theta$ is the random event.

In practice, if one wishes to define an expansion of a random field and either generate random realisations or use it, as it will subsequently be used for the purposes of stochastic FEM, a truncation is performed at $m$th order yielding,
\begin{equation}
    H(x, \theta) = \mu(x) + \sum_{i=1}^{m}\sqrt{\lambda_{i}}\xi_{i}(\theta)\phi_{i}(x)
    \label{eq:realisations_expanded_trunc}
\end{equation}
where it is assumed that the eigenvalues $\lambda_{i}$ are sorted in ascending order.

\subsection*{Solution of static SFEM problems}

After defining the stochastic field in the finite element formulation of a problem, this field is expressed via the KL expansion. This approach leads to the stiffness matrix of equation (\ref{eq:static_fem_eq}) appearing as a summation over stiffness matrices constructed according to the eigenfunctions of equation (\ref{eq:realisations_expanded_trunc}); more specifically,

\begin{equation}
    K(\theta) = K_{0} + \sum_{i=1}^{m}\xi_{i}(\theta)K_{i}
    \label{eq:stiffness_realisations_expanded_trunc}
\end{equation}
where the $K_{i}s$ are deterministic matrices that are calculated using the eigenfunctions $\phi_{i}$ from equation (\ref{eq:realisations_expanded_trunc}). Realisations from the set ${\xi_{i}}$ can be used in order to generate realisations for the stiffness matrix $K(\theta)$, in case a Monte Carlo simulation is to be followed.

Now, substituting equation (\ref{eq:stiffness_realisations_expanded_trunc}) into equation (\ref{eq:static_fem_eq}) yields (assuming a deterministic load),

\begin{equation}
    [K_{0} + \sum_{i=1}^{m}K_{i}\xi_{i}(\theta)]U(\theta) = \sum_{i=0}^{m}[K_{i}\xi_{i}(\theta)]U(\theta) = F
    \label{eq:SFEM_0}
\end{equation}

In order to move further from this point, a \textit{polynomial chaos expansion} (PCE) is used \citep*{ghanem2003stochastic}. PCE has been a very efficient and widely used method in engineering applications, for meta-modelling, structural health monitoring, etc. \citep*{spiridonakos2015metamodeling, spiridonakos2016polynomial}. Here it is used to decompose the field of displacements $U(\theta)$ yielding,

\begin{equation}
    U(\theta) = \sum_{j=0}^{\infty}U_{j}\Psi_{j}(\theta)
    \label{eq:PCE_Us}
\end{equation}
where the ${\Psi_{j}(\theta)}$, for $j = 0, 1, 2...$ are polynomials defined in ${\xi_{i}(\theta)}$ which satisfy,

\begin{subequations}
    \begin{align}
        \begin{split}
            \Psi_{0} \equiv 1
        \end{split}\\
        \begin{split}
            \mathbb{E}_{\xi(\theta)}[\Psi_{j}] = \mathbb{E}_{\theta}[\Psi_{j}] = 0 \qquad   j > 0
        \end{split}\\
        \begin{split}
            \mathbb{E}_{\xi(\theta)}[\Psi_{j}(\theta)\Psi_{k}(\theta)] = \mathbb{E}_{\theta}[\Psi_{j}(\theta)\Psi_{k}(\theta)] = 0 \qquad  j \neq k
        \end{split}
    \end{align}
    \label{eq:polynomials_c}

\end{subequations}
i.e. the terms ${\Psi_{j}(\theta)}$ are orthogonal in expectation ($\mathbb{E}[]$).

By truncating at $P$ terms and substituting into equation (\ref{eq:SFEM_0}), one obtains,

\begin{equation}
    \sum_{i=0}^{m}K_{i}\xi_{i}(\theta)\sum_{j=0}^{P-1}U_{j}\Psi_{j}(\theta) = F
    \label{eq:SFEM}
\end{equation}

The solution to this equation can be found by minimising the error,

\begin{equation}
    \epsilon_{m, P} = \sum_{i=0}^{m}K_{i}\xi_{i}(\theta)\sum_{j=0}^{P-1}U_{j}\Psi_{j}(\theta) - F
\end{equation}

The best approximation of the exact solution $U(\theta)$ in the space $H_{P}$ spanned by the $\{\Psi_{k}\}_{k=0}^{P-1}$ is obtained by minimising this residual in a mean-square sense. In a Hilbert space this is equivalent to requiring that the residual be orthogonal to $H_{P}$, yielding,

\begin{equation}
    \mathbb{E}_{\theta}[\epsilon_{m, P} \Psi_{k}] = 0 \qquad k = 0, ... P-1
    \label{eq:expectation_eq_zero}
\end{equation}

Substituting equation (\ref{eq:expectation_eq_zero}) in equation (\ref{eq:SFEM}), one finds,
\begin{equation}
\mathbb{E}_{\theta}[\sum_{i=0}^{m}\sum_{j=0}^{P-1}K_{i}\xi_{i}\Psi_{j}(\theta)\Psi_{k}(\theta)U_{j}] = \mathbb{E}_{\theta}[\Psi_{k}(\theta)F]
\end{equation}

Introducing the following notation,
\begin{equation}
    c_{ijk} = \mathbb{E}_{\theta}[\xi_{i}\Psi_{j}\Psi_{k}]
\end{equation}

\begin{equation}
    F_{k} = \mathbb{E}_{\theta}[\Psi_{k}F]
\end{equation}
and,
\begin{equation}
    K_{jk} = \sum_{i=0}^{M}c_{ijk}K_{i}
\end{equation}

the stochastic FEM equation finally becomes,

\begin{equation}
    \sum_{j=0}^{P-1}K_{jk} U_{j} = F_{k} \qquad k = 0, 1 ... P-1
\end{equation}

In this equation, every $U_{j}$ is an N-dimensional vector, where N is the number of degrees of freedom in the system. In total, the $P$ equations from above can be written as,

\begin{equation}
    \begin{bmatrix}
        K_{00} & \dots & K_{0, P-1}\\
        K_{10} & \dots & K_{1, P-1}\\
        \vdots &       & \vdots \\
        K_{P-1, 0} & \dots & K_{P-1, P-1}
    \end{bmatrix}
    \begin{bmatrix}
        U_{0}\\
        U_{1}\\
        \vdots\\
        U_{P-1}
    \end{bmatrix}
    =
    \begin{bmatrix}
        F_{0}\\
        F_{1}\\
        \vdots\\
        F_{P-1}
    \end{bmatrix}
    \label{eq:SFEM_system}
\end{equation}

Having solved this system, for $U_{j}$, samples $U(\theta)$ can be generated by sampling $\xi_{i}(\theta)$ values and using equation (\ref{eq:PCE_Us}). Thus, samples of the distribution of all displacements are generated. Solving this system is equivalent to solving the problem for every potential value of the random parameters. The augmented matrices in equation
(\ref{eq:SFEM_system}) are of dimension $NP\times NP$, where $N$ are the degrees of freedom of the deterministic problem and $P$ the order of the PCE. Solving such a system instead of a deterministic one is much more computationally intense, i.e. $\mathcal{O}(N^{3}P^{3})$. However, it might not be as computationally inefficient as a sufficient number of Monte Carlo simulations. Solving the system yields samples and therefore distributions, establishing SFE models as generative models.

\end{document}